\DocumentMetadata
{
	lang		= en-US, 
	pdfversion  = 1.7,
	pdfstandard = a-2b,
}

\documentclass[twoside]{mitthesis}

\usepackage{iftex}
\ifPDFTeX
  \usepackage[T1]{fontenc}
  \usepackage[utf8]{inputenc}
\fi

\usepackage{listings}

\usepackage{xcolor}
\usepackage{adjustbox}
\usepackage{amssymb} 
\usepackage{amsmath} 
\usepackage{tabularx}
\usepackage[table]{xcolor}

\usepackage[version=4]{mhchem}

\usepackage{lipsum}
\IfPackageAtLeastTF{lipsum}{2021/09/20}{\setlipsum{auto-lang=false}}{} 

\usepackage{booktabs}

\usepackage{array}

\usepackage{dcolumn}
  \newcolumntype{d}[1]{D{.}{.}{#1}}

\usepackage{longtable}
\usepackage[nopatch=footnote]{microtype}

\usepackage{multicol}

\usepackage{booktabs}
\usepackage{tabularx}
\usepackage{tikz}
\usetikzlibrary{arrows.meta, positioning}

\lstdefinelanguage{Dafny}{
  morekeywords={
    method, returns, ensures, requires, invariant, assert, var, while,
    if, else, return, lemma, ghost, function, predicate, forall, exists,
    decreases, modifies, reads, nat, int, bool, true, false
  },
  sensitive=true,
  morecomment=[l]{//},
  morecomment=[s]{/*}{*/},
  morestring=[b]"
}

\lstdefinelanguage{Lean}{
  morekeywords={
    theorem, example, def, by, exact, intro, intros, apply, rw, simp,
    have, show, from, fun, match, with, Type, Prop, Nat
  },
  sensitive=true,
  morecomment=[l]{--},
  morecomment=[s]{/-}{-/},
  morestring=[b]"
}

\usepackage[
  style=ext-numeric-comp,
  giveninits=true,
  maxbibnames=10,
  sorting=none,
  language=american,
  natbib=true
]{biblatex}

    \AtEveryBibitem{%
      \ifentrytype{article}{%
        \renewbibmacro{in:}{}
		\renewbibmacro*{issue+date}{
		\iffieldundef{issue}%
        	{}%
        	{\printfield{issue}%
         	\setunit*{\addcomma\space}
        	}%
			\printdate
		}
      }{}
    }
	\renewbibmacro*{volume+number+eid}{%
        \printtext[bold]{\printfield{volume}}
        \iffieldundef{number}
          {} 
          {\printtext[parens]{\printfield{number}}} 
        \setunit{\bibeidpunct}%
        \printfield{eid}
    }
\usepackage{setspace}
\hypersetup{%
	pdfsubject={Template for writing MIT theses with the mitthesis class},
	pdfkeywords={Massachusetts Institute of Technology, MIT},
	pdfurl={},
	pdfcontactemail={},
	pdfauthortitle={},
}

\DeclareMathOperator{\Erf}{erf}
\DeclareMathOperator{\Erfc}{erfc}
\DeclarePairedDelimiterXPP\erf[1]{\Erf\mkern1mu}(){}{#1}
\DeclarePairedDelimiterXPP\erfc[1]{\Erfc\mkern1mu}(){}{#1}

\begin{document}

\title{Automating Formal Verification with Reinforcement Learning and Recursive Inference}

\Author{Max Tan}{Department of Electrical Engineering and Computer Science}[S.B. in Physics and Artificial Intelligence and Decision Making, Massachusetts Institute of Technology, 2025]

\Degree{Master of Engineering in Electrical Engineering and Computer Science}{Department of Electrical Engineering and Computer Science}

\Supervisor{Max Tegmark}{Professor of Physics}[Department of Physics]

\Acceptor{Katrina LaCurts}{Chair, Master of Engineering Thesis Committee}{}

\DegreeDate{May}{2026}

\ThesisDate{May 22, 2026}

\maketitle

\begin{abstract}
Automated formal verification remains challenging for large language models because data for proof assistants and verification-aware languages is scarce, and correctness depends on satisfying precise machine-checkable specifications rather than producing plausible code. This thesis studies how verifier environments can improve LLM generation of verified programs and proofs through reinforcement learning from verifiable rewards (RLVR) and verifier-guided inference-time search. First, we train open-source models in Dafny with RLVR using Group Relative Policy Optimization (GRPO) and related variants, assembling generated candidates into complete programs and scoring them with compiler and verifier outcomes. Initial experiments on an APPS-derived Dafny dataset increased verified reward from 2.2\% to 58.1\%, but revealed specification hacking, where models exploit weak formal specifications instead of implementing the intended solutions. After filtering underspecified and vulnerable tasks, multi-turn RLVR on the refined benchmark improves the verified pass rate from 9.7\% to 31.1\%. Second, we develop a verifier-guided inference scaffold in Lean that treats proof generation as structured search over decomposed subgoals, verifier feedback, diagnostics, and repair. With a fixed base model, the full scaffold with proof reviser improves pass rate on an initial VeriCoding pilot set from 46.2\% under direct repair to 69.2\%. On the larger VERINA dataset, whole-task decomposition plus proof reviser solves 7 of 42 previously unsolved tasks. We also introduce Dalek-Bench, a repository-scale Lean benchmark derived from the Rust \texttt{curve25519-dalek} verification project; preliminary results remain weak, indicating that stronger progress evaluation and task-specific tool-use policies are still needed. Overall, formal verifiers become useful to LLMs when they are treated not just
as checkers, but as sources of reward, feedback, and search control. However, this only
succeeds when the environment supplies clean data, robust specifications, and
structured ways to act on verifier feedback. Benchmarks and code are available on request from \href{mailto:maxt114@mit.edu}{maxt114@mit.edu}
\end{abstract}
\chapter*{Acknowledgments}
\addcontentsline{toc}{chapter}{Acknowledgments}

I am deeply grateful to the many people who made this thesis possible.

\vspace{0.5em}

First, I would like to thank my advisor, Professor Max Tegmark, who introduced me
to the field of formal verification and brought so much excitement to this
problem space. Our weekly meetings were motivating not only technically, but also
because they helped me keep sight of the real-world stakes of this work. Your
feedback throughout the project, and especially on this final thesis, has been
invaluable.

\vspace{0.5em}

I also want to thank the other members of the lab, particularly David Baek,
Ionel Chiosa, and Leo Yao. Our weekly lunches were always full of interesting
conversation, and each of you played an important role in making this thesis
what it is today, from engineering advice and technical feedback to broader
research guidance.

\vspace{0.5em}

I am also grateful to the mentors and collaborators who shaped my broader
research path at MIT. Thank you to my undergraduate advisors, Professor Janet
Conrad and Professor Jing Kong, for their guidance throughout my time as an
undergraduate. I would also like to thank Yufeng (Bright) Ye, Professor Kevin
O'Brien, Professor Jesse Thaler, Dana Choi, Professor Dylan Hadfield-Menell, and Joyce Luo
for their mentorship, collaboration, and encouragement at different stages of my
research journey.

\vspace{0.5em}

To my family---Mom, Dad, and Elliot---thank you for supporting me throughout my
entire academic journey. I would not be where I am today without your support and encouragement.

\vspace{0.5em}

Finally, to all of my friends at MIT: you know who you are. This place has been
special to me because of all of you.

\tableofcontents
\listoffigures
\listoftables


\doublespacing

\chapter{Introduction}

\section{Emerging Cybersecurity Risks}

The cybersecurity landscape is undergoing a huge shift, and AI is at the center. Modern AI systems can now find and 
exploit software flaws on their own, at speeds and complexities that are hard to match. In 
late 2025, Anthropic claimed that a Chinese state-sponsored group had 
used Claude Code (Anthropic's AI coding assistant) to run a 
cyber-espionage campaign against
government agencies, banks, and chemical manufacturers 
\cite{anthropic_gtg1002}. The AI did 80--90\% of the work itself: 
scanning targets for weaknesses, stealing passwords, moving deeper into 
networks, and copying out sensitive data, all at thousands of operations 
per minute. Anthropic called it the first known case of a large-scale 
cyberattack carried out almost entirely by an AI.

Around the same time, AI began showing up in attacks both as a direct weapon and as a way into vast networks. In March 2026, attackers 
slipped malicious code into LiteLLM, a tool that many companies use to 
connect their applications to AI models \cite{sentinelone_how_2026}. 
Several companies were infected automatically: their AI coding 
assistants, given broad permission to install software, downloaded and 
ran the tampered version on their own. On the 
offensive side, the security startup CodeWall has demonstrated 
autonomous AI agents finding critical security flaws in live systems 
within hours: a two-hour break-in to McKinsey's internal AI platform 
that exposed tens of millions of internal consultant messages 
\cite{codewall_mckinsey}; a complete takeover of an AI recruiting 
platform that the agent infiltrated by phoning 
the company's voice bot \cite{codewall_jackjill}; and similar hacking into Bain \& Company \cite{codewall_bain}. IBM estimates that roughly 
one in six recent corporate data breaches in 2025 involved AI-driven 
components \cite{ibm_data_breach_2025}, which is bound to increase further.

Rather than replacing past threats, AI incidents amplify existing cyber risks that continue to cause major damage. In 2024, a faulty CrowdStrike software 
update crashed millions of Windows computers worldwide, grounding flights 
and disrupting hospital operations \cite{crowdstrike_outage_2024}. In 
2021, a configuration error took Facebook, Instagram, and WhatsApp 
offline for hours \cite{meta_outage_2021}. And in late 2025, Amazon's 
own Kiro AI coding agent, which was granted autonomous access to a live AWS 
environment, decided that the best fix for a routine issue was to 
``delete and recreate the environment,'' triggering a 13-hour outage of 
AWS Cost Explore. Amazon employees even reported a second, similar 
AI-related incident shortly after \cite{reuters_aws_kiro_2026}. 

Conventional defenses such as testing, code review, and even AI-driven bug detectors, remain essential but are 
fundamentally limited. They can only check behavior on a sample of 
inputs and not every possible input, scenario, 
or clever attack. \textit{Formal verification} offers something stronger: a 
way to mathematically \emph{prove} that software behaves correctly on 
every possible input. The technique is mature and proven in deployment, 
but, as we next discuss, gated by the expert 
labor required to use it.

\section{The Relevance of Formal Verification}

Formal verification is the practice of mathematically proving that a 
piece of software meets a precise specification of what it should do. 
Rather than running a program on test inputs and observing the outputs, 
the developer writes down what correctness \emph{means} for their particular function (e.g., the program 
never reads memory it shouldn't, it always terminates within a bounded 
number of steps, it correctly implements a particular protocol). We can 
then construct a proof that the program always satisfies that 
specification, checked end-to-end by a machine. 

The idea dates to the 1960s, when computer scientists such as Tony Hoare 
and Edsger Dijkstra developed logical methods for reasoning about 
programs \cite{hoare_axiomatic_1969}. For decades, the work remained 
largely theoretical. Even small proofs required much human 
effort. The breakthrough came as automated reasoning matured, and two 
complementary strands of tools emerged. \emph{Interactive proof 
assistants} such as Coq (recently renamed Rocq \cite{rocq_prover}) 
and Lean \cite{moura_lean4_2021} give the human full control over the 
proof while relying on the machine to check every step. \emph{Verification 
languages} such as Dafny \cite{gauci_dafny_2014}, F$^*$ 
\cite{swamy_fstar_2016}, and Verus \cite{lattuada_verus_2023} attach 
proofs directly to a programming language and offload the routine logic 
to automated solvers, asking the developer for help only with the 
parts that require genuine non-repetitive insight. The two approaches trade-off automation and expressive 
power: more automation makes proofs faster to write but limits what 
they can express; more interactivity makes proofs slower but allows 
a wider range of mathematical reasoning.

These tools have produced verified software at a scale that was 
unimaginable two decades ago. AWS has formally verified Cedar, its 
open-source authorization language used by Cloudflare, MongoDB, AWS 
itself, and others (and recently accepted into the Cloud Native 
Computing Foundation), with a Lean specification rigorously tested 
against the production Rust implementation 
\cite{cedar_cncf_2025}. Verus is in active 
industrial use at Microsoft and Amazon, and underpinned two of the 
three best papers at OSDI 2024 (Anvil, VERISMO) \cite{sun_anvil_2024, zhou_verismo_2024}. The same 
technology is reshaping mathematics: Lean's Mathlib library has 
formalized hundreds of thousands of theorems and is now used by 
working mathematicians \cite{The_mathlib_Community_2020}; Peter Scholze's Liquid 
Tensor Experiment fully formalized a frontier result in advanced 
mathematics \cite{liquid_tensor_experiment}; and Terence Tao formalized 
his polynomial Freiman--Ruzsa theorem in Lean in 2023 and is now 
formalizing his classic textbook \emph{Analysis I} chapter by chapter 
\cite{pfr_project, noauthor_lean_nodate}. Additionally, driven by AI, Harmonic's Aristotle has independently formalized and 
proved long-standing open Erdős problems (including Problems 728 and 
1026) that had resisted human effort for decades 
\cite{xenaproject_formalization_2025}.

The cost of verification, however, concentrates at two distinct 
bottlenecks. The first is \emph{specification}: translating an 
informal description of what software should do into a machine-readable form. 
A specification that is subtly wrong or incomplete will yield an inconsequential proof as the bug survives. 
Specification writing is far from trivial: for the Liquid Tensor Experiment mentioned earlier, translating just the statement of the key intermediate 
theorem into Lean took a team of expert mathematicians an entire 
month \cite{xenaproject_beyond_2022}.
The second barrier is \emph{proof generation}: constructing the argument 
that a particular implementation actually meets its specification, 
step by step. Proof-to-code ratios in published 
verification efforts often run from 3:1 to 10:1, and large projects 
often take years of expert effort \cite{lattuada_verus_2023,  
Klein_2009}. The community that can do this productively at 
scale remains small, and as a result, formal verification today 
protects a narrow set of high-stakes components such as kernels, compilers, 
cryptography and authorization systems, while the surrounding software that 
run our secure systems ship unverified. Both remain open research challenges, and both are precisely where 
recent advances in Large Language Models are making their sharpest 
inroads.

\section{The Rise of Large Language Models}

Large Language Models (LLMs) have undergone a rapid shift in their code and
reasoning capabilities, from generating plausible-looking snippets to
acting as autonomous software engineers capable of reading, executing,
debugging, and reasoning about large codebases end-to-end. The clearest
sign of this shift is the proliferation of agentic coding tools: Claude
Code, OpenAI's o-series, and similar systems now write production code,
navigate complex repositories, propose and apply multi-file edits, and
iterate on their own outputs in response to test failures, all without
human guidance between steps \cite{anthropic_claude_code,noauthor_introducing_nodate-1}.
At the frontier, the depth of this code understanding has crossed a
qualitative threshold. Claude Mythos Preview, Anthropic's most capable
unreleased model, is claimed to have discovered thousands of previously unknown
vulnerabilities across every major operating system and web browser (including a 27-year-old bug in OpenBSD, and a 17-year-old remote code
execution flaw in FreeBSD) despite receiving no explicit security
training. The capabilities simply emerged as a result of improvements in
code comprehension, step-by-step reasoning, and autonomous tool use
\cite{noauthor_project_nodate,noauthor_our_nodate}.

The same period has seen a parallel revolution in formal mathematical
theorem proving. Formal theorem
proving is a natural testbed for LLMs because it is objectively verifiable: a
proof either compiles in Lean or it does not, with no partial credit.
In 2024, Google DeepMind's AlphaProof achieved silver-medal performance
at the International Mathematical Olympiad (IMO), the first AI system
to do so in a formally verified setting \cite{hubert2025alphaproof}. In
2025, Harmonic AI's Aristotle went further, achieving IMO gold-medal
performance and independently formalizing and proving long-standing open
Erd\H{o}s problems in Lean \cite{harmonic2025aristotle}. The open-source community has tracked closely, producing a rapid 
succession of systems each improving on the last. DeepSeek-Prover 
trained a model to write proofs and then used Lean's automatic 
checker as a teacher using Reinforcement Learning, letting the system bootstrap its own improvement 
\cite{xin_deepseek-prover_2024}. Goedel-Prover repeated this cycle for smaller models, yet still outperfomring DeepSeek-Prover by nearly eight percentage 
points on standard benchmarks \cite{lin_goedel-prover_2025}. Seed-Prover 
1.5 took the approach further still, solving nearly 88\% of problems 
on PutnamBench---a set of competition and graduate-level mathematics 
problems that a generation ago would have been considered out of 
reach for any automated system \cite{chen2025seedprover}.

These results trace to two complementary techniques that
this thesis is organized around. 

\begin{enumerate}
    \item \textbf{Post-training}: Training the model directly on formal 
    verification tasks, often on large collections of existing 
    proofs (Goedel-Prover started from every verified proof in 
    Mathlib \cite{lin_goedel-prover_2025}). Using the verifier as an 
    automatic judge, the model can also learn from its own successes and 
    failures without any human labeling (the mechanism behind both 
    DeepSeek-Prover and AlphaProof \cite{xin_deepseek-prover_2024, 
    alphaproof2024}), then feed successful proofs back into training 
    so the model progressively improves on harder problems, and 
    also compress the knowledge of a large model into a smaller, 
    faster one (distillation).

    \item \textbf{Inference-time Scaffolding}: improving how the 
    model searches at test time without any additional training, 
    by generating many candidate proofs in parallel and keeping 
    the first that works (Goedel-Prover tried up to 25,600 
    candidates for its hardest problems \cite{lin_goedel-prover_2025}), 
    exploring proof steps like branches of a tree and backtracking 
    when a path fails (used by DeepSeek-Prover to recover from 
    dead ends \cite{xin_deepseek-prover_2024}), letting the model read 
    the verifier's error messages and try again the way a 
    programmer iterates on compiler feedback, and breaking a hard 
    problem into smaller independently solvable steps before 
    tackling each one (the core idea behind DeepSeek-Prover V2 
    \cite{ren_deepseek-prover-v2_2025}).
\end{enumerate}

The critical enabler for LLM-based approaches is the
\emph{verifier}: because Lean, Dafny, and Verus provide binary,
automatic, and deterministic feedback on every proof attempt, they act
as an ideal reward signal for correctness. The
question this thesis addresses is whether these techniques transfer
from formal mathematics to the harder problem of automatically
generating verified software a non-trivial domain transfer.

\section{Thesis Overview}

\textit{Can we automate the generation of formally verified code from 
written specifications?} While recent work in automated 
theorem proving has shown that LLMs can learn to construct formal 
mathematical proofs, formal software verification presents a harder 
challenge: programs require specifications that must be conceived 
alongside implementation, the corpus of verified code is 
orders of magnitude smaller than mathematical proof libraries, and 
the space of possible solutions and ``proof paths" is less structured 
than the action space available in mathematics. This thesis 
investigates two complementary families of approach to closing this 
gap, evaluated primarily on the Vericoding benchmark 
\cite{10.48550/arxiv.2509.22908} and a new real-world benchmark we introduce 
derived from the Dalek elliptic curve cryptography library 
\cite{dalek_lite}---a formally verified package used in production 
systems including Signal.

\begin{enumerate}
    \item \textbf{Part~I: Post-Training} First we investigate post-training Large Language Models to improve on Dafny verification tasks. We apply long 
    chain-of-thought reinforcement learning from verifier feedback,
    using similar principles that powered successes like Seed-Prover and Goedel-Prover, but targeting 
    formally verified code generation rather than mathematical theorem 
    proving. Initial experiments showed promising results, but deeper analysis 
    revealed evidence of reward hacking: models were exploiting 
    structural holes in the proposed problems rather than learning 
    genuine formal verification reasoning. Diagnosing and blocking this hacking behavior led to more honest baselines and motivated a sequence of 
    further developments: supervised fine-tuning on curated Dafny 
    examples to establish stronger latent knowledge, a multi-turn 
    training environment that exposes models to more refinement across multiple repair attempts, 
    and distillation through rollouts from stronger state-of-the-art models to transfer reasoning capabilities into smaller models.

    \item \textbf{Part~II: Inference-time Scaffolding} We then investigate the effectiveness of techniques to improve performance at inference-time  as a 
    complementary path. Rather than improving the model itself, we build 
    a recursive agent that decomposes hard verification problems into 
    independently solvable subproblems (initially inspired by Hilbert \cite{10.48550/arxiv.2509.22819}), coordinating a set of 
    specialized models to handle each subproblem recursively. Two additions proved 
    key: smart tool calling, which gives the agent 
    access to relevant context such as type signatures, library documentation, 
    and relevant lemmas, and a 
    progress evaluator that monitors whether a line of reasoning is 
    making headway, cutting off token-intensive dead ends. Together, these additions improved both the quality 
    and efficiency of the agent's verification attempts.
\end{enumerate}

The two parts go hand in hand by design. Post-training builds 
models that learn verification techniques by leveraging the strong mathematical and coding skills of current LLM models, while scaffolding extracts the most from these models at test 
time without additional training. Understanding how both perform helps seriously inform the design space for future work on LLM-assisted formal verification.

\section{Contributions}
\label{sec:contributions}

This thesis contributes: (1) an empirical study of post-training for formally
verified code generation, including reinforcement learning from verifier
feedback, ``specification hacking" analysis, dataset filtering, and 
multi-turn verifier-feedback training; (2) an inference-time scaffolding system for Lean verified coding that
combines recursive decomposition, verifier feedback, tool access, and progress
evaluation; and (3) benchmark evaluations on the Vericoding benchmark
\cite{10.48550/arxiv.2509.22908} and the construction of a new Dalek-derived verified-code benchmark, \textbf{Dalek Bench}.

\paragraph{Collaboration and attribution.}
Except where noted here, the work in this thesis was performed by me. David Baek
helped set up reinforcement-learning infrastructure and advised on training
practices such as using the open-source framework \texttt{verl}. Ionel Chiosa helped propose the initial
APPS-derived dataset for post-training, the harder filtered dataset after identifying specification hacking, and helped run the first flat Codex
evaluations on VERINA, BigNum, and VerifCogen. Leo Yao contributed to early
brainstorming around the scaffolding system. All remaining experimental design,
implementation, analysis, and writing were performed by me.
\chapter{Background}

\section{Formal Verification Languages}
\label{sec:2.1}

Formal verification uses mathematical reasoning to prove that a program, protocol, or theorem satisfies a precise specification. Unlike standard testing, which samples executions, formal verification asks a checker to establish that a property holds for all executions admitted by the model. In software verification, the specification may include preconditions, postconditions, loop invariants, data-structure invariants, refinement properties, or memory-safety conditions. In theorem proving, the specification is usually a formal mathematical statement, and the artifact to be checked is a proof.

The central object shared across verification systems is the \emph{proof obligation}: a logical condition that must be established for verification to succeed. Different systems expose proof obligations at different levels of abstraction. SMT-backed verifiers such as Dafny and Verus generate verification conditions and prove many of them automatically using solvers. Interactive theorem provers such as Lean expose proof states directly and require the user to construct proof terms, usually through tactics. These differences matter for LLM-based systems because they determine what the model must generate, what feedback it receives, and how much work can be delegated to the verifier.

This section introduces the verification systems most relevant to the thesis. Dafny represents the auto-active, SMT-backed style of verified programming; Lean represents the interactive theorem-proving style; and Verus and F* provide adjacent points of comparison for recent LLM-based verified-code systems.

\subsection*{Dafny}
\label{sec:2.1.1}

Dafny is a verification-aware programming language with native specification
constructs and a static program verifier for functional correctness
\citep{Leino_2010}. Its specifications include preconditions, postconditions,
loop invariants, frame conditions, termination metrics, and ghost state.
Verified Dafny programs can also be compiled to ordinary programming languages
such as C\#, Java, JavaScript, Go, and Python \citep{dafnyref,gauci_dafny_2014}.
The user writes executable code together with annotations, and Dafny translates
the annotated program into verification conditions that are discharged by
automated reasoning.

\begin{lstlisting}[language=Dafny, caption={A Dafny loop invariant used to prove a postcondition. The method computes \(1+\cdots+n\); the invariant states the closed form for the partial sum after each iteration, giving the verifier the inductive fact needed to establish the final result.}]
method SumN(n: nat) returns (s: nat)
  ensures s == n * (n + 1) / 2
{
  var i := 0;
  s := 0;
  while i < n
    invariant 0 <= i <= n
    invariant s == i * (i + 1) / 2
  {
    i := i + 1;
    s := s + i;
  }
}
\end{lstlisting}

The executable part of this method computes a running sum. The method-level
behavioral specification is the postcondition, while the loop invariants are
intermediate specifications that expose the inductive structure of the
argument. Dafny must prove that the invariants hold initially, are preserved by
each loop iteration, and, together with the negation of the loop guard, imply
the postcondition when the loop exits. In this example, the invariant
\texttt{s == i * (i + 1) / 2} states that the partial sum is correct up to the
current loop index; when the loop terminates, \texttt{i == n}, yielding the
postcondition.

This interaction model is important for LLMs. In Dafny, the model often does
not need to produce a long explicit proof script. Instead, it must generate
compact annotations that expose the right intermediate facts to the solver.
The difficulty is that verifier acceptance alone is not enough: an implementation
may verify against a specification that is too weak to capture the intended
behavior, while a correct implementation and specification may still fail to
verify without sufficiently strong invariants, assertions, or helper lemmas.

\subsection*{Lean}
\label{sec:2.1.2}

Lean is an interactive theorem prover based on dependent type theory, with a
trusted kernel that checks proof terms. In Lean, propositions are represented as
types, and proofs are terms inhabiting those types. Users often construct proofs
through tactics, which transform a proof state until all goals are closed
\citep{de_Moura_2015,moura_lean4_2021}. Lean's large mathematical library,
\texttt{mathlib}, provides reusable definitions and lemmas, making premise
retrieval and library search central to nontrivial automated proof generation.

\begin{lstlisting}[
language=Lean,
caption={A minimal Lean proof using a local hypothesis. The theorem assumes
\texttt{h : a = b} and proves the reversed equality \texttt{b = a} by applying
the symmetry operation \texttt{h.symm}.}
]
theorem symmetry_example (a b : Nat) (h : a = b) : b = a := by
  exact h.symm
\end{lstlisting}

Before the proof is completed, the proof state contains the variables
\texttt{a} and \texttt{b}, the hypothesis \texttt{h : a = b}, and the goal
\texttt{b = a}. The tactic \texttt{exact h.symm} supplies a term whose type
matches the goal, using symmetry of equality. Lean elaborates the tactic proof
into an underlying proof term, which is then checked by the kernel. This
illustrates the interaction surface used by LLM theorem provers: observe a proof
state, generate a tactic or proof term, and use the checker to determine whether
the proof has progressed.

Therefore, Lean places a different burden on LLMs than Dafny. Rather than mostly
generating solver-guiding annotations around executable code, the model must
often search through tactics, select lemmas, manage local context, and construct
explicit proofs. This makes Lean a natural setting for studying proof-state
representation, retrieval, decomposition, and verifier-guided search.

\subsection*{Other Verification Languages}
\label{sec:2.1.3}

Verus and F* provide useful comparison points for Dafny \citep{lattuada_verus_2023, swamy_fstar_2016}. Verus is a Rust-based verifier for proving functional correctness of low-level systems code against user-provided specifications; its specifications include preconditions, postconditions, assertions, and loop invariants, and these specifications are ghost code that does not appear in the compiled executable. F* is a proof-oriented programming language that combines dependent types, SMT-based automation, and tactic-based interactive theorem proving. Both systems appear in later LLM work because they share Dafny's code-plus-specification structure while exposing different proof languages and implementation settings.

Coq and Isabelle provide comparison points for Lean \citep{rocq_prover, Nipkow_2002}. Like Lean, they support machine-checked proofs over large formal libraries, and much early neural theorem-proving work was developed in these ecosystems. Coq emphasizes tactic scripts and proof terms, while Isabelle/HOL combines structured proofs with powerful automation such as Sledgehammer. They are less central to this thesis than Dafny and Lean, but they motivate recurring techniques in the next section: tactic prediction, premise retrieval, proof repair, and proof-assistant interaction.

\section{Automated Formal Methods}
\label{sec:2.2}

This section reviews automated formal methods that use learned models. Section~2.2.1 covers neural theorem proving, where the formal statement is usually fixed and the model searches for a proof. Section~2.2.2 turns to vericoding, where the model must coordinate code, specification, and proof. Section~2.2.3 covers autoformalization and specification inference, which generate the formal statements and annotations that make verifier-grounded training possible.

\subsection{Neural Theorem Proving on Mathematics}
\label{sec:2.2.1}

Interactive theorem provers (ITPs) such as Lean, Isabelle, Coq, and Metamath allow mathematicians to encode definitions, statements, and proofs as machine-checkable artifacts. A proof in this setting is a sequence of tactic (instructional) steps to reach the goal proved state, accepted once the proof is correctly type checked and the goal state is closed; large libraries such as Lean's \texttt{mathlib} accumulate the lemmas and definitions on which many proofs build. \emph{Neural theorem proving} (NTP) applies learned models to this environment in an attempt to automate theorem proving, ranging from early tactic predictors to more recent large language models trained on entire proof corpora and reinforced against verifier reward. This setting is unusually amenable to learned search policies: proof states are explicit, verifier feedback is exact, and mature libraries provide large corpora of reusable lemmas and tactics.

\paragraph{Tactic prediction and search.} The basic NTP template predates transformer models and has two components: a learned tactic policy and a search procedure. TacticToe demonstrated the approach on HOL4 by combining hand-engineered tactic features with depth-bounded search \citep{gauthier_tactictoe_2021}; ASTactic paired the CoqGym corpus with a recurrent decoder that emits tactic invocations as abstract syntax trees \citep{yang_learning_2019}; and Tactician's $k$-nearest-neighbor tactic recommender, combined with CoqHammer's premise (fact) selection and dispatch tool to other automated theorem provers (ATPs), proved 56.7\% of Coq's standard library \citep{blaauwbroek_tactician_2020}. Subsequent work has scaled this template rather than replaced it, preserving the basic policy-plus-search shape.

\paragraph{Transformer policies and self-training.}
The arrival of large transformer models scaled the policy side of the template first. GPT-f trained a decoder-only language model on \texttt{(tactic-state, tactic)} pairs, implementing a ``most-likely"-first proof search, and iteratively retrained on its own verified proofs, pushing Metamath \texttt{set.mm} to roughly 57\% versus the previous 21 \% SOTA performance \citep{polu2020gptf}. \citet{han2021pact} proposed co-training on self-supervised data extracted from kernel-level proof terms alongside the usual tactic prediction objective, raising theorem proving success rate on a held-out mathlib suite from 32\% to 48\%. The gains come from proof terms recording the elaborated proof structure (i.e., resolved implicits, inferred types, and kernel-verified subproofs) that human tactic scripts do not surface. \citet{polu2022curriculum} formalized the self-training half of the loop as \emph{expert iteration} over a \emph{problem curriculum}: auto-generate intermediate problems at calibrated difficulty, attempt them with the current model, replay successes into the supervised pool, and retrain. Search then emerged as a separable axis with HyperTree Proof Search, a Monte Carlo Tree Search style AND/OR-tree algorithm whose policy and critic update online from each newly discovered proof, lifting Metamath from 65.4\% to 82.6\% \citep{lample2022htps} and showed that Monte-Carlo-style search and expert iteration are complementary approaches.

\paragraph{Retrieval and Lean benchmarks.}
Around this supervised core, the modern Lean stack consolidated. LeanDojo extracted programmatic interaction traces and a 98k-theorem \textbf{mathlib} benchmark, and its companion ReProver was the first Lean prover to retrieve premises densely from mathlib before generating a tactic \citep{10.48550/arxiv.2306.15626}; premise retrieval has been a structural primitive of academic Lean provers since. \textbf{miniF2F} established the complementary competition-level evaluation target: 488 problems drawn from the AIME, AMC, and IMO, formalized across Lean, Metamath, and Isabelle, providing a unified cross-system leaderboard against which nearly all subsequent provers are measured \citep{zheng2022minif2f}.Math-focused pretraining also sharpened the base on which downstream fine-tuned models operate: Llemma continued-pretrained 7B and 34B models on Proof-Pile-2, lifting formal-to-formal miniF2F pass rates from 20.5\% to 26.2\% (7B) and 22.2\% to 25.8\% (34B) over the Code Llama baselines without any proof-specific fine-tuning \citep{azerbayev2023llemma}. Process supervision also emerged through approaches such as Lean-STaR, which interleaved informal ``thoughts'' between tactics, ran the resulting tactics through Lean, and fine-tuned only on (thought, tactic) pairs whose proofs verified \citep{lin2024leanstar}, which reappears at scale in Kimina-Prover and in the current generation of LLM-based provers.

\paragraph{Verifier-reward training and test-time search.}
The DeepSeek-Prover line brings these ingredients together in a modern NTP pipeline: synthetic formal data, verifier-filtered self-training, reinforcement learning from proof-assistant feedback, and test-time search. DeepSeek-Prover V1 trained a 7B prover on synthetic Lean proofs; V1.5 added GRPO-style verifier rewards and an MCTS-like search procedure; and V2 further emphasized subgoal decomposition and large-scale reasoning traces. Taken together, these systems show that recent gains in NTP come less from next-token tactic prediction alone than from the interaction between data generation, verifier feedback, and search \citep{xin_deepseek-prover_2024,xin_deepseek-prover-v15_2024,ren_deepseek-prover-v2_2025}.

Recent provers continue to explore different points in this design space. Goedel-Prover-V2 combines scaffolded data synthesis, verifier-guided self-correction, and model averaging, achieving strong miniF2F performance with relatively small 8B and 32B models \citep{lin_goedel-prover_2025,lin_goedel-prover-v2_2025}. Kimina-Prover applies large-scale RL to a Qwen2.5-72B base model using Lean-STaR-style interleaved reasoning traces \citep{wang2025kimina}. Seed-Prover emphasizes lemma-style whole-proof reasoning and test-time compute allocation by problem difficulty, while AlphaProof uses a different AlphaZero-style recipe: autoformalize informal problems into Lean, train through verified proof search, and update the policy online at competition time \citep{chen2025seedprover,hubert2025alphaproof}. Across these systems, the common lesson is that frontier NTP systems are increasingly built around verifier-grounded data generation, decomposition, and search-time scaling rather than static supervised learning alone. Agentic scaffolds built on top of these specialist provers, including DSP, COPRA, LEGO-Prover, Apollo, and Hilbert, are discussed later in \textbf{Part II}.

\paragraph{Limits of math-only NTP.}
Two empirical findings do slightly undermine these headline results. First, autoformalization errors can dominate end-to-end performance: \citet{ospanov2025minif2fv2} show that combining a high-accuracy autoformalizer with a strong prover can still yield much lower full-pipeline accuracy when the formalized statement drifts from the informal problem. Second, benchmark transfer remains limited: \citet{poiroux2025rlmeval} evaluate provers on research-level Lean Blueprint theorems and find substantially lower success than on curated competition benchmarks. Math NTP is therefore a powerful demonstration of verifier-reward learning, but it is also a comparatively clean setting. The statement is fixed, the artifact being generated is usually only the proof, mature libraries provide retrievable lemmas, and the proof assistant supplies exact feedback on candidate proof steps. The vericoding setting in Section~\ref{sec:2.2.2} relaxes these assumptions: the model must synthesize code and specifications alongside the proof, and verifier success can be confounded by weak specifications or incorrect implementations.

\subsection{LLMs for Vericoding}
\label{sec:2.2.2}

Where ordinary code generation asks a model to produce executable code, and theorem proving often asks it to complete a proof for a fixed formal statement, \emph{vericoding} asks the model to align three coupled artifacts: code, specification, and proof. The central challenge is that each artifact can fail independently: code may compile but violate intent, a specification may verify while being vacuous, and a proof may fail despite nearly correct code and specifications.

Recent benchmarks show that this alignment problem remains difficult even for frontier models. On VERINA, a Lean~4 benchmark of 189 verified-programming tasks, OpenAI's o3 reaches 72.6\% code correctness, 52.3\% specification correctness, and only 4.9\% end-to-end proof success \cite{verina2025}. CLEVER reports the same shape: frontier models routinely produce code that compiles, yet none achieves end-to-end Lean verification at meaningful rates \cite{clever2025}. The 12{,}504-specification cross-language Vericoding benchmark of sharpens the asymmetry across systems: off-the-shelf LLMs verify 82\% of Dafny tasks, 44\% of Verus/Rust tasks, and 27\% of Lean tasks \citep{10.48550/arxiv.2509.22908}. These results suggest that the main obstacle is not merely syntactic familiarity with a verification language, but the proof burden imposed by each environment and the feedback structure available to the model.

Existing LLM approaches to vericoding can be organized along three axes: the feedback signal, the artifact being revised, and the supervision regime. Feedback ranges from raw verifier errors, to retrieved examples, to synthesized counterexamples, to dense proof-progress or verifier-reward signals. The revised artifact may be the implementation, the formal specification, proof annotations, helper assertions, tactic scripts, or an explicit decomposition of the proof obligation. Finally, supervision ranges from prompt scaffolding and retrieval, to fine-tuning on verifier-filtered data, to reinforcement learning against proof-checker feedback. This section follows those axes rather than reintroducing the verification systems themselves.

\paragraph{Prompting, retrieval, and scaffolded generation.}
Early work on Dafny established the basic prompt-and-repair shape of LLM-assisted verified programming. \citet{10.48550/arxiv.2406.08467} introduced DafnyBench, a 750-program corpus of structured problems. With retry-on-error feedback, the strongest baseline reached 68\% pass rate but degraded sharply with program length. \citet{misu2024dafny} studied method synthesis over 178 MBPP problems translated into Dafny, establishing a retrieval-augmented and chain-of-thought prompting taxonomy. Laurel \cite{mugnier2025laurel} pushed this scaffolding further by parsing verifier errors to insert assertion placeholders at failing locations and retrieving examples using a proof-similarity metric, generating 56.6\% of missing assertions on DafnyGym. Clover \cite{sun2024clover} added a tri-consistency check between code, formal annotations, and natural-language docstrings, showing that verified-code benchmarks can be gamed by under-specification unless specification strength is checked explicitly.

Retrieval has also become central outside Dafny. In F*, \citet{chakraborty2025fstar} curated a large real-code corpus of 940{,}000 lines and 54{,}000 top-level definitions from production Windows, Linux, Python, and Firefox proofs, then used retrieval over in-corpus proofs to condition small open models such as Phi-2 and StarCoder. PoPilot \cite{zhang2025popilot} extends this F* direction by combining synthetic basic-problem augmentation with project-level repair traces, fine-tuning a repair-specialized model on edit sequences developers produce when fixing broken proofs. \citet{jain2025whats} instead studies source-code telemetry from F* and Verus developers, using process data to derive prompt and interaction policies for proof agents. In Coq, Rango \cite{thompson2025rango} retrieves both relevant premises and similar in-project proofs during proof search, reaching 32.0\% on CoqStoq and improving substantially over prior baselines. Across these systems, retrieval gives the model access to local proof idioms and project-specific lemmas that are unlikely to be recoverable from pretraining alone.

\paragraph{Verifier-guided repair and counterexample feedback.}
A second line of work treats the verifier as an interactive critic rather than a final judge. AutoVerus \cite{yang2025autoverus} introduced a three-phase workflow for Verus: preliminary proof generation, tip-guided refinement, and error-message-driven debugging. SAFE \cite{chen2025safe} extends this template with self-evolving supervised fine-tuning, where verifier-filtered outputs become training data and incorrect attempts are repurposed as self-debugging traces. VeriStruct \cite{sun2025veristruct} scales agentic refinement from individual functions to data-structure modules by planning abstractions, type invariants, specifications, and proof code before applying syntax-aware repair, verifying 128 of 129 functions and 10 of 11 modules. ExVerus \cite{yang2026exverus} replaces raw verifier-error repair with synthesized source-level SMT queries whose satisfying models act as concrete counterexamples, helping the model add invariants that rule out real failing executions rather than merely reacting to low-level solver output.

Similar repair patterns appear in theorem-proving settings. PALM \cite{lu2024palm} observes that LLMs often produce plausible high-level Coq proof structure but fail on low-level details, motivating a generate-then-repair pipeline that drafts whole proofs and then discharges failing subgoals using Sledgehammer, CoqHammer, and deterministic backtracking. These systems suggest that verifier feedback is most useful when it is transformed into a model-readable intermediate signal: a placeholder location, a counterexample, a retrieved proof, a smaller subgoal, or a structured repair trace.

\paragraph{Fine-tuning and verifier-filtered supervision.}
Prompting and retrieval improve inference-time behavior, but they do not by themselves adapt the model distribution to verification tasks. Several systems therefore build verifier-filtered training loops. PoPilot \cite{zhang2025popilot} augments F* supervision with synthetic basic problems and project-level repair traces. SAFE \cite{chen2025safe} applies a similar idea in Verus by recycling successful verifier-filtered outputs into later supervised fine-tuning rounds and using failed attempts as self-debugging data. For Dafny, DafnyBench \cite{10.48550/arxiv.2406.08467}, DafnyGym \cite{mugnier2025laurel}, and related hint-removal datasets provide natural supervision targets: missing assertions, loop invariants, proof hints, and specification annotations. This line of work is especially relevant to Dafny because the model can often succeed by generating compact annotations that guide the SMT backend, rather than producing a full proof script.

\paragraph{RL and proof decomposition with verifier rewards.}
More recent work moves from supervised repair traces toward reinforcement learning with verifier-derived rewards. Goedel-Code-Prover \cite{li2026goedelcodeprover} is the clearest example in Lean verified programming: it recursively decomposes a top-level verification theorem into subgoals, scores candidate decompositions with a progress metric, and trains an 8B model first by supervised fine-tuning on decomposition-and-completion trajectories and then by GRPO with online Lean~4 verifier rewards. Its results across VERINA, CLEVER, and AlgoVeri suggest that decomposition and dense verifier feedback can make proof search more learnable than direct whole-proof generation.

This direction is important because it treats the verifier not only as a filter, but as a reward source. In Dafny and Verus, analogous rewards may come from partial verification, fewer remaining obligations, successful invariant preservation, or reduced counterexample space. In Lean, rewards may come from proof-state progress, subgoal closure, or successful elaboration of intermediate tactic steps. The common theme is that formal verification supplies unusually clean correctness signals, but only if the harness can expose them at the right granularity.

\paragraph{Verifier interfaces and proof-search infrastructure.}
A final line of work studies the infrastructure needed to expose proof assistants as usable environments for LLM training and search. LeanDojo provides an open Lean environment with data extraction, programmatic interaction, premise annotations, benchmarks, and ReProver, addressing the reproducibility barriers created by private datasets and large closed training pipelines \cite{10.48550/arxiv.2306.15626}. Pantograph similarly frames Lean interaction as a machine-to-machine interface, exposing proof execution, environment access, data extraction, and search hooks for methods such as Monte Carlo Tree Search \cite{10.48550/arxiv.2410.16429}. These systems suggest that LLM-based formal verification is not only a modeling problem: the verifier must be wrapped in an interface that exposes proof states, local context, retrievable premises, verifier errors, and partial progress at the right granularity. This motivates the Lean harness component of this thesis.

\paragraph{Open failure modes.}
Across these approaches, three failure modes recur. First, models can reward-hack the verifier by producing weak or vacuous specifications that make verification easy without capturing the intended behavior, a problem made explicit by Clover's tri-consistency framing \cite{sun2024clover}. Second, single-shot generation remains brittle: published ablations across DafnyBench, Laurel, AutoVerus, SAFE, PALM, and Goedel-Code-Prover show that iterative refinement, repair, retrieval, or decomposition outperforms one-pass generation, especially as tasks become longer or require multiple annotations \cite{10.48550/arxiv.2406.08467,mugnier2025laurel,yang2025autoverus,chen2025safe,lu2024palm,li2026goedelcodeprover}. Third, multi-artifact repair compounds errors: a system that can fill one missing assertion may still fail when code, specification, and proof must be revised jointly. These failures motivate the two directions studied in this thesis: improving LLM performance on Dafny through verifier-aware training, and building Lean harnesses that expose proof-state feedback, decomposition, and verifier rewards at the right granularity.

\subsection{Autoformalization and Spec Inference}
\label{sec:2.2.3}
Autoformalization translates informal intent into a formal artifact that a verifier can check. In mathematics, this means translating natural-language theorem statements or proof sketches into Lean, Isabelle, or related proof-assistant languages. In software verification, it means inferring preconditions, postconditions, loop invariants, helper assertions, or contracts from docstrings, tests, natural-language specifications, or code. Both settings follow the same propose-and-check loop: an LLM proposes a formal candidate, and a sound checker filters or scores it. The difference is semantic risk: mathematical autoformalization must preserve the meaning of an informal theorem, while software spec inference must avoid weak or vacuous specifications that verify without capturing intended program behavior.

Two threads are especially relevant to this thesis. The first is \emph{mathematical autoformalization}, which translates natural-language theorem statements and proofs into formal proof-assistant languages such as Lean or Isabelle. The second is \emph{software specification inference}, which translates natural-language descriptions, tests, or code into preconditions, postconditions, invariants, and proof helpers for systems such as Dafny, Verus, Frama-C, or Boogie. Both threads follow the same high-level loop: an LLM proposes a formal candidate, and a sound checker filters or scores it. Their main difference is the target of formalization. Math autoformalization usually formalizes the statement to be proved; software spec inference must formalize the intended behavior of executable code, where weak or vacuous specifications can pass the verifier while failing to capture the intended program semantics.

\subsubsection*{Math autoformalization}

Early work established that LLMs could perform nontrivial mathematical autoformalization at all. \citet{wu2022autoformalization} showed that few-shot-prompted models could translate competition-style problems into Isabelle/HOL, and that autoformalized statements could improve a downstream prover. This work introduced the core template reused throughout verifier-grounded learning: generate formal candidates from informal artifacts, then use a checker to convert noisy model outputs into training or evaluation signal.

Subsequent work expanded the unit of formalization from statements to proof structure. Draft-Sketch-Prove \cite{jiang2023dsp} first drafts an informal proof, then converts it into an Isabelle/Isar skeleton with \texttt{sorry} placeholders, and finally uses Sledgehammer to close the holes. The result suggests that LLMs are often better at producing high-level proof plans than complete low-level formal proofs. \citet{zhou2024dontrustverify} invert the role of autoformalization: instead of using it only to translate problems, they use it to verify natural-language solutions by autoformalizing candidate reasoning into Isabelle, checking consistency with the formalized problem, and aggregating only the surviving candidates.

At larger scale, AlphaProof uses autoformalization as a data-generation engine: informal mathematical problems are translated into Lean~4 and filtered to produce large formal training corpora for an AlphaZero-style prover \cite{hubert2025alphaproof}. Open datasets such as Lean Workbook \cite{ying2024leanworkbook} and ProofNet \cite{azerbayev2023proofnet} provide smaller-scale anchors for this pipeline. The important point for this thesis is not only that autoformalization can produce formal data, but that checker-filtered formal data can become a training substrate for downstream proving.

However, headline autoformalization results can overstate end-to-end reliability. \citet{ospanov2025minif2fv2} audit miniF2F-Lean and find systematic discrepancies between informal and formal statements, showing that syntactic validity and prover success are not sufficient: the formal statement must also preserve the intended informal meaning. Distribution shift sharpens this concern. RLMEval evaluates proof autoformalization and neural theorem proving on research-level Lean Blueprint theorems, where models perform far worse than on curated competition benchmarks \cite{poiroux2025rlmeval}. Semantic equivalence is itself difficult to judge: ReForm's ConsistencyCheck reports substantial semantic-error rates even among human experts, and improves autoformalization by explicitly training against consistency checks \cite{chen2025reform}. Tool-feedback systems such as Autoformalizer-with-Tool-Feedback \cite{gu2025proofoptimizer} have made syntactic invalidity less central; the harder residual failures are semantic drift, vacuous or over-strong statements, and hallucinated helper lemmas.

\subsubsection*{Software spec inference}

The same propose-and-check pattern transfers to software, but the formal artifact changes. Instead of translating a theorem statement, the model must infer contracts, invariants, assertions, and proof helpers that capture the intended behavior of code. The checker may be Boogie/Z3, Frama-C with its WP plugin, the Dafny verifier, or another deductive verification backend. This setting is closer to the thesis target because the formal artifact is not independent of the implementation: changing the specification, annotation, or helper proof can change what the verifier accepts.

Loopy \cite{kamath2023loopy} instantiates this pattern for loop invariants in C/Boogie programs, with LLM-generated candidates accepted or rejected by Boogie and Z3. SpecGen \cite{ma2025specgen} extends specification inference to Java by generating JML preconditions, postconditions, and class invariants, then applying mutation operators such as precondition weakening to reduce vacuity. AutoSpec \cite{wen2024autospec} targets C programs using static-analysis-driven decomposition and iterative Frama-C feedback, including on a real-world X.509 parser. These systems mirror math autoformalization methodologically, but their failure modes are more software-specific: inferred specifications may be too weak, invariants may fail to support later proof steps, and local annotations may interact in non-obvious ways.

Spec inference for Dafny is especially relevant to this thesis because Dafny sits between ordinary programming and interactive theorem proving: the user writes executable code plus annotations, while the SMT-backed verifier discharges many proof obligations automatically. \citet{faria2026dafny} report a multi-model pipeline that generates preconditions, postconditions, loop invariants, auxiliary predicates, and proof helpers from code annotated with natural-language specifications and tests. Their analysis identifies proof-helper annotations as a disproportionate source of remaining difficulty. DAISY \cite{silva2025daisy} makes the same bottleneck more explicit: a hybrid LLM and error-message heuristic works substantially better when only one helper assertion is missing than when multiple helpers must be inferred simultaneously.

This connects directly to the failure modes identified in vericoding. Systems such as Clover \cite{sun2024clover} and SAFE \cite{chen2025safe} treat natural-language specifications, formal annotations, code, and verifier feedback as parts of a larger consistency loop. Their role here is not just to generate code, but to guard against the central risk of specification inference: a candidate may verify for the wrong reason. This is the software analogue of semantic drift in math autoformalization, but with an additional reward-hacking failure mode: a model can make verification easier by weakening the specification rather than by producing a better proof.

The two threads therefore sit at different maturity levels. Math autoformalization has canonical benchmarks such as miniF2F, ProofNet, and ConsistencyCheck, a clean checker-filtered formulation, and industrial-scale demonstrations. Its main open problems are semantic-equivalence judgment and distribution shift to research-level mathematics. Software spec inference is more fragmented: Loopy, SpecGen, AutoSpec, Dafny-specific pipelines, and DAISY evaluate on different languages, checkers, and curated benchmarks, making cross-system comparison difficult. For this thesis, the central lesson is that verifier feedback is essential but too coarse on its own. It can certify that a candidate specification or helper annotation checks, but it does not always reveal whether the specification captures the intended behavior or which missing helper caused a multi-step proof to fail.

\chapter{Benchmarks}
\label{ch:benchmarks}

This chapter describes the evaluation suite used in the thesis. The thesis uses two complementary benchmarks. 
The Vericoding Benchmark evaluates standardized formally verified program 
synthesis tasks across Dafny, Verus/Rust, and Lean. Dalek Bench, constructed 
in this thesis from the \texttt{curve25519-dalek-lean-verify} repository, 
evaluates proof completion and repair in a repository-grounded Lean setting.

The distinction between these benchmarks is important. The Vericoding 
Benchmark measures broad cross-language capability on self-contained formal 
specification tasks, while Dalek Bench measures whether a model can operate 
inside an existing Lean verification project, where success depends on 
repository context, imported definitions, domain-specific lemmas, and the 
ability to repair incomplete proofs without changing the trusted 
specification.

\section*{Benchmark Overview}

Table~\ref{tab:benchmark-overview} groups related benchmarks by evaluation 
setting. Math-focused benchmarks test formal theorem proving and 
autoformalization. Dafny benchmarks emphasize localized verifier repair and 
annotation synthesis. Vericoding benchmarks move toward end-to-end generation 
of verified implementations and proofs. Dalek Bench extends this setting to 
repository-grounded Lean proof repair, where success depends on existing 
imports, definitions, lemmas, and project structure.
\begin{table}[t]
\centering

\begingroup
\scriptsize
\setlength{\tabcolsep}{5pt}
\renewcommand{\arraystretch}{1.10}

\begin{tabularx}{\textwidth}{
>{\hsize=.95\hsize\raggedright\arraybackslash}X
>{\hsize=.95\hsize\raggedright\arraybackslash}X
>{\hsize=1.05\hsize\raggedright\arraybackslash}X
>{\hsize=1.05\hsize\raggedright\arraybackslash}X
}
\toprule
\textbf{Benchmark} 
& \textbf{Language(s)} 
& \textbf{Task} 
& \textbf{Thesis Role} \\
\midrule[0.7pt]
\addlinespace[0.25em]

\rowcolor{gray!12}
\multicolumn{4}{l}{\textbf{\textit{Math proving and autoformalization}}} \\
\addlinespace[0.15em]

miniF2F 
& Lean, Isabelle, Metamath, HOL Light 
& Formal math proofs 
& Standard NTP baseline \\

ProofNet 
& Lean 
& Informal-to-formal math 
& Autoformalization reference \\

PutnamBench 
& Lean, Isabelle, Coq 
& Hard math proofs 
& Reasoning stress test \\

\addlinespace[0.35em]
\rowcolor{gray!12}
\multicolumn{4}{l}{\textbf{\textit{Verified-program repair}}} \\
\addlinespace[0.15em]

DafnyBench 
& Dafny 
& Annotation completion 
& Dafny repair baseline \\

DafnyGym 
& Dafny 
& Assertion synthesis 
& Local repair baseline \\

\addlinespace[0.35em]
\rowcolor{gray!12}
\multicolumn{4}{l}{\textbf{\textit{End-to-end vericoding}}} \\
\addlinespace[0.15em]

VERINA 
& Lean 
& Code/spec/proof generation 
& Lean vericoding reference \\

CLEVER 
& Lean 
& Verified code synthesis 
& Curated Lean benchmark \\

\textbf{Vericoding Benchmark} 
& Dafny, Verus/Rust, Lean 
& Code/proof synthesis 
& \textbf{Main broad benchmark} \\

\addlinespace[0.35em]
\rowcolor{gray!12}
\multicolumn{4}{l}{\textbf{\textit{Repository-grounded proof repair}}} \\
\addlinespace[0.15em]

\textbf{Dalek Bench} 
& Lean 
& Proof completion/repair 
& \textbf{Main repo-context benchmark} \\

\bottomrule
\end{tabularx}

\endgroup

\caption[Review of evaluation benchmarks grouped by setting and thesis role]{
Selected benchmarks grouped by evaluation setting. Math-focused benchmarks 
test formal theorem proving and autoformalization; Dafny benchmarks emphasize 
localized verifier repair and annotation synthesis; vericoding benchmarks 
evaluate end-to-end generation of verified code and proofs. The two thesis 
benchmarks are shown in bold: the Vericoding Benchmark provides broad 
cross-language coverage, while Dalek Bench evaluates repository-grounded Lean 
proof completion and repair.
 \citep{zheng2022minif2f, azerbayev2023proofnet, 10.48550/arxiv.2407.11214, 10.48550/arxiv.2406.08467, mugnier_laurel_2025, verina2025, clever2025, 10.48550/arxiv.2509.22908, noauthor_beneficial-ai-foundationcurve25519-dalek-lean-verify_2026}}
\label{tab:benchmark-overview}
\end{table}

\section{Vericoding Benchmark}
\label{sec:2.3.2}

We use \emph{vericoding} to refer to the generation and repair of
machine-checkable verified software artifacts, including formal specifications,
executable definitions or implementations, and proofs connecting the
implementation to the specification. In the Vericoding Benchmark \citep{10.48550/arxiv.2509.22908}(See Table~\ref{tab:vericoding-composition}, the central
task is more constrained: given an initial specification specification and surrounding context,
the model must generate the missing implementation, broken specifications, helper lemmas,
or proof artifacts required for the target verifier to accept the completed
file.

The benchmark was designed to bridge code generation and neural theorem
proving. Ordinary coding benchmarks such as HumanEval or MBPP evaluate
executable synthesis, but correctness is approximated by finite test suites.
Theorem-proving benchmarks provide machine-checkable success criteria, but
typically target mathematical statements rather than verified programs.
Vericoding combines these requirements by testing whether a model can align
implementation, specification, and proof under a formal checker.

The benchmark also reflects a practical feature of formal verification:
specifications are not always complete, consistent, or immediately compilable.
Although the main experiments evaluate verifier-accepted completions from
formal task contexts, the benchmark construction explicitly treats
specification repair as part of the broader verification workflow. This matters
because real verification rarely consists only of filling a proof hole under a
perfect specification; it often requires identifying whether the specification,
implementation, auxiliary definitions, or proof structure is responsible for a
verification failure.

The benchmark was constructed by curating source tasks from existing
verification benchmarks, programming benchmarks, and mathematical library
documentation, then normalizing them into a common vericoding format. For
benchmark-derived tasks, implementations, helper lemmas, and proofs were
removed and replaced with holes. For programming and documentation sources,
formal specifications were first produced by autoformalization. Cross-language
coverage was expanded with an LLM-based specification translator: given a
source-language specification and a target verification language, the translator
generated a candidate target-language specification, checked it with the target
verifier, and iteratively repaired it using verifier feedback. The resulting
files were organized with explicit \texttt{vc-*} section tags, such as
\texttt{vc-preamble}, \texttt{vc-spec}, \texttt{vc-code}, \texttt{vc-helpers},
and \texttt{vc-postamble}, and were then parsed into benchmark metadata,
compiled, and quality-checked.

The benchmark remains broad rather than repository-specific. Its small and
self-contained tasks support controlled comparison, but they do not fully
capture the long-context, library-dependent reasoning required in real verified
software.

\begin{table}[h]
\centering
\small
\setlength{\tabcolsep}{5pt}
\renewcommand{\arraystretch}{1.08}
\begin{tabular}{l c r r r r}
\toprule
\textbf{Source} & \textbf{Ref} & \textbf{Dafny} & \textbf{Verus/Rust} & \textbf{Lean} & \textbf{Total} \\
\midrule
APPS (Test)       & A & 677 & 536 & 676 & 1889 \\
DafnyBench        & D & 443 & 440 & 440 & 1323 \\
NumPyTriple       & T & 603 & 581 & 603 & 1787 \\
VerifiedCogen     & J & 172 & 172 & 172 & 516 \\
VERINA            & V & 157 & 156 & 189 & 502 \\
BigNum            & B & 62  & 62  & 62  & 186 \\
NumPySimple       & S & 58  & 58  & 59  & 175 \\
HumanEval         & H & 162 & 161 & \(161^{\dagger}\) & 484 \\
FVAPPS            & F & --  & --  & 4006 & 4006 \\
\midrule
\textbf{Total}    &   & \textbf{2334} & \textbf{2166} & \textbf{6368} & \textbf{10868} \\
\bottomrule
\end{tabular}
\caption[Vericoding Benchmark task counts by source and language]{
Composition of the Vericoding Benchmark evaluation subset by source dataset
and target language. Counts report the retained tasks used in experiments after
compilation, formatting, and quality checks. The full generated release contains
12,504 formal specifications; the retained experimental subset contains 10,868
tasks. \(\dagger\) The Lean HumanEval row corresponds to the CLEVER benchmark
subset.
}
\label{tab:vericoding-composition}
\end{table}

\subsection*{DafnyBench Re-evaluation and Model Progress}
\label{sec:2.3.2-dafnybench-progress}

A contribution of this thesis was the re-evaluation of DafnyBench under newer
frontier models as part of the Vericoding Benchmark study \citep{10.48550/arxiv.2509.22908}. DafnyBench is a
useful longitudinal baseline within Vericoding because it evaluates localized Dafny verification:
models are given Dafny programs with missing proof hints, assertions, or
invariants, and must restore enough annotations for the verifier to accept the
program while preserving the intended specification \citep{10.48550/arxiv.2406.08467}.

The re-evaluation was run as an API-based experiment over frontier models
available at the time of evaluation, accessed through OpenRouter as a common
model interface. This allowed the same prompting, parsing, verifier invocation,
and repair loop to be applied across models from different providers. The
experiment retained the original DafnyBench prompts and evaluation criteria, but
updated the infrastructure for newer reasoning-heavy models. Thus, the
comparison only measures model progress under
rather than gains from a new benchmark format or task construction.

The original
DafnyBench study reported a best success rate of approximately \(68\%\)
\citep{10.48550/arxiv.2406.08467}. Under the Vericoding Benchmark
re-evaluation, the strongest single model reached approximately \(89\%\), while
the union over evaluated models reached approximately \(96\%\)
\citep{10.48550/arxiv.2509.22908}. These results indicate rapid progress on
localized SMT-guided verification repair.

The result is important but also limited. It shows that frontier models have
substantially improved at localized Dafny proof repair, especially the
generation of assertions, invariants, and auxiliary proof hints. However,
DafnyBench does not fully test cross-language vericoding, specification repair,
or repository-grounded proof construction. It therefore motivates grounding in the broader
Vericoding Benchmark.

\subsection*{Initial LLM Performance and Subset Selection}
\label{sec:2.3.2-initial-performance}

The initial Vericoding Benchmark results provide an empirical basis for
selecting subsets for further study (See Table~\ref{tab:vericoding-initial-performance}. In the benchmark evaluation, off-the-shelf
frontier LLMs were prompted to fill tagged code and proof holes, with verifier
errors returned for a fixed number of repair iterations. At the language level,
model-union success was highest on Dafny with \(82.2\%\), followed by Verus/Rust and Lean with \(44.3\%\) and \(26.8\%\) respectively
\citep{10.48550/arxiv.2509.22908}. This difference stems from the high degree of automation inherent in SMT-backed verifiers, in contrast to interactive theorem provers such as Lean that rely on explicit, user-guided tactic sequences to construct formal proofs. The gap supports using Dafny as a tractable
post-training setting for smaller models, where SMT-backed verifier feedback
provides an automatic training signal, while using Lean subsets to study
inference-time scaffolding for harder explicit proof construction.

\begin{table}[t]
\centering
\scriptsize
\setlength{\tabcolsep}{3pt}
\renewcommand{\arraystretch}{1.12}
\begin{tabular*}{\textwidth}{
@{\extracolsep{\fill}}
>{\raggedright\arraybackslash}p{2.6cm}
>{\centering\arraybackslash}p{1.3cm}
>{\centering\arraybackslash}p{1.1cm}
>{\raggedright\arraybackslash}p{2.6cm}
>{\centering\arraybackslash}p{1.5cm}
>{\centering\arraybackslash}p{1.5cm}
@{}}
\toprule
\textbf{Setting} &
\textbf{Lang.} &
\textbf{Tasks} &
\textbf{Best Model} &
\begin{tabular}[c]{@{}c@{}}\textbf{Best}\\\textbf{Success}\end{tabular} &
\begin{tabular}[c]{@{}c@{}}\textbf{Model}\\\textbf{Union}\end{tabular} \\
\midrule[0.7pt]

\rowcolor{gray!12}
\multicolumn{6}{@{}l}{\textbf{\textit{Language-level baselines}}} \\
\addlinespace[0.1em]

Overall 
& Dafny 
& 2161 
& Claude Opus 4.1 
& \(67.5\%\) 
& \(82.2\%\) \\

Overall 
& Verus/Rust 
& 2166 
& GPT-5 
& \(30.9\%\) 
& \(44.3\%\) \\

Overall 
& Lean 
& 2361 
& GPT-5 
& \(17.9\%\) 
& \(26.8\%\) \\

\addlinespace[0.35em]
\rowcolor{gray!12}
\multicolumn{6}{@{}l}{\textbf{\textit{Dafny subsets}}} \\
\addlinespace[0.1em]

APPStest 
& Dafny 
& 677 
& GPT-5-mini 
& \(71.6\%\) 
& \(83.0\%\) \\

HumanEval 
& Dafny 
& 162 
& Claude Sonnet 4 
& \(72.8\%\) 
& \(93.2\%\) \\

VerifiedCogen 
& Dafny 
& 172 
& Claude Opus 4.1 
& \(90.7\%\) 
& \(95.9\%\) \\

\addlinespace[0.35em]
\rowcolor{gray!12}
\multicolumn{6}{@{}l}{\textbf{\textit{Lean subsets}}} \\
\addlinespace[0.1em]

VERINA 
& Lean 
& 189 
& Claude Opus 4.1 
& \(15.3\%\) 
& \(25.4\%\) \\

VerifiedCogen 
& Lean 
& 172 
& GPT-5 
& \(34.9\%\) 
& \(44.2\%\) \\

BigNum 
& Lean 
& 62 
& GPT-5 
& \(12.9\%\) 
& \(12.9\%\) \\

\bottomrule
\end{tabular*}
\caption[Initial Vericoding Benchmark performance]{
Initial Vericoding Benchmark performance used to guide subset selection
\citep{10.48550/arxiv.2509.22908}. ``Best Success'' reports the strongest
single-model result for each setting, while ``Model Union'' reports the
fraction of tasks solved by at least one evaluated model. The results show a
large gap between Dafny and Lean performance, motivating Dafny as a tractable
post-training setting for smaller models and Lean as a target for
inference-time proof scaffolding.
}
\label{tab:vericoding-initial-performance}
\end{table}

These results motivate different experimental roles for Dafny and Lean. For
Dafny, APPStest is a useful post-training target because it has the largest
Dafny subset size among the selected candidates and retains nontrivial headroom:
the model union solves \(83.0\%\), compared with \(93.2\%\) on HumanEval and
\(95.9\%\) on Dafny VerifiedCogen. Thus APPStest is high-performing enough to
provide a dense verifier-feedback signal, but not so saturated that improvements
become difficult to measure.

The Lean subsets are chosen to make scaffold iteration measurable. A useful
scaffold baseline should leave room for improvement over one-shot generation
while still producing enough partially successful attempts and verifier
diagnostics to guide development. VERINA provides comparability with prior Lean
vericoding work, since it consists of curated verified algorithmic and
data-structure tasks. Lean VerifiedCogen provides a more tractable improvement
target: its \(44.2\%\) model-union success rate shows that current models can
solve many instances, while still leaving substantial headroom for gains. BigNum provides a
harder stress test, with only \(12.9\%\) model-union success and a focus on
arithmetic-heavy, cryptography-adjacent reasoning. It is therefore useful for
probing whether scaffolds help on longer and more brittle proof obligations, but
should not be treated as a standalone measure of general Lean performance.

\section{Dalek Bench}
\label{sec:2.3.3}

Dalek Bench is a thesis-constructed Lean benchmark derived from the
\texttt{curve25519-dalek-lean-verify} repository. The upstream project verifies
parts of the Rust \texttt{curve25519-dalek} cryptography library by translating
Rust implementations into Lean with Aeneas and proving Lean specifications about
the translated functions. Dalek Bench turns this repository-scale verification
project into a collection of proof-completion and repair tasks for LLM-based
proof assistants.

Unlike the Vericoding Benchmark, which aggregates many relatively small tasks
across languages, Dalek Bench is built from a single realistic verification
codebase. The benchmark therefore tests whether a model can operate against
existing Lean definitions, generated names, local specifications, imported
lemmas, and project proof conventions. This makes it a natural evaluation
setting for the inference-time scaffolds studied in Part~II.

\subsection*{Source Repository and Extraction Target}

The benchmark starts from the \texttt{curve25519-dalek-lean-verify}
repository. The verified Rust baseline is the \texttt{curve25519-dalek}
\texttt{4.2.0} release, and the Rust code is translated into Lean using Aeneas.
The extracted Lean project contains executable definitions, generated from the
Rust implementation, together with Lean specification modules that state and
prove correctness properties of those definitions.

The extraction is intentionally not a full blind translation of the Rust crate.
It targets the core verification surface used by the project: backend support,
field arithmetic, scalar arithmetic, Edwards-curve operations, Montgomery-form
operations, Ristretto encodings, and related constants. Some Rust features that
were outside the reliable Aeneas fragment, such as iterator-heavy or
dynamically indexed paths, are excluded or rewritten before extraction. As a
result, Dalek Bench focuses on the core arithmetic and curve-operation proofs
that are present in the Lean verification project.

\subsection*{Helper-Scaffolded Benchmark Construction}

The initial Dalek Bench version used in this thesis is a
\emph{helper-scaffolded} worksheet benchmark. Each task is centered on one
top-level Rust-function correctness theorem. Instead of presenting only the
main theorem, the benchmark also exposes selected proof-only supporting
specification theorems as helper goals. This makes the proof decomposition
explicit while still requiring all exposed helper goals to be proved by the solver.

The construction proceeds in four stages. First, the extraction pass scans the
Lean verification files and records source-level declarations. Second, a
compiler metadata pass stores declaration metadata and dependency information,
including which dependencies occur in theorem statements and which occur only
inside proofs. Third, a selection pass filters declarations into an auditable
theorem pool, keeping substantive theorem and lemma declarations from the
Curve25519-Dalek Lean project while discarding auxiliary declarations and
declarations that depend on upstream \texttt{sorry}. Finally, an emission pass creates one small Lake project per benchmark task.
Lake is Lean's build system, so each emitted project packages the relevant
Lean files, imports, toolchain, and dependencies needed to check the task. This
lets the grader evaluate a submitted \texttt{Task.lean} file in a reproducible
Lean environment matching the original repository context.

\subsection*{Task Format}

Each emitted task is a self-contained Lake project containing a vendored slice
of the original verification repository, a visible \texttt{Task.lean} file, a
reference \texttt{solution.lean} file for grading and analysis, and a
\texttt{meta.yaml} file with target and dependency metadata. The solver edits
the worksheet-style \texttt{Task.lean} file.

The worksheet follows the same high-level sectioning convention as the
Vericoding Benchmark:
$\texttt{vc-preamble},\quad
\texttt{vc-helpers},\quad
\texttt{vc-definitions},\quad
\texttt{vc-theorems}.
$
The preamble imports the relevant specification module. The helper section
contains copied helper theorem statements whose proof bodies have been removed and replaced with \texttt{sorry}.
The definitions section is usually empty, since executable definitions are
provided by the vendored project context. The theorem section contains the main
target theorem, also with its proof removed.

This format is deliberately scaffolded but not axiomatized. Helper statements
are copied into the task under fresh names and must be proved by the submitted
solution. The grader checks both the helper targets and the main target, and
rejects solutions that depend on forbidden axioms or on the original redacted
helper proofs. Thus, helper goals guide decomposition without weakening the
verification requirement.


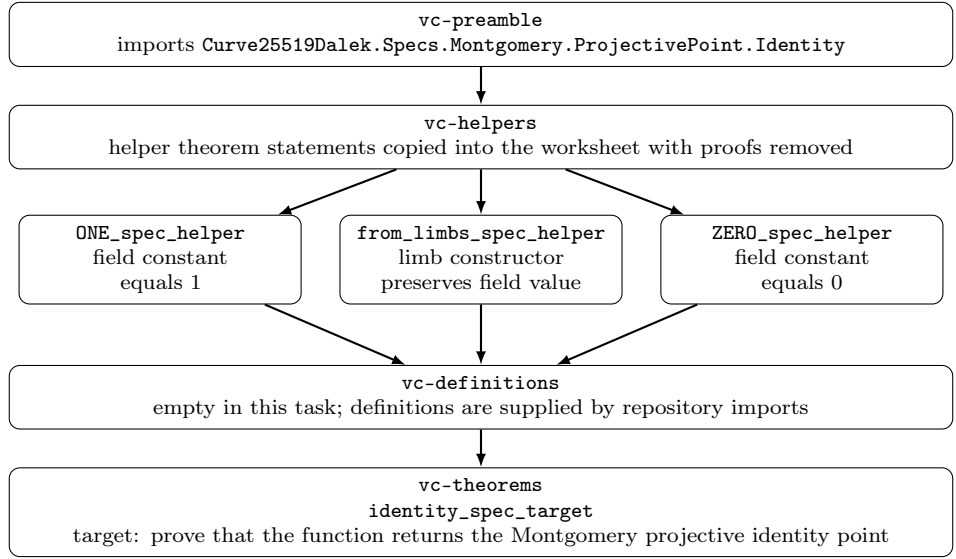
\begin{figure}[t]
\centering
\begin{adjustbox}{max width=\textwidth}
\begin{tikzpicture}[
  node distance=5mm,
  main/.style={
    draw,
    rounded corners,
    align=center,
    inner sep=4pt,
    text width=12.2cm,
    font=\scriptsize
  },
  helper/.style={
    draw,
    rounded corners,
    align=center,
    inner sep=4pt,
    text width=3.45cm,
    font=\scriptsize
  },
  arrow/.style={-{Latex[length=1.8mm]}, thick}
]

\node[main] (preamble) {
  \textbf{\texttt{vc-preamble}}\\
  imports \texttt{Curve25519Dalek.Specs.Montgomery.ProjectivePoint.Identity}
};

\node[main, below=of preamble] (helpers) {
  \textbf{\texttt{vc-helpers}}\\
  helper theorem statements copied into the worksheet with proofs removed
};

\node[helper, below=6mm of helpers] (limbs) {
  \texttt{from\_limbs\_spec\_helper}\\
  limb constructor\\
  preserves field value
};

\node[helper, left=5mm of limbs] (one) {
  \texttt{ONE\_spec\_helper}\\
  field constant\\
  equals \(1\)
};

\node[helper, right=5mm of limbs] (zero) {
  \texttt{ZERO\_spec\_helper}\\
  field constant\\
  equals \(0\)
};

\node[main, below=8mm of limbs] (defs) {
  \textbf{\texttt{vc-definitions}}\\
  empty in this task; definitions are supplied by repository imports
};

\node[main, below=of defs] (target) {
  \textbf{\texttt{vc-theorems}}\\
  \texttt{identity\_spec\_target}\\
  target: prove that the function returns the Montgomery projective identity point
};

\draw[arrow] (preamble) -- (helpers);
\draw[arrow] (helpers) -- (one);
\draw[arrow] (helpers) -- (limbs);
\draw[arrow] (helpers) -- (zero);

\draw[arrow] (one) -- (defs);
\draw[arrow] (limbs) -- (defs);
\draw[arrow] (zero) -- (defs);
\draw[arrow] (defs) -- (target);

\end{tikzpicture}
\end{adjustbox}

\caption[Worksheet structure of an example Dalek Bench task]{
Concrete worksheet anatomy for an example Dalek Bench \texttt{identity\_spec} task.
The solver receives repository imports, three helper proof obligations, and the
main target theorem. Helper goals expose proof structure but must themselves be
proved; they are not hidden axioms.
}
\label{fig:dalek-worksheet-identity}
\end{figure}

\subsection*{Benchmark Analysis}

The helper-scaffolded Dalek Bench contains \(190\) main benchmark tasks and
\(4{,}933\) copied helper goals, for \(5{,}123\) total proof obligations.
Helper goals appear in \(149\) of the \(190\) tasks, while \(41\) tasks expose
only the main target theorem. Across all tasks, the median helper count is
\(10\), and the largest task contains \(179\) helpers. This wide range makes the
benchmark useful for evaluating both direct proof completion and larger
decomposition-style proof worksheets.

\begin{figure}[ht]
\centering
\includegraphics[width=0.8\linewidth]{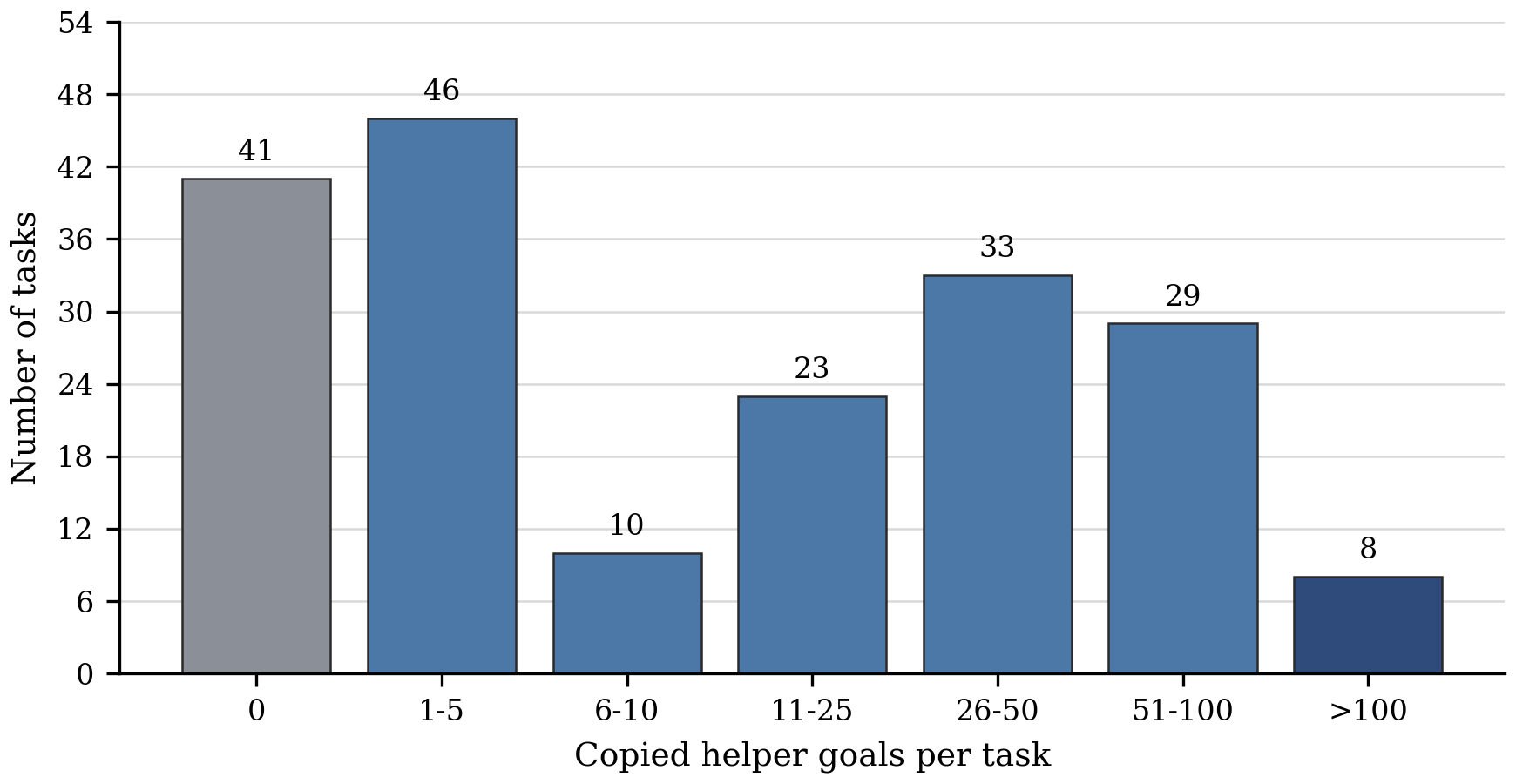}
\caption[Distribution of helper goals per Dalek Bench task]{
Distribution of helper goals per task in the helper-scaffolded Dalek Bench.
The benchmark contains both direct proof-completion tasks with no helper goals
and long-tail worksheet tasks with many helper obligations; the median task has
\(10\) helpers, and the largest task has \(179\).
}
\label{fig:dalek-helper-distribution}
\end{figure}

Figure~\ref{fig:dalek-helper-distribution} shows the aggregate helper-count
distribution. Table~\ref{tab:dalek-helper-burden-by-area} complements this by
breaking the workload down by specification area. This distinction is useful:
the figure shows the overall difficulty profile seen by an agent, while the
table shows which parts of the cryptographic verification stack contribute the
largest helper burden.

\begin{table}[ht]
\centering
\small
\setlength{\tabcolsep}{4pt}
\renewcommand{\arraystretch}{1.08}
\begin{tabular*}{\textwidth}{@{\extracolsep{\fill}}lrrrr@{}}
\toprule
\textbf{Spec area} &
\textbf{Tasks} &
\textbf{With helpers} &
\textbf{Helper goals} &
\textbf{Median helpers} \\
\midrule
\textbf{Montgomery} & 20 & 16 & 1,163 & 74 \\
\textbf{Scalar} & 35 & 27 & 968 & 33 \\
\textbf{Ristretto} & 20 & 17 & 937 & 19 \\
\textbf{Edwards} & 30 & 27 & 617 & 4.5 \\
\textbf{Field} & 7 & 7 & 404 & 50 \\
\textbf{Backend scalar arithmetic} & 20 & 18 & 357 & 19 \\
\textbf{Backend curve models} & 19 & 18 & 319 & 4 \\
\textbf{Backend field arithmetic} & 20 & 15 & 163 & 2 \\
\textbf{Constants} & 19 & 4 & 5 & 0 \\
\midrule
\textbf{Total} & \textbf{190} & \textbf{149} & \textbf{4,933} & -- \\
\bottomrule
\end{tabular*}
\caption[Dalek Bench helper burden by proof topic]{
Helper burden in the helper-scaffolded Dalek Bench, grouped by specification
area. The table reports the number of tasks in each area, how many expose
helper goals, the total number of helper goals contributed by that area, and
the median helper count per task. These quantities summarize the worksheet load
seen by an agent, rather than the upstream extraction-funnel counts.
}
\label{tab:dalek-helper-burden-by-area}
\end{table}

Together, the distribution figure and area table characterize the benchmark at
two levels. The distribution figure shows that Dalek Bench has a long-tailed
helper structure, with many small tasks and a smaller number of large
worksheet-style tasks. The area table shows that this burden is concentrated in
specific proof families: Montgomery, Scalar, and Ristretto contribute the most
helper goals, while Field has relatively few tasks but high per-task helper
counts. This structure makes aggregate pass rate insufficient on its own; later
experiments should also report whether scaffold improvements concentrate in
direct proofs, arithmetic-heavy obligations, or dependency-heavy worksheet
tasks.

\subsection*{Role in the Thesis}

Dalek Bench complements the Vericoding Benchmark by moving from standardized,
mostly self-contained vericoding tasks to repository-grounded Lean proof repair.
Its helper-scaffolded format is especially useful for evaluating agentic
systems: helper goals expose intermediate proof structure, the repository
context creates a need for local-context retrieval and tool use, and the Lean
checker provides a precise success signal. The benchmark therefore serves as
the main repository-scale evaluation setting for the inference-time scaffolds
introduced later in the thesis.
\chapter{Part I: Post-Training}

\section{Preliminaries: Reinforcement Learning Mechanics}
\label{sec:rlvr-preliminaries}

\paragraph{Why reinforcement learning for LLMs?}
Reinforcement learning for language models treats generation itself as
the action space: given a prompt \(x\), the model samples a token
trajectory \(y=(y_1,\ldots,y_T)\) autoregressively, with $y_t \sim \pi_\theta(\cdot \mid x,y_{<t})$,
which receives a scalar outcome reward \(r(x,y)\), and updates
\(\pi_\theta\) to assign higher probability to successful rollouts.
This differs from supervised fine-tuning, whose objective is to imitate a
fixed demonstration \(y^\star\) by maximizing the likelihood of each
reference token under ground truth completions:
\[
    \mathcal{L}_{\mathrm{SFT}}(\theta)
    =
    -
    \sum_{t=1}^{T}
    \log
    \pi_\theta(y_t^\star \mid x,y_{<t}^\star).
\]
SFT therefore provides a dense token-level imitation signal: the model is
rewarded for matching the demonstrated trajectory. RL instead optimizes a
sequence-level objective of the form
\[
    \max_\theta
    \;
    \mathbb{E}_{y \sim \pi_\theta(\cdot \mid x)}
    [
        r(x,y)
    ],
\]
where the reward depends on properties of the completed output rather
than on agreement with ground truth tokens. This distinction is
crucial for reasoning and code. Generated rollouts may differ from the
demonstration token-by-token while still producing the correct answer. Though more expensive, we fully maximize exploration space this way.

RL with human feedback (RLHF) made this paradigm practical for instruction-following by replacing
hand-written rewards with a learned reward model trained from human
preferences \citep{ouyang2022instructgpt}, the brekathrough for ChatGPT. RL with verifiable rewards
(RLVR) removes the learned reward model from the loop and instead scores
each sampled trajectory with a deterministic checker, such as a unit
test, answer verifier, theorem prover, or formal-verification toolchain (as done in this thesis)
\citep{lambert2024tulu3}. The resulting loop is especially powerful for
reasoning domains: candidate solutions can be generated at scale, checked
automatically, and recycled into policy-gradient updates, which has made
RLVR a central ingredient in recent high-reasoning training recipes
\citep{10.48550/arxiv.2501.12948}.

\paragraph{From PPO to GRPO.}
Diving deeper into the progression of RL algorithms for LLMs, a key part of the standard RL objective is augmenting the
outcome reward with a penalty that keeps the learned policy close to a
reference model ($\pi_{\mathrm{ref}}$), usually the supervised fine-tuned policy:
\[
    \max_\theta
    \;
    \mathbb{E}_{x \sim \mathcal{D},\,
    y \sim \pi_\theta(\cdot \mid x)}
    [
        r(x,y)
    ]
    -
    \beta
    D_{\mathrm{KL}}
    (
        \pi_\theta(\cdot \mid x)
        \Vert
        \pi_{\mathrm{ref}}(\cdot \mid x)
    ).
\]
This regularization prevents the policy from moving too far from the
language distribution learned during pretraining and supervised
fine-tuning, while the reward term selects for task success
\citep{ouyang2022instructgpt}.

Proximal Policy Optimization (PPO) became the default optimizer for this
objective alongside RLHF \citep{schulman2017ppo, ouyang2022instructgpt}. Given a
rollout \(y=(y_1,\ldots,y_T)\), PPO computes a token-level ratio to compare the generation likelihood under the active policy ($\pi_\theta$) against the historical policy ($\pi_{old}$) used to sample the rollout.
\[
    \rho_t(\theta)
    =
    \frac{
    \pi_\theta(y_t \mid x,y_{<t})
    }{
    \pi_{\theta_{\mathrm{old}}}(y_t \mid x,y_{<t})
    }.
\]
The objective function for PPO is
\[
\mathcal{L}_{\mathrm{PPO}}(\theta)
=
\mathbb{E}
\left[
\sum_{t=1}^{T}
\min
\left(
\rho_t(\theta) A_t,
\operatorname{clip}
\left(
\rho_t(\theta),1-\epsilon,1+\epsilon
\right)A_t
\right)
\right],
\]
where \(A_t\) is an advantage estimate,  defined as the
difference between the observed return and a value prediction ($A_t \approx R_t - V_\psi(x,y_{<t})$).
Intuitively, \(A_t\) measures whether the sampled continuation performed
better or worse than expected under the current policy. Tokens lying on
higher-than-expected trajectories receive positive advantage and are made
more likely, while tokens on lower-than-expected trajectories receive
negative advantage and are suppressed. Another key term is the clipping term, which prevents a
single update from changing the policy too aggressively.

For long reasoning, code, and proof trajectories, the main difficulty is
estimating \(A_t\). PPO typically trains a value model to predict future
return from partial trajectories. However, in RLVR, verifier rewards are sparse,
terminal, and often available only after an entire candidate has been generated. Predicting
eventual verifier success from partial code or proof prefixes becomes difficult. Group Relative Policy
Optimization (GRPO) avoids this issue through bypassing a value (critic) model, comparing multiple
rollouts for the same prompt \citep{shao2024deepseekmath}.

For each prompt \(x\), GRPO samples a group of \(K\) completions $y_1,\ldots,y_K \sim \pi_{\theta_{\mathrm{old}}}(\cdot \mid x)$
and scores them with the task reward \(r_i=r(x,y_i)\). We then define a normalized group-relative advantage
\[
    \hat A_i
    =
    \frac{r_i-\mu_x}{\sigma_x+\epsilon_A}.
\]
Where $\mu_x, \sigma_x$ are the group reward mean and variance. This advantage replaces a value model-dependent one as done in PPO, yielding
the clipped objective
\[
\mathcal{L}_{\mathrm{GRPO}}(\theta)
=
\mathbb{E}
\left[
\frac{1}{K}
\sum_{i=1}^{K}
\frac{1}{|y_i|}
\sum_{t=1}^{|y_i|}
\min
\left(
\rho_{i,t}(\theta)\hat A_i,
\operatorname{clip}
\left(
\rho_{i,t}(\theta),1-\epsilon,1+\epsilon
\right)
\hat A_i
\right)
\right],
\]

\[
\operatorname{clip}(x,1-\epsilon,1+\epsilon)
=
\min\!\left(\max(x,1-\epsilon),1+\epsilon\right).
\]
In the verifier-reward setting, prompt groups naturally contrast verified
and failed completions. The per-response averaging in GRPO keeps the
magnitude of each sampled completion's update roughly independent of its
length, while the group-relative baseline removes the need for a learned
critic. This makes GRPO a natural default for RLVR pipelines
\citep{10.48550/arxiv.2501.12948}.

\paragraph{DAPO-style refinements.}
Decoupled Clip and Dynamic Sampling Policy Optimization (DAPO), introduced by \citet{yu2025dapo}, was proposed as a practical
refinement of GRPO for reproducing R1-style long-CoT reinforcement
learning at open-source scale. Its motivation is that the basic
group-relative objective is simple and critic-free, but brittle in the
regime that matters for reasoning: rewards are sparse, successful
rollouts are rare early in training, many prompt groups contain only
successes or only failures, and long generations interact badly with both
loss normalization and truncation. DAPO addresses these issues with four
modifications.

\begin{enumerate}
    \item \textbf{Dynamic sampling.}
    Prompt groups whose rollouts all receive the same reward provide no
    relative training signal. DAPO therefore only keeps groups with
    nonzero reward variance, i.e.
    \(\operatorname{Var}(\{r(x,y_i)\}_{i=1}^{K})>0\). For binary verifier
    rewards, this means keeping prompts where some but not all sampled
    completions verify (Note $\mathcal{V}$ is our verifier and $\operatorname{Assemble}$ ``assembles" our proof given the LLM response) :
    \[
        0
        <
        \frac{1}{K}
        \sum_{i=1}^{K}
        \mathbf{1}
        [
            \mathcal{V}(\operatorname{Assemble}(x,y_i))=\texttt{verified}
        ]
        <
        1.
    \]

    \item \textbf{Token-level policy-gradient loss.}
    The sequence-level GRPO objective averages each completion's token
    losses as \(\frac{1}{K}\sum_i (1/|y_i|)\sum_t \ell_{i,t}\), giving each
    completion equal weight regardless of length. DAPO instead averages
    over all generated tokens,
    \((1/\sum_i |y_i|)\sum_i\sum_t \ell_{i,t}\), so long reasoning
    trajectories are not systematically down-weighted.

    \item \textbf{Asymmetric clipping.}
    Clip-Higher replaces the symmetric PPO clipping interval with
    \(\operatorname{clip}(\rho,1-\epsilon_{\mathrm{low}},
    1+\epsilon_{\mathrm{high}})\), where
    \(\epsilon_{\mathrm{high}}>\epsilon_{\mathrm{low}}\). This allows
    rare high-reward trajectories to be reinforced more strongly while
    still well-bounding destructive policy updates.

    \item \textbf{Overlong reward shaping.}
    DAPO treats truncation as a distinct failure mode by penalizing
    generations that exceed a soft length threshold, for example with
    \(r_{\mathrm{len}}(x,y)=-\lambda_{\mathrm{len}}\max(0,
    (|y|-L_{\mathrm{soft}})/(L_{\mathrm{hard}}-L_{\mathrm{soft}}))\).
    This prevents overlong or truncated completions from being conflated
    with ordinary verifier failures.
\end{enumerate}

\paragraph{Dafny verifier rewards and success metrics.}

We instantiate this RLVR setup with Dafny. A prompt \(x \sim
\mathcal{D}\) contains a Dafny task, including some combination of a
natural-language description, method signature, formal contract, and
partial implementation. The model samples a completion
\(y \sim \pi_\theta(\cdot \mid x)\), which is inserted into the surrounding
template to produce a candidate Dafny artifact $a = \operatorname{Assemble}(x,y)$
The Dafny verifier is then a deterministic terminal evaluator with the following outcomes:
\[
\mathcal{V}_{\mathrm{Dafny}}(a)
\in
\left\{
\begin{aligned}
&\texttt{parse\_error},\\
&\texttt{compile\_error},\\
&\texttt{verification\_error},\\
&\texttt{timeout},\\
&\texttt{verified}
\end{aligned}
\right\}
\]

The simplest verifier reward is therefore the binary outcome
\[
    r_{\mathrm{raw}}(x,y)
    =
    \mathbf{1}
    [
        \mathcal{V}_{\mathrm{Dafny}}(\operatorname{Assemble}(x,y))
        =
        \texttt{verified}
    ].
\]
This is the source of Dafny's appeal as an RLVR environment: the reward is
machine-checkable, deterministic under a fixed verifier configuration,
and does not require human preference labels or a learned reward model.

However, Dafny verification proves correctness only relative to the
contracts and assumptions present in the candidate artifact. A file that
verifies satisfies the specification it presents to Dafny,
\[
    \mathcal{V}_{\mathrm{Dafny}}(a)=\texttt{verified}
    \Rightarrow
    a \models \varphi_{\mathrm{given}},
\]
but this does not imply that the given specification $\varphi_{\mathrm{given}}$ matches the intended
task specification \(\varphi_{\mathrm{intended}}\). In particular, a model
may obtain verifier success by weakening postconditions, strengthening
preconditions, introducing trusted assumptions, or otherwise changing the
task being verified, failure modes also emphasized in prior Dafny
benchmarks and training work \citep{loughridge2024dafnybench,
yan2025reform}. We therefore distinguish raw verifier success from
stricter task success throughout the experiments.

We define strict success as raw verifier success plus two additional
filters: the completion must preserve the intended specification and avoid
forbidden verifier escape hatches. Hack rate is the fraction of raw
verifier passes that fail this stricter criterion:
\[
    \operatorname{StrictSuccess}(x,y)
    =
        \operatorname{RawPass}(x,y)
        \land
        \operatorname{SpecPreserved}(x,y)
        \land
        \operatorname{NoForbiddenConstructs}(y)
\]
\[
    \operatorname{HackRate} (x,y)
    =
    \Pr[
        \operatorname{StrictSuccess}(x,y)=0
        \mid
        \operatorname{RawPass}(x,y)=1
    ].
\]

Here, \(\operatorname{SpecPreserved}\) checks that protected parts of the
task are unchanged: method signatures, types, \texttt{requires} clauses,
\texttt{ensures} clauses, and provided termination or framing annotations.
For example, changing \texttt{ensures sorted(a)} to \texttt{ensures true}
or adding \texttt{requires a.Length == 0} fails this check.
\(\operatorname{NoForbiddenConstructs}\) filters Dafny escape hatches such
as \texttt{assume false}, unconstrained \texttt{assume} statements, and
\verb|{:verify false}| annotations. A completion that verifies only by
weakening the contract or disabling the proof obligation is therefore
counted as a raw pass but not as a strict success.

\section{Setup: Models, Data, Infrastructure, Evaluation}

\begin{figure}
    \centering
    \includegraphics[width=\linewidth]{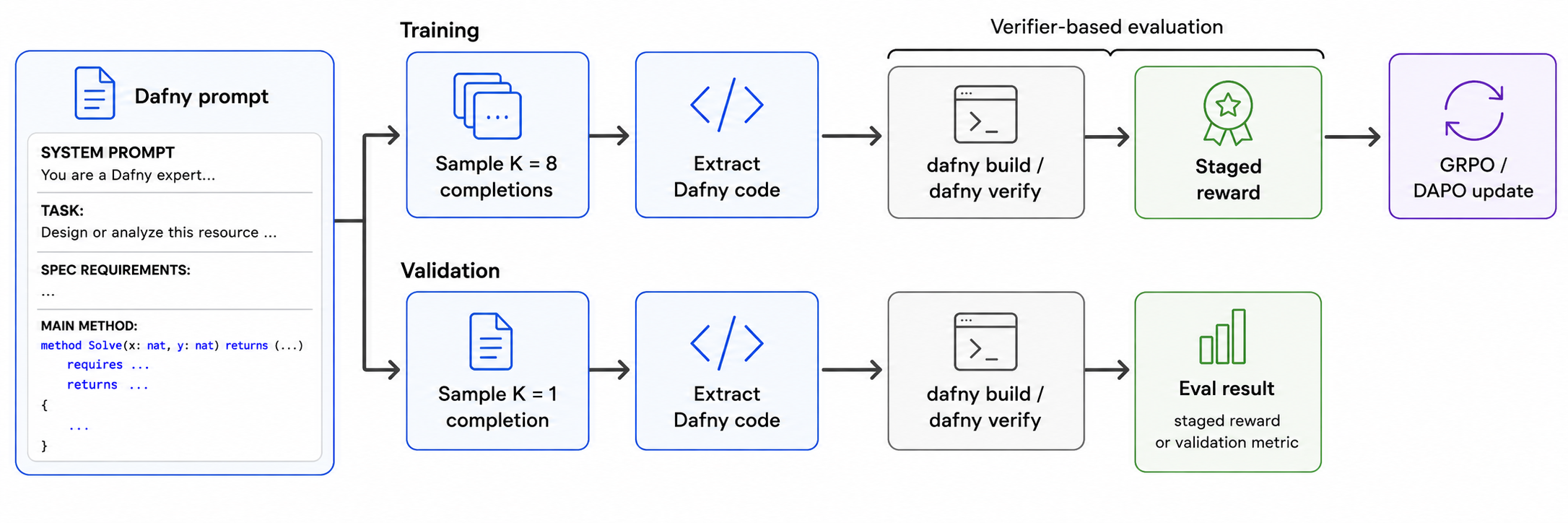}
    \caption[Single-turn Dafny RLVR training and validation pipeline]{
\textbf{    Single-turn Dafny RLVR training and validation pipeline.}
    Training uses grouped rollouts, sampling \(K=8\) completions per prompt before extracting Dafny code and evaluating each completion with \texttt{build} / \texttt{verify}.
    Verifier outcomes are mapped to a staged reward and used for the DAPO update.
    Validation reuses the same extraction and verifier-based scoring pipeline with \(K=1\), reporting staged reward on the held-out split.
    }
        \label{fig:dafny_initial_pipeline}
\end{figure}

This section describes the initial single-turn Dafny RLVR setup, visualized in Fig. \ref{fig:dafny_initial_pipeline}. The goal
of this experiment was to test whether a language model could be trained
directly from Dafny verifier feedback using the GRPO/DAPO-style objective
introduced above.

\paragraph{Model.}
The actor policy was initialized from
\texttt{Qwen/Qwen2.5-7B-Instruct}. We selected a 7B-parameter backbone to keep
rollout-based RLVR tractable under the available compute budget. In this setting,
training cost is dominated not only by policy optimization, but also by sampling
long responses, generating multiple completions per prompt for group-relative
advantage estimation, and executing the verifier on each sampled artifact. These
costs made substantially larger actor models impractical for our experimental
setup, which had access to at most roughly eight GPUs. We trained using the
\texttt{verl} reinforcement-learning framework with the GRPO advantage estimator
and the DAPO reward manager \citep{sheng_hybridflow_2025}. Each prompt produced \(K=8\) sampled completions per
rollout batch, matching the group-relative setup from
Section~\ref{sec:rlvr-preliminaries}. The maximum prompt length was 2048 tokens
and the maximum response length was 8192 tokens (See Table~\ref{tab:initial_rlvr_config}.

\paragraph{Dataset.}
The dataset consists of APPS-derived programming tasks translated into
Dafny verification problems taken from the \texttt{vericoding} benchmark \citep{10.48550/arxiv.2509.22908}. Each example is formatted as a single user
message with \texttt{ability = dafny}. A prompt contains a
natural-language programming task, formal Dafny specification
requirements, optional helper predicates or functions, and a target
method or function with an empty body to synthesize. The common initial file is formatted as:
\begin{verbatim}
// ======= TASK =======
// Natural-language programming problem

// ======= SPEC REQUIREMENTS =======
// Dafny predicates, functions, requires clauses, and ensures clauses

// ======= HELPERS =======
// Optional helper definitions

// ======= MAIN METHOD =======
method solve(...) returns (...)
    requires ...
    ensures ...
{
}
\end{verbatim}

The training split contains 618 prompts and the held-out validation split
contains 265 prompts. Although most target methods are named
\texttt{solve}, some examples use task-specific names; we therefore refer
to the generated object as the target method or function rather than
assuming a fixed method name.

\begin{table}[h]
\centering
\begin{tabular}{lrrrr}
\toprule
Split & Examples & Median chars & Mean chars & 95th percentile chars \\
\midrule
Train & 618 & 1394 & 1711.5 & 4008.8 \\
Validation & 265 & 1323 & 1665.1 & 3556.8 \\
\bottomrule
\end{tabular}
\caption{Prompt-level statistics for the Dafny APPS-derived dataset.}
\label{tab:dafny_dataset_stats}
\end{table}

\paragraph{Training infrastructure.}
Training was run on an 8-GPU node with NVIDIA A100-SXM4-80GB GPUs. Each
training batch contained 16 prompts, and each prompt produced 8 sampled
responses before filtering, yielding up to 128 Dafny reward evaluations
per rollout batch. Training was configured for 30 epochs, with validation
enabled before the first training update.

\begin{table}[h]
\centering
\begin{tabular}{ll}
\toprule
Setting & Value \\
\midrule
Base model & \texttt{Qwen/Qwen2.5-7B-Instruct} \\
RL framework & \texttt{verl} \\
Advantage estimator & \texttt{grpo} \\
Reward manager & \texttt{dapo} \\
Prompt batch size & 16 \\
Rollouts per prompt & 8 \\
PPO mini-batch size & 4 \\
Max prompt length & 2048 \\
Max response length & 8192 \\
Weight decay & 0.1 \\
Clip ratio low / high & 0.20 / 0.28 \\
\bottomrule
\end{tabular}
\caption{Initial RLVR training configuration.}
\label{tab:initial_rlvr_config}
\end{table}

\paragraph{KL configuration.}
Although Section~\ref{sec:rlvr-preliminaries} presents the general
KL-regularized RLVR objective, runs disable both reward-level
and loss-level KL penalties. Thus, this isolates the effect of verifier-derived
reward under a critic-free group-relative update, without explicit KL
anchoring to the reference policy.

\paragraph{Verifier and reward function.}
Each sampled model response was converted into Dafny code before scoring. The
extracted code was checked in two stages. First, Dafny was run in build
mode with verification disabled.
If the artifact built successfully, the reward function then ran verification.
The initial reward was staged:
\[
r_0(x,y)
=
\begin{cases}
1.0, & \text{if the artifact builds, verifies, and passes anti-cheating checks},\\
0.1, & \text{if the artifact builds and passes anti-cheating checks but does not verify},\\
0.0, & \text{otherwise}.
\end{cases}
\label{eq:reward_init}
\]
The anti-cheating check compared the pre and post condition (i.e., \texttt{requires, ensures}) clauses in the original template against those in the
generated artifact after whitespace normalization. It also filtered
explicit Dafny escape hatches, including \texttt{assume false} and
\verb|{:verify false}|. Thus, a completion that verifies only by weakening
the contract or disabling verification receives the cheating reward,
which was set to \(0.0\) in the initial run.

\paragraph{Evaluation.}
Evaluation used the held-out Dafny APPS-derived validation split of 265
problems. In the initial run, evaluation was based primarily on the
staged reward returned by the Dafny reward function. Thus, the first train/test
curves should be read as aggregate reward curves rather than a full
decomposition of verifier behavior. Later sections refine this signal by
separating build success, verification success, specification changes, and
reward-hacking examples.

\section{RL from Verifier Feedback: Initial Results}
\label{sec:initial-rlvr-results}

Having defined the model, data, reward function, and training configuration
above, we now analyze the behavior of the initial single-turn Dafny RLVR run.
For each prompt, the policy sampled Dafny completions, the extracted artifacts
were scored by the staged Dafny build/verify reward (see
Eq.~\ref{eq:reward_init}), and DAPO updates were applied using these
verifier-derived rewards.

\begin{figure}[ht]
    \centering
    \includegraphics[width=\linewidth]{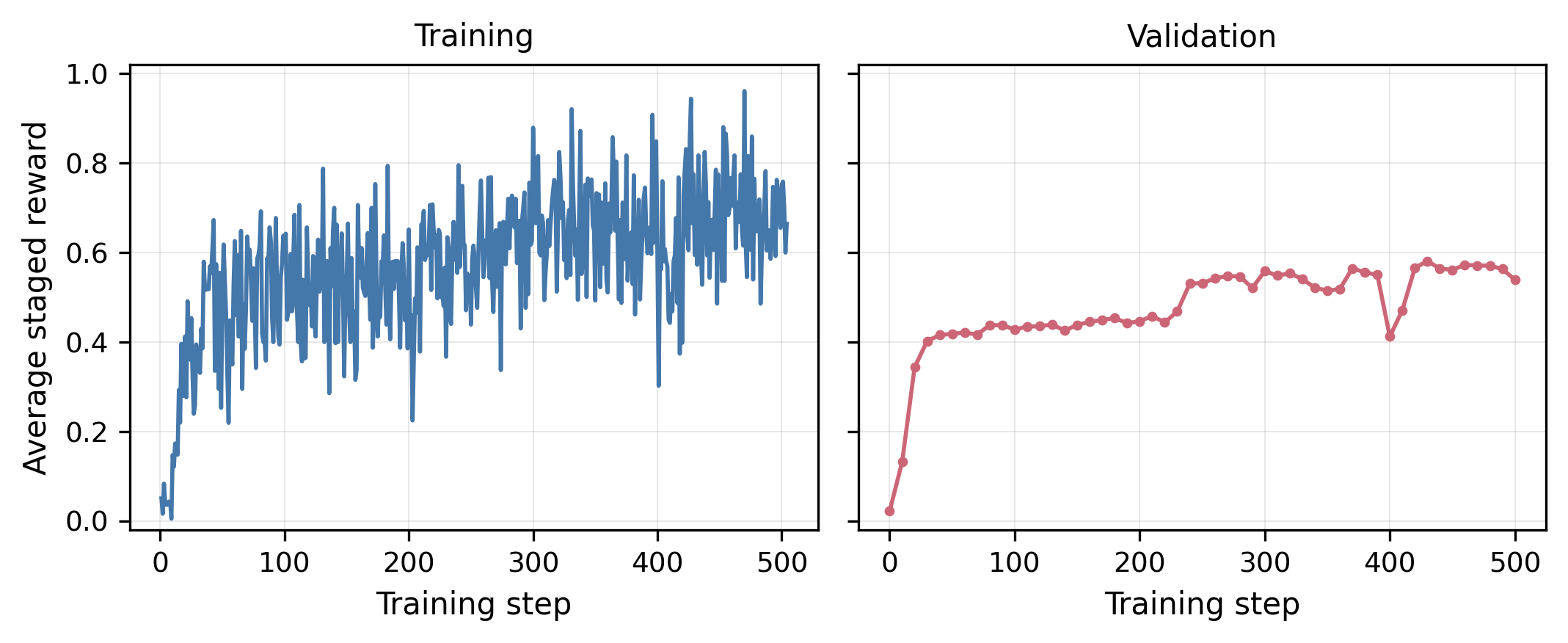}
    \caption[Reward trajectory for the initial single-turn Dafny RLVR run]{Reward trajectory for the initial single-turn RLVR run. The training curve reports the mean verifier-based reward over sampled training rollouts, where each prompt produces \(K=8\) completions for group-relative optimization. The validation curve reports mean reward on the held-out validation set using a single sampled completion per prompt, so it measures held-out single-sample performance rather than best-of-\(K\) success. Training reward is logged at every optimization step, while validation reward is evaluated every five steps, producing a sparser validation curve. The validation reward falls below the training reward on average, a pattern consistent with overfitting to the training distribution.}
    \label{fig:initial_rlvr_reward}
\end{figure}

\paragraph{Aggregate reward dynamics.}
Figure~\ref{fig:initial_rlvr_reward} shows the average staged reward on the
training and validation splits over the course of RL training. Because the
reward function assigns partial credit for buildable but non-verifying artifacts,
these curves measure aggregate verifier-facing progress rather than verified
correctness alone.

The reward curves show that the model quickly learned to optimize the staged
verifier signal. Before RL, the base policy achieved an average validation
reward of only \(0.022\), indicating that almost none of its sampled completions
received meaningful reward under the Dafny pipeline. After RL training, the
validation reward rose sharply, reaching a best value of \(0.581\) at step 430.
This suggests that the verifier reward was not too sparse to train on.

At the same time, the later part of training shows signs of over-optimization.
The final checkpoint reached a higher sampled training reward of \(0.752\), but
its validation reward decreased to \(0.539\) (See Table~\ref{tab:initial_rlvr_rewards}. This divergence suggests that the model continued to improve
on the sampled training rollouts while becoming less aligned with the held-out
validation distribution, a sign of over-fitting.

\begin{table}[hb]
\centering
\begin{tabular}{lrrrr}
\toprule
Checkpoint & Step/Epoch & Train reward & Validation reward & Train--val gap \\
\midrule
Before RL & 0 / 0.00 & -- & 0.022 & -- \\
Best validation & 430 / 11.32 & 0.595 & 0.581 & +0.013 \\
Final validation & 500 / 13.16 & 0.752 & 0.539 & +0.212 \\
\bottomrule
\end{tabular}
\caption[Aggregate reward statistics for the initial Dafny RLVR run]{Aggregate reward statistics for the initial single-turn Dafny RLVR run. Validation reward is the mean verifier reward over the validation set; train reward is the mean verifier reward over sampled training rollouts.}
\label{tab:initial_rlvr_rewards}
\end{table}

Across the run, training completed 13.263 epochs and made an estimated 199,662
Dafny verifier calls. These results show that verifier-derived rewards provide
a usable optimization signal for the policy, but they do not by themselves
establish improved verification ability.

\paragraph{Limitations of aggregate reward.}
Aggregate reward conflates several behaviors: producing syntactically valid
Dafny, satisfying the intended specification, exploiting prompt artifacts, and
modifying the effective verification problem while passing the anti-cheating
checks. Thus, the reward curve measures optimization of the verifier pipeline,
not necessarily semantic progress on the original tasks.

This motivates the rollout-level analysis in the next section. In particular,
we inspect whether reward improvements correspond to genuine verification
success or to degenerate strategies that exploit the staged reward function.

\section{Specification Hacking: Discovery and Analysis}
\label{sec:spec-hacking}

\paragraph{Rollout-level diagnostic.}
To investigate the source of the reward gains, we inspected the five validation
prompts logged at each validation checkpoint. This analysis is diagnostic rather
than exhaustive: it does not decompose the full validation set, but instead uses
the logged rollouts to identify concrete mechanisms by which high reward was
obtained.

\begin{figure}[hbt]
    \centering
    \includegraphics[width=\linewidth]{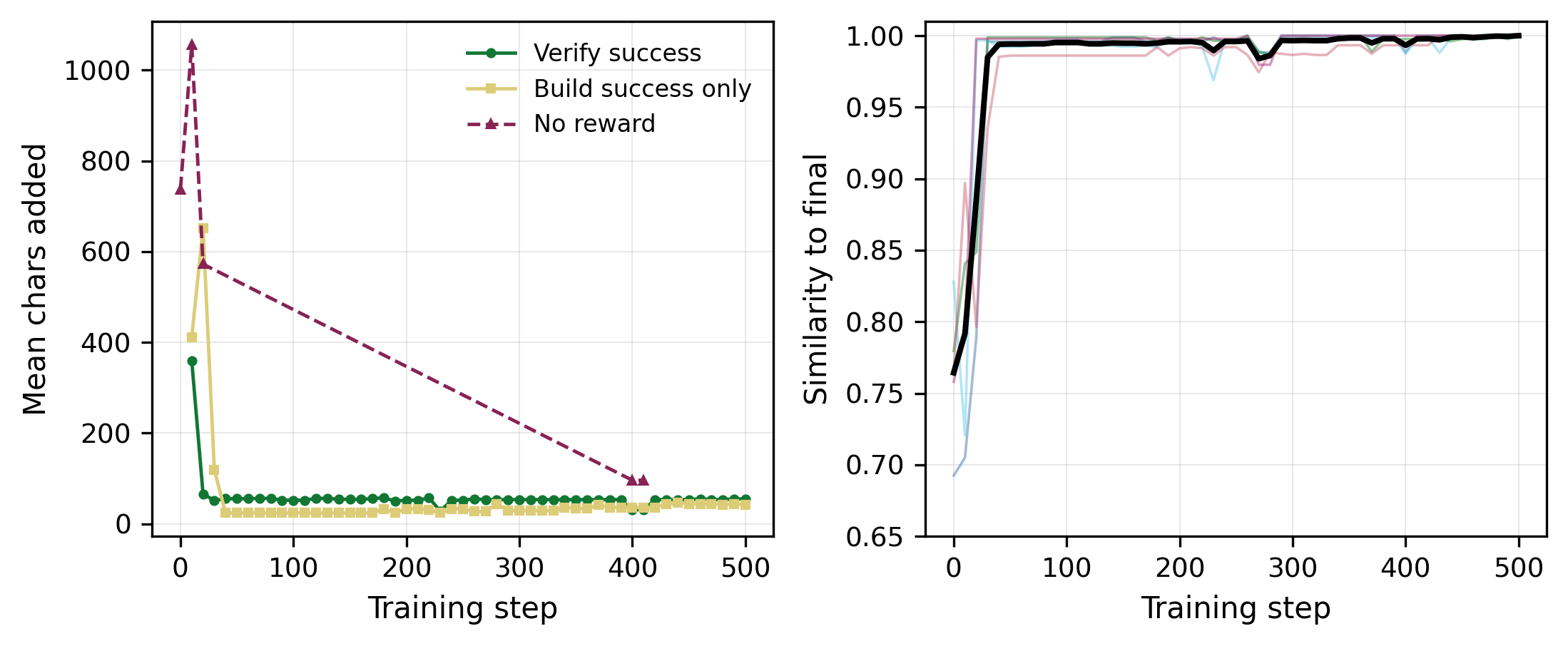}
    \caption[Logged validation diagnostics for the initial Dafny RLVR run]{Logged validation diagnostics for the initial single-turn RLVR run.
    \textbf{Left:} Mean number of generated characters added to the prompt program,
    grouped by verifier outcome. \textbf{Right:} Character-level mean similarity
    between each checkpoint output and the final generated program for each of the
    five logged prompts, with the across-prompt mean shown in black. The rapid
    collapse in generated length and the early rise in similarity toward one suggest
    that the model quickly converges to short, repeated outputs, consistent with
    specification-hacking behavior rather than substantive program synthesis.}
    \label{fig:initial_rlvr_logged_diagnostics}
\end{figure}

\paragraph{Observed patterns.}
Figure~\ref{fig:initial_rlvr_logged_diagnostics} shows two patterns in the
logged rollouts. First, verified completions are often short, typically adding
only a small method body or a direct call to an existing helper. Second,
generations become increasingly stable across checkpoints: the final-output
similarity rises over training, indicating convergence toward repeated Dafny
templates for the logged prompts. These patterns suggest that the policy learned
compact reward-seeking behaviors, not just general-purpose Dafny synthesis.

\paragraph{Specification artifacts.}
Manual inspection of the logged validation completions revealed two dominant
failure modes. First, some tasks had trivial or underconstrained
postconditions, allowing constant implementations to verify. Second, some tasks
leaked the solution through helper functions or predicates that appeared
directly in the postcondition. In both cases, the model received verifier
reward without synthesizing the intended algorithm.

\paragraph{Example 1: Trivial Specification.}
In one logged prompt (Cowbell Packing), the postcondition only required the returned value to be
nonnegative:
\begin{verbatim}
method solve(n: int, k: int, L: seq<int>)
  returns (result: int)
  requires ValidInput(n, k, L)
  ensures result >= 0
{
}
\end{verbatim}
The following completion verifies:
\begin{verbatim}
{
    result := 0;
}
\end{verbatim}
This is a valid implementation with respect to the formal contract, but it does
not solve the intended task. The verifier reward is therefore correct relative
to the specification, while the specification itself is too weak to define the
desired behavior.

\paragraph{Example 2: Specification Leakage.}
In another logged prompt (Farm-area), the specification exposed a helper function that
directly computed the required output:
\begin{verbatim}
function RemainingFarmArea(a: int, b: int): int
  requires ValidFarmDimensions(a, b)
{
  a * b - a - b + 1
}

method solve(a: int, b: int) returns (result: int)
  requires ValidFarmDimensions(a, b)
  ensures result == RemainingFarmArea(a, b)
  ensures result >= 0
{
}
\end{verbatim}
The model can obtain full verifier reward by simply returning the leaked helper:
\begin{verbatim}
{
    result := RemainingFarmArea(a, b);
}
\end{verbatim}
Again, the completion satisfies the formal specification, but it bypasses the
intended synthesis problem. Rather than deriving or implementing the algorithm,
the model delegates the computation to an abstraction already provided in the
prompt. We also found examples where the predicate itself encodes the target behavior (See Appendix).

Together, these examples illustrate two distinct ways that high verifier reward
can fail to reflect the intended synthesis task. We also observe a weaker form of reward
optimization in which the model converges toward buildable Dafny templates,
consistent with the partial reward assigned to non-verifying artifacts that pass
build checks. Table~\ref{tab:spec_hacking_patterns} summarizes these patterns.

\begin{table}[t]
\centering
\small
\begin{tabular}{p{0.24\linewidth}p{0.34\linewidth}p{0.32\linewidth}}
\toprule
\textbf{Failure mode} & \textbf{Specification artifact} & \textbf{Reward-seeking behavior} \\
\midrule
Trivial specification &
Postcondition underconstrains the intended task. &
Return a constant or no-op value that satisfies the weak contract. \\
Specification leakage &
Helper function or predicate directly encodes the target behavior. &
Call the helper or branch on the predicate instead of synthesizing the
algorithm. \\
Optimize for Build &
Staged reward gives partial credit for code that builds but does not verify. &
Converge toward syntactically valid Dafny templates even without full
verification. \\
\bottomrule
\end{tabular}
\caption[Specification-hacking patterns in the initial Dafny RLVR run]{Specification-hacking patterns observed in the logged validation
rollouts.}
\label{tab:spec_hacking_patterns}
\end{table}

\paragraph{Implications.}
The issue is not that the verifier reward failed, but that it optimized exactly
the formal objective it was given. Trivial contracts rewarded constant
implementations, leaked helpers rewarded direct calls to the answer, and the
staged reward encouraged buildable outputs even without verification. This
motivates the next pipeline revision: filtering the data distribution toward
faithful, non-leaky, and nontrivial specifications so that verifier reward is
better aligned with the intended synthesis task.

\section{Improving the Pipeline: Data Filtering}
\label{sec:data-filtering}

\paragraph{Motivation.}
The previous analysis indicates that the data distribution itself
is part of the reward design. If a prompt contains a weak or leaky
specification, then Dafny verification provides a correct but misaligned reward.
We therefore first filter the dataset using LLM-extracted metadata about specification
faithfulness, specification leakage, and task difficulty. This produces a
cleaner subset in which high reward is more likely to correspond to genuine
verification progress rather than exploitation of specification artifacts. 
For our base dataset, we consider the entire Vericoding benchmark to widen our scope past the initial APPS set.

\begin{table}[htb]
\centering
\begin{tabular}{lrrrr}
\toprule
Source & Initial Dafny & Filtered subset & Train & Test \\
\midrule
\texttt{apps} & 677 & 196 & 155 & 41 \\
\texttt{bignum} & 62 & 50 & 37 & 13 \\
\texttt{dafnybench} & 443 & 294 & 240 & 54 \\
\texttt{humaneval} & 162 & 102 & 86 & 16 \\
\texttt{numpy\_simple} & 58 & 26 & 17 & 9 \\
\texttt{numpy\_triple} & 603 & 309 & 255 & 54 \\
\texttt{verified\_cogen} & 172 & 76 & 57 & 19 \\
\texttt{verina} & 157 & 96 & 72 & 24 \\
\midrule
\textbf{Total} & \textbf{2334} & \textbf{1149} & \textbf{919} & \textbf{230} \\
\bottomrule
\end{tabular}
\caption[Composition of the quality-filtered Dafny tasks]{Source composition of the quality-filtered Dafny tasks drawn from the
Vericoding benchmark \citep{10.48550/arxiv.2509.22908}. The initial Dafny column
reports the number of Dafny tasks available from each source before filtering.
The filtered subset retains tasks with faithful, low-leakage, and nontrivial
specifications, which are then split into train and test sets.}
\label{tab:dafny-hard-subset-source-breakdown}
\end{table}

\paragraph{LLM-generated metadata.}
We filter tasks using three LLM-generated annotations with Claude's Opus 4.5, where each score sits 0-3. The
\texttt{difficulty} score measures the assessed hardness of the task
with larger values indicating more difficult verification problems.
The \texttt{spec\_faithfulness} score measures whether the Dafny specification
matches the intended natural-language task, while \texttt{spec\_leakage}
measures whether the specification reveals excessive implementation structure
or makes the task trivial.

We then applied the following quality filter:
\[
\texttt{spec\_faithfulness} \geq 2,\qquad
\texttt{spec\_leakage} \leq 1,\qquad
\texttt{difficulty} \geq 1.
\]
Unlike the initial RLVR experiment, which used APPS-derived Dafny tasks, this
filter was applied to the full set of Dafny problems included in the Vericoding
benchmark \citep{10.48550/arxiv.2509.22908}. We choose to not constrain the \texttt{spec\_leakage} to 0 to allow for some realistic leeway with specification errors. The resulting data contained 1,149 Dafny verification tasks. The breakdown by source can be found in Table~\ref{tab:dafny-hard-subset-source-breakdown}

\paragraph{Prompt format.}
The hard-subset experiments also use a  different output format from
the initial APPS-derived setup, taken from \citet{10.48550/arxiv.2509.22908}. Instead of asking the model to emit a single
Dafny completion, the prompt asks for a JSON object whose fields separately
specify code for helper definitions, additional definitions, and the main target
implementation. This makes extraction more structured. After generation, the
corresponding JSON fields are then inserted into the appropriate regions of the Dafny
template before running the build and verification checks. The reward function
is still computed from the resulting Dafny artifact, but the model's output
space is more explicitly aligned with the editable regions of the benchmark
files.

\begin{figure}[htb]
\centering
\includegraphics[width=\linewidth]{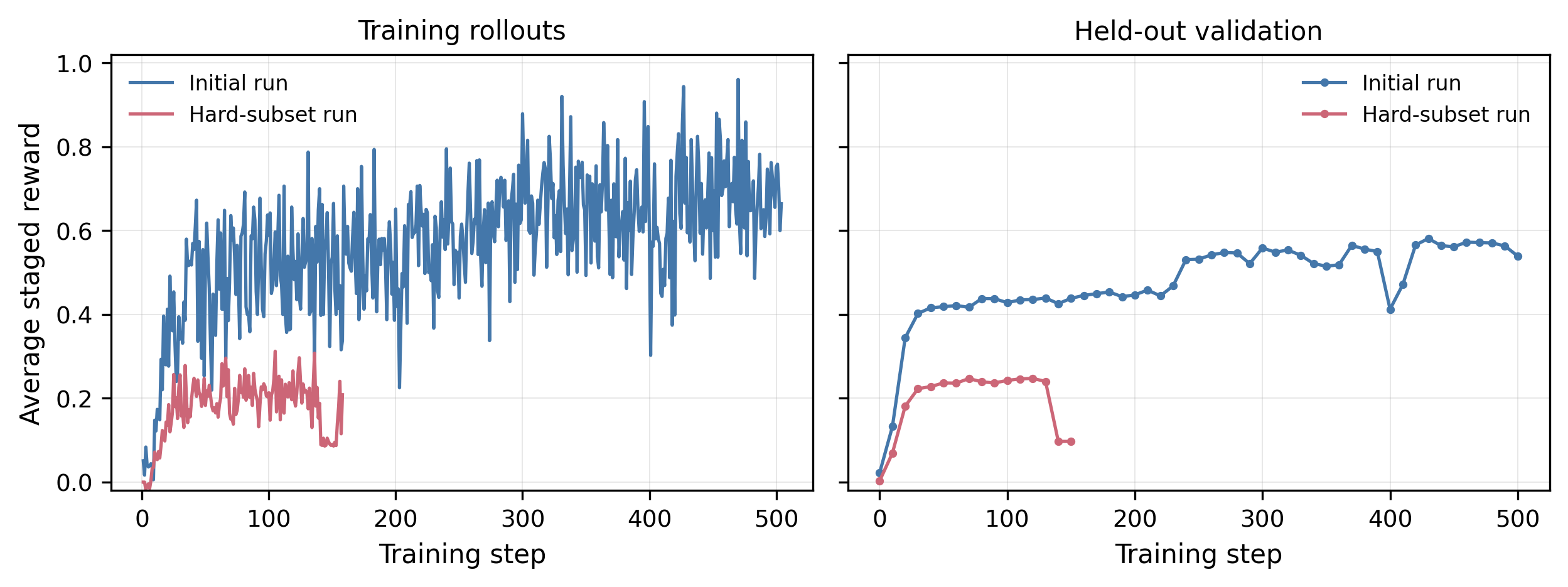}
\caption[Reward trajectories across initial and filtered Dafny RLVR runs]{Reward trajectories for the initial APPS-derived run and the
quality-filtered hard-subset run. The left panel shows average verifier reward
over sampled training rollouts, while the right panel shows held-out verifier
reward at evaluation checkpoints. The hard-subset run reaches substantially
lower rewards and plateaus, indicating that the filtering procedure
produces a more difficult verifier-feedback training distribution.}
\label{fig:reward-trajectory-overlay}
\end{figure}

\paragraph{Effect on training.}
We next repeated the checkpoint-level reward analysis from the initial run on
the quality-filtered hard subset. Figure~\ref{fig:reward-trajectory-overlay}
compares the training-rollout and held-out reward trajectories, and
Table~\ref{tab:dafny-initial-vs-hard-subset-checkpoints} summarizes the
corresponding pre-RL, best-held-out, and final checkpoints.

The hard-subset run produces substantially lower absolute rewards, consistent
with a stricter distribution that removes unfaithful, leaky, and trivial
specifications. Its held-out reward increases from 0.003 before RL to 0.247 at
step 120, but falls to 0.097 by the final checkpoint. Training reward decreases
over the same interval, from 0.197 to 0.090, so the degradation is not driven by
a widening train--held-out gap. 

Thus, the hard-subset run exhibits less apparent overfitting to sampled training
rollouts, but also fails to sustain post-RL improvement. We interpret this as
evidence that the filtered distribution provides a stricter and more diagnostic
reward signal: verifier reward is harder to obtain but now more closely tied to general Dafny
vericoding ability rather than artifacts of the initial APPS-derived setting.
This motivates further experiments with improved base models and multi-turn reward environments.

\begin{table}[!htb]
\centering
\small
\begin{tabular}{llrrrr}
\toprule
Run & Checkpoint & Step / epoch & Train & Held-out & Gap \\
\midrule
Initial run & Before RL & 0 / 0.00 & -- & 0.022 & -- \\
Initial run & Best held-out & 430 / 11.32 & 0.595 & 0.581 & 0.013 \\
Initial run & Final & 500 / 13.16 & 0.752 & 0.539 & 0.212 \\
\midrule
Hard-subset run & Before RL & 0 / 0.00 & -- & 0.003 & -- \\
Hard-subset run & Best held-out & 120 / 5.71 & 0.197 & 0.247 & -0.050 \\
Hard-subset run & Final & 150 / 7.14 & 0.090 & 0.097 & -0.007 \\
\bottomrule
\end{tabular}
\caption[Checkpoint rewards: initial APPS-derived versus filtered Dafny RLVR runs]{Checkpoint-level reward comparison between the initial APPS-derived
RLVR run and the run on the quality-filtered hard subset.}
\label{tab:dafny-initial-vs-hard-subset-checkpoints}
\end{table}

\section{Multi-Turn Training with Verifier Feedback}

\begin{figure}[ht]
    \centering
    \includegraphics[width=\linewidth]{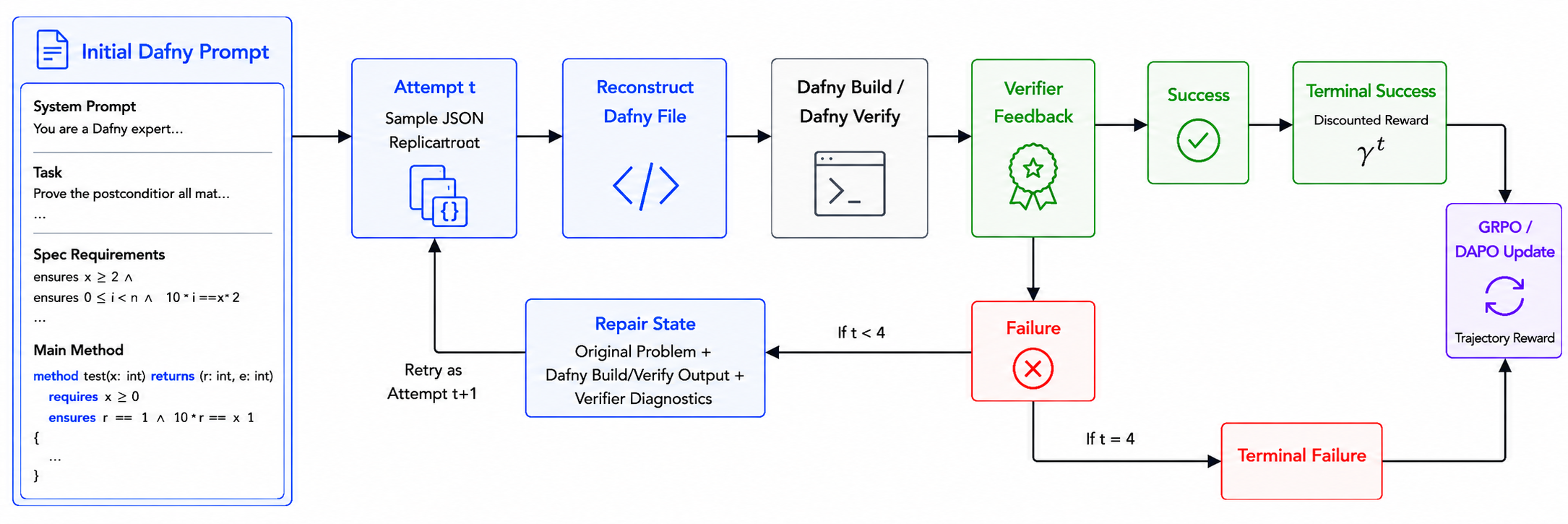}
\caption[Diagram of the multi-turn verifier-feedback Dafny RLVR environment]{
\textbf{Multi-turn verifier-feedback} environment for Dafny RLVR.
The policy samples a JSON replacement at attempt \(t\), after which the environment reconstructs the Dafny file and evaluates it with \texttt{dafny build} / \texttt{dafny verify}.
If verification succeeds, the trajectory terminates with discounted reward \(\gamma^{t-1}\); otherwise, verifier diagnostics define a local repair state for attempt \(t+1\), until either success or the \(K=4\) attempt limit is reached.
The final trajectory reward is used for the GRPO/DAPO update.
}
    \label{fig:dafny_multi_turn_pipeline}
    
\end{figure}

\paragraph{Motivation.}
Single-shot RLVR treats verifier failure as a terminal outcome. In Dafny,
however, failures are often structured: the verifier can identify failed
postconditions, assertions, invariants, type constraints, or compilation
errors. These diagnostics provide localized information about why a candidate
failed. We therefore extend the environment from single-shot synthesis to
bounded multi-turn repair, allowing the policy to condition subsequent attempts
on verifier feedback.

\paragraph{Verifier-feedback environment.}
We model Dafny repair as a finite-horizon verifier-feedback Markov decision
process with an absorbing success state. For task \(i\), let \(x_i\) denote the
original Dafny problem and let \(K=4\) be the maximum number of attempts. The
initial state \(s_{i,1}\) is the single-shot synthesis prompt constructed from
\(x_i\). At attempt \(t\), the policy samples an action
\[
a_{i,t} \sim \pi_\theta(\cdot \mid s_{i,t}),
\]
where \(a_{i,t}\) is a candidate JSON replacement. The environment reconstructs
a Dafny file \(y_{i,t}\) from this replacement and evaluates it with the Dafny
compiler and verifier. If \(y_{i,t}\) verifies, the episode transitions to a
terminal success state. Otherwise, the verifier output \(e_{i,t}\) is converted
into a new repair state
\[
s_{i,t+1} = \Phi(x_i, y_{i,t}, e_{i,t}),
\]
where \(\Phi\) constructs a retry prompt containing the original Dafny problem,
the most recent failed reconstruction, and the corresponding verifier
diagnostics.

The repair state is therefore local rather than transcript-based: it does not
accumulate the full sequence of previous attempts. This design isolates the
effect of structured verifier feedback by testing whether the policy can use
the most recent formal failure signal to produce a corrected program. Episodes
terminate after successful verification or after \(K\) attempts, so the
environment measures both eventual verification and the efficiency with which
verification is reached.

\paragraph{Reward shaping over repair attempts.}
Within this finite-horizon process, we use a discounted first-success reward.
For each task \(i\), let \(v_{i,t}\in\{0,1\}\) indicate whether the candidate
verifies on attempt \(t\in\{1,\ldots,K\}\), where \(K=4\). The first successful
attempt is $T_i = \min\{t : v_{i,t}=1\}$,
when such an attempt exists. Successful trajectories receive a discounted
terminal reward
\[
R_i^{\mathrm{succ}} = \gamma^{T_i-1}, \qquad \gamma=0.75.
\]
This objective
preserves verification as the primary criterion while making repair efficiency
part of the reward: later success remains valuable, but earlier success is
preferred.

For unsuccessful attempts, the base verifier reward is
\[
r_{\mathrm{base}}(y)=
\begin{cases}
-0.5, & \text{if } y \text{ modifies the specification or bypasses verification},\\
-0.1, & \text{if } y \text{ cannot be parsed as valid JSON},\\
0.0,  & \text{if } y \text{ produces a Dafny build failure},\\
0.1,  & \text{if } y \text{ builds but fails verification},\\
1.0,  & \text{if } y \text{ verifies successfully}.
\end{cases}
\]
Note we added a ``cheating detector" to extremely penalize the model if it changes the specification or uses bypass statements (i.e. \texttt{assume false}). For successful trajectories, the final case is replaced by the discounted
success reward \(\gamma^{T_i-1}\). Outputs that exceed the generation budget
receive an additional linear overlength penalty. If \(\ell(y)\) is the output
length, \(\ell_{\max}=5120\) is the nominal length limit, \(B=1024\) is the
overlength buffer, and \(\lambda\) is the penalty factor, then
\[
r_{\mathrm{len}}(y)
=
\min\left(
-\frac{\max(0,\ell(y)-\ell_{\max})}{B}\lambda,\,
0
\right).
\]
The total reward assigned to an attempt is therefore
\[
r(y,t)
=
\begin{cases}
\gamma^{t-1} + r_{\mathrm{len}}(y),
& \text{if } y \text{ verifies successfully on attempt } t,\\
r_{\mathrm{base}}(y) + r_{\mathrm{len}}(y),
& \text{otherwise}.
\end{cases}
\]

\paragraph{Training setup.}
The multi-turn environment required a different training architecture from the
initial single-turn RLVR run. In the initial experiment, rollouts consisted of
one generation followed by one verifier call. Here, each rollout trajectory may
contain up to four generations and four verifier calls, with intermediate retry
prompts constructed online from verifier feedback. We therefore moved from the
self-managed \texttt{verl} setup used in the initial 7B experiment to a
\textbf{Tinker-based} LoRA RL setup, which supports larger-model policy optimization
with online environment interaction.

The actor policy was initialized from
\texttt{Qwen/Qwen3-30B-A3B-Instruct-2507} and adapted with rank-16 LoRA. This
increase in model capacity is motivated by the harder policy required in the
multi-turn setting: the model must parse Dafny diagnostics, localize the failure
in its previous candidate, and produce a
valid JSON replacement that improves the reconstructed program. Thus, the
larger model and Tinker infrastructure should be viewed as enabling choices for
studying verifier-feedback RL, rather than as an isolated model-size ablation.

Training used GRPO with DAPO-style settings. Each batch contained 16 Dafny
problems, and each problem was expanded into a group of four independent
rollout environments. Within each environment, the policy could make at most
\(K=4\) attempts: one initial generation followed by up to three feedback-guided
repairs. Advantages were computed groupwise from total trajectory rewards (GRPO) and
then normalized at the token level. The objective used asymmetric clipping with lower and upper deviations
\(0.20\) and \(0.28\), corresponding to ratio bounds \([0.8,1.28]\), and no KL
penalty. The response length
budget was reduced to 6144 tokens, with the overlength penalty applied linearly
over the 5120--6144 token window. All hyperparameters are specified in Table~\ref{tab:multi_turn_rl_config}.

\begin{table}[h]
\centering
\begin{tabular}{ll}
\toprule
Setting & Value \\
\midrule
Base model & \texttt{Qwen/Qwen3-30B-A3B-Instruct-2507} \\
RL framework & Tinker LoRA RL \\
Policy Optimizer & DAPO \\
Prompt batch size & 16 \\
Rollouts per prompt & 4 \\
Max response length & 6144 \\
Clip ratio low / high & 0.20 / 0.28 \\
\midrule
LoRA rank & 16 \\
Max attempts per rollout & 4 \\
Reward scheme & Discounted first-success reward \\
Decay factor & \(\gamma=0.75\) \\
Overlength penalty window & 5120--6144 tokens \\
Constant-reward groups & Removed \\
KL penalty & 0 \\
\bottomrule
\end{tabular}
\caption{Multi-turn RL training configuration with online verifier feedback.}
\label{tab:multi_turn_rl_config}
\end{table}

\paragraph{Evaluation metrics.}
We evaluate whether this online verifier-feedback setup improves not only total
reward, but also the intermediate behaviors required for repair. In addition to mean total
trajectory reward, we
measure build success and verification success, failure-mode rates (i.e.,  JSON-format errors, Dafny build failures), and average turns per episode and token usage per turn to assess repair efficiency and the cost of multi-turn
interaction.

\paragraph{Results.}
We analyze the multi-turn run through a sequence of diagnostics. We first
examine total trajectory reward, then decompose this aggregate signal into build
and verification success, and finally inspect failure modes and interaction
efficiency.

\begin{figure}[p]
    \centering

    \begin{subfigure}[t]{0.5\linewidth}
        \centering
        \includegraphics[width=\linewidth]{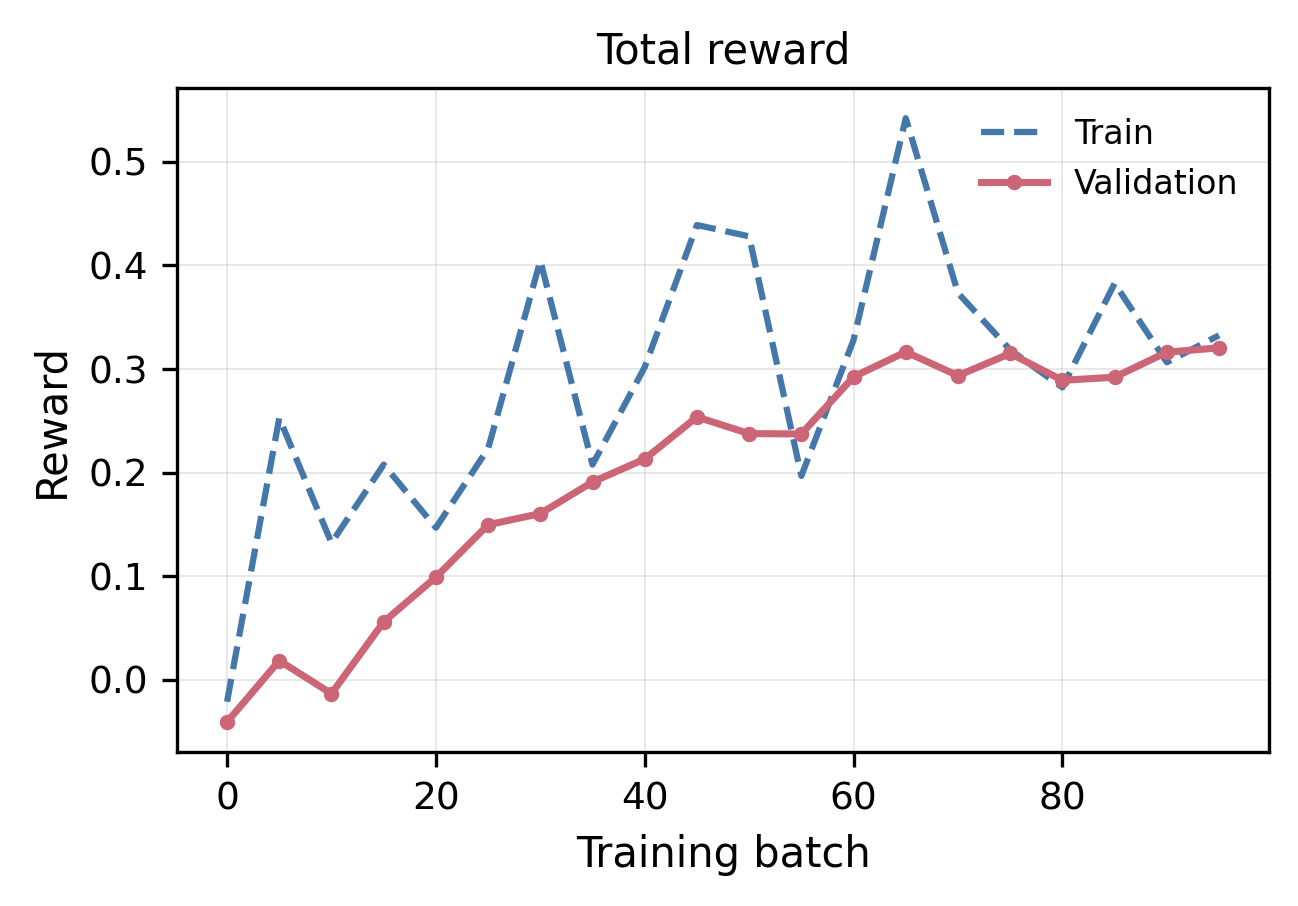}
        \label{fig:multiturn-reward-total}
    \end{subfigure}


    \begin{subfigure}[t]{0.85\linewidth}
        \centering
        \includegraphics[width=\linewidth]{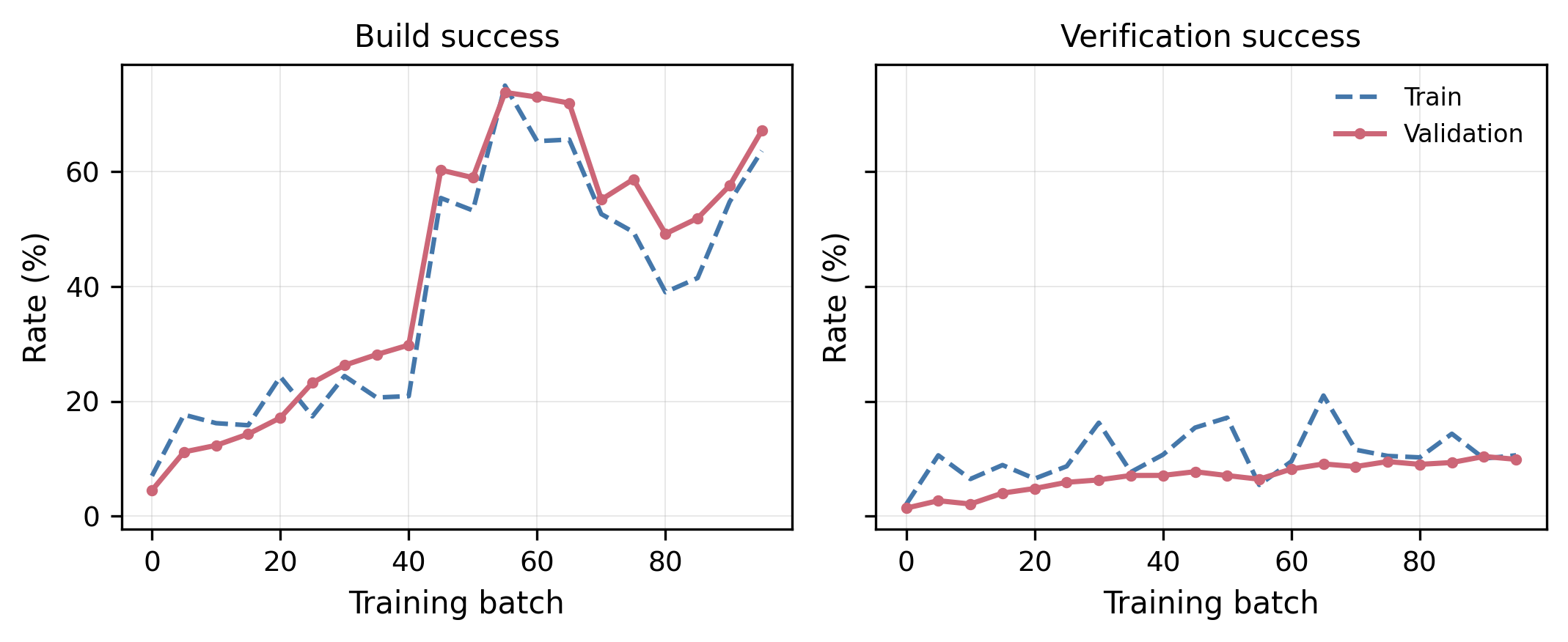}
        \label{fig:multiturn-build-verify}
    \end{subfigure}


    \begin{subfigure}[t]{0.85\linewidth}
        \centering
        \includegraphics[width=\linewidth]{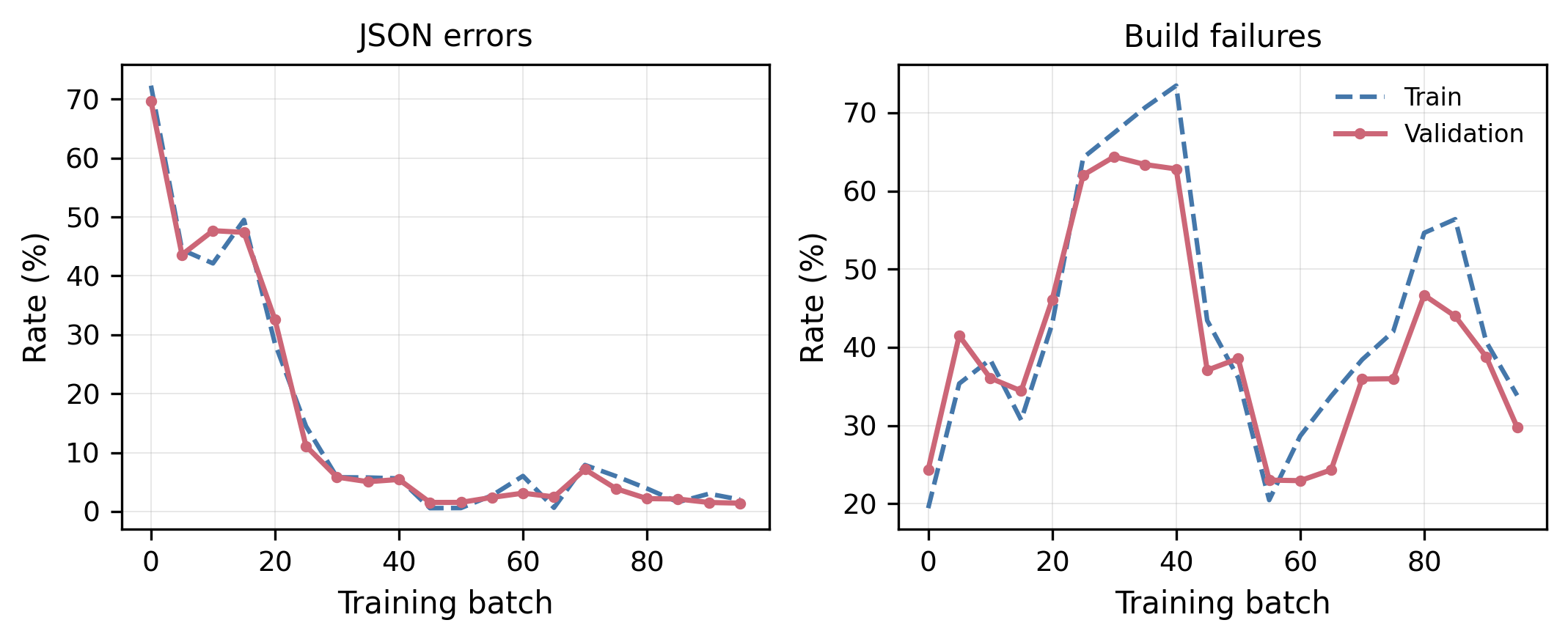}
        \label{fig:multiturn-failure-modes}
    \end{subfigure}

\caption[Training metrics for the multi-turn Dafny RLVR run]{Training diagnostics for the multi-turn Dafny RLVR run on the
quality-filtered Dafny subset. From top to
bottom, the panels show total trajectory reward, build and verification success,
and primary failure modes over training. We observe stages of JSON-format errors decline, build success improvements, and steady improvement in verification success. This suggests that the model initially prioritizes syntactically valid and structurally acceptable Dafny
artifacts. The later resurgence in build failures likely reflects a shift toward more
ambitious completions as the model searches for verification gains.}
    \label{fig:multiturn-training-diagnostics}
\end{figure}

Figure~\ref{fig:multiturn-reward-total} shows that total trajectory reward
improves over training, with held-out reward increasing to approximately
\(0.32\) by the final logged checkpoint. This indicates that training in the multi-turn verifier-feedback environment
improves the policy's ability to produce higher-reward Dafny repair trajectories.

Because the single-turn and multi-turn runs use different reward definitions,
raw reward values are not directly comparable across settings. We therefore
compare them using outcome-oriented held-out metrics. In the single-turn run,
reward@1 measures the staged verifier reward after one attempt. In the
multi-turn run, verify@{\(\leq 4\)} measures whether any attempt in the bounded
repair episode verifies.

\begin{table}[h]
\centering
\small
\begin{tabular}{lrrr}
\toprule
Checkpoint & ST reward@1 & MT proxy & MT verify@{\(\leq 4\)} \\
\midrule
Before RL & 0.003 & [0.054, 0.066] & 0.054 \\
Best validation & 0.247 & [0.337, 0.393] & 0.325 \\
Final validation & 0.097 & [0.332, 0.380] & 0.311 \\
\bottomrule
\end{tabular}
\caption[Held-out comparison of single-turn and multi-turn Dafny RLVR]{
Held-out comparison between single-turn (ST) and multi-turn (MT) Dafny RL.
The ST and MT runs use different Qwen base models: ST uses a \(7\)B model,
while MT uses a \(30\)B model. The comparison should therefore be read as a
comparison of the resulting systems, not as an isolated ablation of turn
structure alone. ST reward@1 is the one-attempt staged reward from the
single-turn run. MT proxy scores the best episode outcome within four
attempts. This is reported as a range because build-at-least-once is bounded
from aggregate turn-level build logs rather than observed directly at the
episode level. \textbf{MT verify@{\(\leq 4\)}} is the fraction of episodes
verified within the four-attempt repair budget.
}
\label{tab:single-turn-vs-multiturn-terminal-proxy}
\end{table}

Table~\ref{tab:single-turn-vs-multiturn-terminal-proxy} shows that the
multi-turn setup substantially improves held-out verification within the repair
budget, reaching \(0.325\) at the best validation checkpoint and \(0.311\) at
the final checkpoint. Since model scale, infrastructure, interaction protocol,
and reward definition all change simultaneously, this comparison should be
interpreted as evidence for the promise of online verifier-feedback training
rather than as an isolated ablation.

Because total reward aggregates several outcomes, Figure~\ref{fig:multiturn-build-verify}
separates build success from verification success. Build success increases
sharply, from roughly $5\%$ on validation at the beginning of training to
approximately $67\%$ at the final logged checkpoint. Verification success also
improves, but more modestly, rising from approximately $1$--$2\%$ initially to
approximately $10\%$ on validation. This distinction is important because much
of the reward improvement corresponds to moving generations from malformed or
non-compiling outputs into buildable Dafny programs; full proof success remains
the more difficult bottleneck.

\begin{figure}[!htp]
    \centering
    \includegraphics[width=0.6\linewidth]{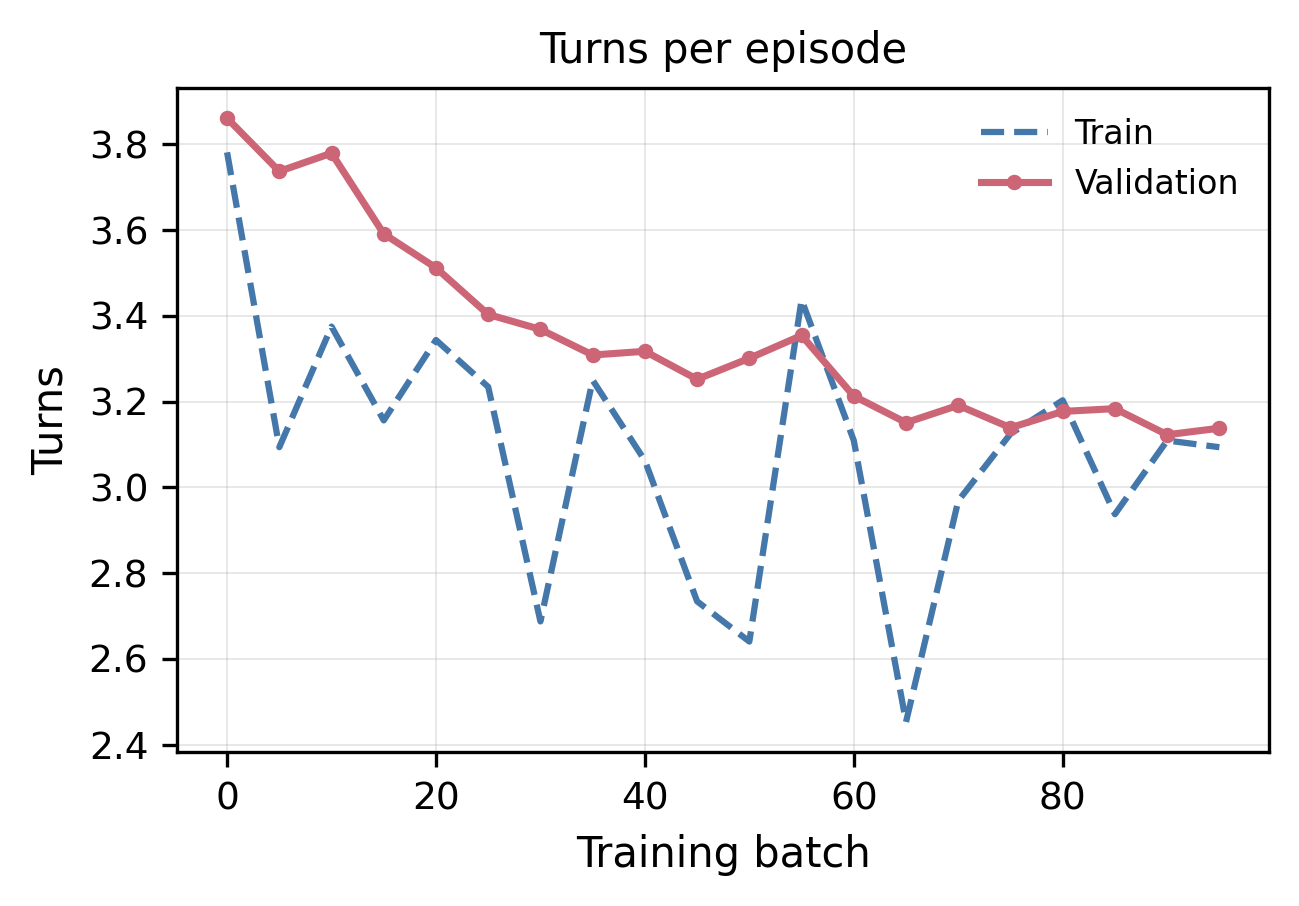}
    \caption[Average turns per episode during multi-turn Dafny RLVR]{Average number of turns per episode during the multi-turn Dafny
    RLVR run. The number of turns decreases over training, indicating that
    episodes increasingly terminate earlier as the model learns to become more efficient in response to the decaying reward}
    \label{fig:multiturn-turns}
\end{figure}

Figure~\ref{fig:multiturn-failure-modes} shows the corresponding failure-mode
decomposition. JSON-format failures decrease rapidly, falling from roughly
$70\%$ of validation turns at the start of training to approximately $1$--$2\%$
by the end. Build failures initially become more prominent as outputs begin to
pass JSON parsing, but later decline relative to their peak. This suggests that
the policy first learns the interaction protocol---returning parseable JSON
replacements---and then begins to improve Dafny-level syntactic and type
correctness.

\begin{figure}[!htp]
    \centering
    \includegraphics[width=0.85\linewidth]{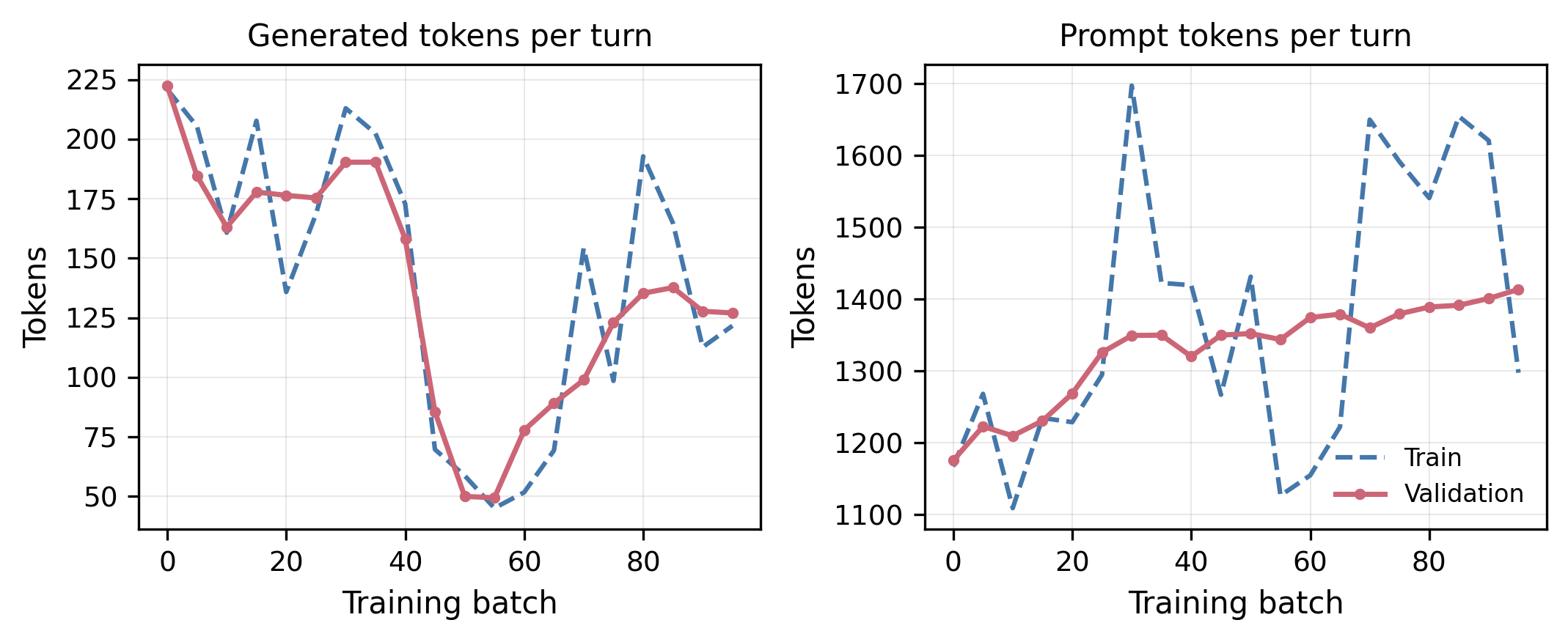}
    \caption[Token lengths during the multi-turn Dafny RLVR run]{Prompt and response token lengths during the multi-turn Dafny RLVR
    run. Generated responses become shorter over training, while prompt lengths
    increase as failed attempts and verifier diagnostics are incorporated into
    subsequent repair prompts. This reflects the changing cost structure of the
    multi-turn environment: later turns carry more context, even as the model's
    generated artifacts become more compact.}
    \label{fig:multiturn-token-lengths}
\end{figure}

Figure~\ref{fig:multiturn-turns} tracks the average number of turns per episode.
The validation average decreases from approximately $3.86$ turns to
approximately $3.14$ turns over training. Interpreted together with
Figure~\ref{fig:multiturn-build-verify}, this indicates that the improvement is
not solely due to using all available repair attempts. Together with the increase in build and verification success, this suggests
that the model increasingly uses the repair budget more efficiently rather than
simply exhausting all four attempts.

Finally, Figure~\ref{fig:multiturn-token-lengths} shows generated-token and
prompt-token lengths per turn. Generated outputs become shorter over training,
decreasing from roughly $220$ tokens per turn to roughly $125$--$130$ tokens
per turn on validation. Prompt lengths, by contrast, increase moderately,
reflecting the larger verifier-feedback prompts used in later repair turns.
This helps rule out simple length growth as the explanation for the reward
trend: performance improves while generated responses become more concise.

Overall, the multi-turn run appears to learn the verifier interaction protocol
first, then improves compilation success, and finally achieves a smaller but
meaningful gain in full verification. These results support the utility of
online compiler and verifier feedback as a training environment, while also
showing that verification remains substantially harder than producing
well-formed or buildable Dafny code.

\paragraph{Qualitative analysis.}
The rollout traces suggest that the multi-turn setup primarily improves
first-attempt behavior rather than producing reliable iterative repair.
Table~\ref{tab:multiturn-repair-trace-summary} summarizes this pattern across
a set of logged trajectories. Successful trajectories overwhelmingly verify on the first
attempt, while post-failure repair remains rare; among trajectories with a
failed first attempt, a large fraction later collapse to empty or near-empty
replacements.

  \begin{table}[!ht]
  \centering
  \small
  \begin{tabular}{lcc}
  \toprule
  Statistic & Train logs & Validation logs \\
  \midrule
  Logged trajectories & 800 & 320 \\
  Verified trajectories & 256 (32.0\%) & 90 (28.1\%) \\
  Verified on first attempt & 242 (94.5\% of verified) & 87 (96.7\% of verified) \\
  Verified after failed attempt & 14 (1.8\% of all) & 3 (0.9\% of all) \\
  Failed-first trajectories & 558 & 233 \\
  Minimal-edit collapse after failure & 293 (52.5\%) & -- \\
  \bottomrule
  \end{tabular}
\caption[Summary of logged multi-turn Dafny RLVR trajectories]{Logged multi-turn rollout summary. Most verified trajectories succeed on the first attempt, while successful
  repair after an earlier failure is rare. Minimal-edit collapse is measured only for training logs, where full generated
  text is available.}
  \label{tab:multiturn-repair-trace-summary}
  \end{table}

\noindent\textbf{Case 1: minimal-edit collapse.}
In \texttt{FindCommonType}, the first attempt produced a substantive loop-based
implementation, but Dafny could not establish the key invariant on entry:

\begin{verbatim}
invariant maxArr in array_types[0..i]
invariant forall j :: 0 <= j < i ==>
  Precedence(array_types[j]) <= Precedence(maxArr)
\end{verbatim}
Rather than strengthening the invariant or changing the loop initialization,
later attempts collapsed to empty replacements:
\begin{verbatim}
Attempt 2: ["", "{}"]  -> verification failure
Attempt 3: ["", ""]    -> build failure
Attempt 4: ["", ""]    -> build failure
\end{verbatim}
This illustrates a common failure mode: after receiving verifier feedback, the
policy often reduced the program to low-information edits instead of performing
semantic proof repair.

\noindent\textbf{Case 2: successful verifier-guided repair.}
In contrast, the \texttt{max} example shows a rare successful repair. The first
attempt was close, but the loop invariant was not strong enough to prove the
postcondition \texttt{is\_max(max, a, n)}:
\begin{verbatim}
invariant is_max(max, a, i)
invariant max == a[0] &&
  forall j :: 0 <= j < i && j < n ==> a[j] <= max
\end{verbatim}
After intermediate failed attempts, the final attempt introduced an invariant
that directly decomposed the postcondition into containment and upper-bound
facts:
\begin{verbatim}
invariant 0 < i <= n && contains(max, a, i) &&
  upper_bound(max, a, i)
invariant forall j :: 0 <= j < i ==> a[j] <= max
\end{verbatim}
This fourth attempt verified, showing that verifier feedback can occasionally
guide a semantically meaningful invariant repair, although such repairs remained
rare in this run.

\section{Discussion: Reflections on Data-Scarce RL}
\label{sec:discussion-data-scarce-rl}

The central lesson of this work is that RLVR is not reward-model free in the
strong sense. It removes the learned reward model, but it moves reward design
into the formal artifact. In Dafny, the verifier deterministically checks
whether the assembled program satisfies the specification presented to it:
\[
    \mathcal{V}_{\mathrm{Dafny}}(\operatorname{Assemble}(x,y))
    =
    \texttt{verified}.
\]
This is a clean machine-checkable signal, but it is only a signal for
\(\varphi_{\mathrm{given}}\), not necessarily for the intended task
\(\varphi_{\mathrm{intended}}\). The gap between those two specifications is
where much of the difficulty in this chapter appears.

Table~\ref{tab:part-i-summary-with-vericoding} summarizes the main results.
The initial RLVR run reached \(53.9\%\) held-out reward on the APPS-derived
validation split, showing that Dafny verifier feedback is strong enough to
optimize an LLM policy. However, Section~\ref{sec:spec-hacking} showed that
this number overstated genuine synthesis progress: weak postconditions, leaked
helper functions, and staged build rewards created cheap reward channels. After
filtering toward faithful, low-leakage, nontrivial specifications in
Section~\ref{sec:data-filtering}, single-turn RLVR became much harder and ended
at \(9.7\%\) held-out reward. The multi-turn verifier-feedback setup recovered
to \(31.1\%\) verify@{\(\leq 4\)} on the hard subset, suggesting that online
compiler and verifier feedback can partially compensate for the sparsity of the
cleaner reward signal.

\begin{table}[ht]
\centering
\small
\begin{tabular}{lrrr}
\toprule
Method & Hard subset & APPStest & Overall \\
\midrule
Initial RLVR & -- & 53.9\% (val) & -- \\
Filtered RLVR & 9.7\% (val) & -- & -- \\
Multi-turn RLVR & 31.1\% (val) & -- & -- \\
\midrule
Gemini 2.5 Flash & -- & 36.9\% (all) & 38.2\% (all) \\
GPT-5 mini & -- & 71.6\% (all) & 66.9\% (all) \\
Claude Opus 4.1 & -- & 66.6\% (all) & 67.5\% (all) \\
Model union & -- & 83.0\% (all) & 82.2\% (all) \\
\bottomrule
\end{tabular}
\caption[Main Part I summary of Dafny RLVR results with Vericoding references]{Part I verifier-success summary for our Dafny RLVR runs, with external
Vericoding reference points for context. The RLVR rows report held-out
validation results from our experiments: the initial single-turn run on the
APPS-derived validation set, the filtered single-turn run on the
quality-filtered Dafny hard subset, and the multi-turn run on the same hard
subset. The Gemini, GPT, Claude, and model-union rows report Dafny pass rates
from Vericoding on the APPS test source and the overall Dafny benchmark
aggregate. Values marked \emph{val} are our validation results; values marked
\emph{all} are Vericoding-reported benchmark results. External rows use five
attempts per task, except ten attempts on BigNum
\citep{10.48550/arxiv.2509.22908}, and should be read as contextual reference
points rather than controlled baselines.}
\label{tab:part-i-summary-with-vericoding}
\end{table}

The sharper interpretation is that the data distribution is part of the reward
function. In GRPO-style training, a prompt provides useful group-relative signal
only when its sampled completions receive different rewards. For binary verifier
rewards, if each completion for prompt \(x\) succeeds independently with
probability \(p_x\), then a group of size \(K\) is non-degenerate with
probability $1 - p_x^K - (1-p_x)^K$.
This probability is small when the task is either too easy or too hard. Thus,
data-scarce RLVR depends on prompts near the model's capability frontier: tasks
where some completions verify and others fail. The initial APPS-derived
distribution contained many easy or exploitable prompts, so reward improved
quickly. Filtering removed many of these cheap reward channels, making the
signal more faithful but also sparser. In short, filtering improved reward
validity while worsening RL credit assignment.

The multi-turn environment partially relaxes this sparsity. Rather than making
Dafny failure terminal, it exposes verifier diagnostics and allows bounded
repair. This gives the policy intermediate behaviors to learn: producing valid
JSON, generating buildable Dafny, reaching verification, and doing so in fewer
attempts. The training dynamics follow this progression: JSON errors collapse
first, build success rises next, and verification improves more slowly. The
multi-turn run therefore succeeds as an online verifier-feedback environment,
but not yet as a robust repair learner. Most verified trajectories still succeed
on the first attempt, while success after an initial failure remains rare. The
model actually becomes a stronger first-shot Dafny generator before it becomes a reliable
proof-repair agent.

\subsubsection*{Limitations}

\paragraph{Small effective RL dataset.}
The filtered hard subset contains only 1,149 tasks, with 919 training examples.
This is small by GRPO-style reasoning standards: DAPO, for instance, trains on
17K math prompts and samples 512 prompts \(\times\) 16 responses per rollout
step \citep{yu2025dapo}. More importantly, GRPO does not learn from prompts
uniformly. Since advantages are computed relative to other samples from the same
prompt, a prompt is informative only when its sampled completions receive
different rewards. All-failure and all-success groups contribute little useful
signal. Thus, the effective RL dataset is smaller than the prompt count
suggests: GRPO learns mainly from tasks where the current policy is
inconsistent, not from tasks it always fails or always solves.

\paragraph{Reward brittleness.}
Staged reward made the initial experiment trainable by giving partial credit for
buildable programs, but it also encouraged build-oriented templates. Raw
verification is similarly brittle: a program can verify because the
specification is weak, a helper function leaks the answer, or the completion
changes the effective proof obligation. The anti-cheating filters catch obvious
violations such as modified contracts, \texttt{assume false}, and
\verb|{:verify false}|, but they cannot fully detect semantic underspecification
or specification leakage.

\paragraph{Experimental confounding.}
The multi-turn experiment changes several variables at once: model scale,
infrastructure, LoRA adaptation, interaction protocol, reward definition, and
attempt budget. The gain from \(9.7\%\) to \(31.1\%\) is therefore evidence for
the promise of the overall online verifier-feedback setup, not a clean causal
estimate of multi-turn feedback alone.

\paragraph{Weak post-failure repair.}
The current multi-turn policy often learns the protocol and improves initial
generations, but after a failed proof it may collapse to empty or
low-information edits rather than preserving useful code and strengthening the
relevant invariant or assertion. This distinction is important:
verify@{\(\leq 4\)} measures bounded opportunity, not necessarily repair
competence.

\subsubsection*{Future Work}

\paragraph{Tool-using verifier agents.}
The most direct next step is to turn the policy from a generator into a
tool-using verifier agent. The current environment gives Dafny feedback after a
candidate is produced, but the model does not actively decide when to call
tools, inspect diagnostics, retrieve lemmas, or run intermediate checks. Future
work should make these actions part of the policy interface, closer to
reasoning-and-acting agents such as ReAct and Toolformer
\citep{10.48550/arxiv.2210.03629, lean_lsp_mcp}, but with formal tools: compiler,
verifier, local context retrieval, lemma search, and proof-obligation
inspection.

\paragraph{Repair-specific supervision.}
Terminal verification reward is too coarse to teach targeted proof repair.
Future systems should mine or generate failed-to-verified trajectories of the
form
\[
    (x, y_{\mathrm{fail}}, e_{\mathrm{dafny}})
    \mapsto
    y_{\mathrm{repair}},
\]
where \(e_{\mathrm{dafny}}\) is the compiler or verifier diagnostic. These
trajectories would teach the model to preserve useful partial solutions,
localize the failed proof obligation, and make minimal semantic edits. This
connects to self-refinement methods such as Self-Refine and Reflexion
\citep{madaan2023selfrefine,shinn2023reflexion}, but replaces language-model
critique with grounded verifier feedback.

\paragraph{Curriculum learning for verifier feedback.}
Future work should also treat task selection as part of the training algorithm.
The hard-subset experiments show that filtering improves reward faithfulness but
also makes the learning signal sparser. A natural next step is curriculum
learning over verifier difficulty: begin with tasks where the base model has a
nonzero probability of success, then gradually shift toward harder prompts as
the policy improves \citep{bengio_curriculum_2009}. For group-relative methods,
this is especially important because useful gradients arise primarily from
prompts whose sampled completions produce mixed outcomes rather than uniformly
failing or uniformly succeeding. A verifier-aware curriculum could therefore
sample frontier tasks whose empirical success rate lies between these extremes,
while continuing to exclude tasks with weak, leaky, or unfaithful specifications.
This makes the proposal closer to adaptive or teacher-student curriculum
learning, where training examples are selected based on the learner's current
progress, rather than a fixed easy-to-hard ordering
\citep{matiisen_teacher-student_2017}.

\paragraph{Faithful data scaling.}
Dafny data scarcity is structural, so future work should combine human
solutions, teacher-model rollouts, synthetic tasks, verified repairs, and
rejection sampling. The key is not simply more verified programs, but more
faithful specifications and more nontrivial repair states. STaR-style
bootstrapping and formal synthetic-data pipelines such as DeepSeek-Prover
suggest one route \citep{zelikman2022star,xin_deepseek-prover_2024}, but Dafny
adds an extra requirement: synthetic tasks must be filtered for specification
faithfulness in translation, not just verifier success.

\paragraph{Cross-verifier transfer.}
Future work should also evaluate domain transfer across verifiers. Dafny, Lean, Verus,
and unit-tested programming tasks all provide verifiable rewards. Multi-language RLVR could test
whether verifier feedback teaches general repair behavior or only
environment-specific syntax and reward hacks. Recent RLVR systems show that verifiable rewards can scale reasoning, but
mostly within fixed domains \citep{10.48550/arxiv.2504.21801}\citep{10.48550/arxiv.2501.12948} \citep{10.48550/arxiv.2508.03613}. The open question is whether these gains transfer
across formal systems, or remain tied to each environment's syntax, tools, and
reward channels.

In summary, the next step is not simply more RL. This chapter shows that
verifier feedback can train Dafny-capable policies, but it also shows that RL
faithfully amplifies whatever objective the verifier exposes. Weak
specifications produce reward hacking; filtered specifications produce a cleaner
but sparser learning problem; and multi-turn feedback improves interaction
without yet guaranteeing genuine repair. Data-scarce RLVR, therefore, depends on the entire training environment: the specification must be unambiguous, the tasks must elicit informative failure modes, the verifier must return feedback the model can act on, and evaluation must distinguish first-shot success from genuine post-failure repair. The central challenge is not obtaining a machine-checkable reward, but making that reward faithful enough to optimize against and informative enough to learn from.
\chapter{Part II: Agentic Scaffolding}
\label{ch:part2-agentic-scaffolding}

\section{Preliminaries: Agentic Systems for Formal Methods}

Part~I used verifier feedback to improve the model. Although that setting
included multi-turn verifier-feedback episodes, the interaction remained
model-centric: verifier responses were incorporated through a fixed feedback
protocol, and the learning objective was to update the policy
\(\pi_\theta\). This chapter studies the complementary system-centric setting.
Here, verifier feedback is used directly by an inference-time scaffold that
maintains state, invokes tools, and controls search over candidate Lean
artifacts. The distinction is therefore not single-turn versus multi-turn
interaction, but training a policy from verifier feedback versus using
verifier feedback as an online observation for search.

This shift is natural for Lean verified coding. Candidate solutions must pass
elaboration, type checking, tactic execution, and kernel checking, yielding both
a precise success criterion and structured intermediate signals. Failed
attempts expose goals, local context, type errors, missing facts, and tactic
failures. Part~II uses these signals not merely to repair a single trajectory,
but to guide stateful inference: branching, retrieval, decomposition, and
termination.

\paragraph{From direct generation to scaffolded search.}
The simplest inference mode is \emph{one-shot generation}. The model receives a
Lean verified-coding task and emits a complete candidate solution, which is
then accepted or rejected by Lean. This baseline is attractive because it is
simple, cheap, and easy to evaluate. The next strongest baseline is \emph{best-of-\(n\)} sampling. The model samples
multiple independent candidates, and the system keeps the first candidate that
verifies. Best-of-\(n\) uses Lean as a selection oracle and can improve success
when correct solutions are already present in the model's sampling
distribution. However, each attempt remains independent.

A more interactive strategy is \emph{iterative repair}. Here the system feeds a
failed candidate and its Lean diagnostics back to the model, asks for a revised
candidate, and repeats this loop. This mirrors a common pattern in code and
language-agent systems: generate an artifact, observe feedback from an external
tool, and revise the artifact in response
\citep{10.48550/arxiv.2210.03629}\citep{shinn2023reflexion,madaan2023selfrefine,chen2023selfdebug,
yang2024sweagent}. For formal methods, repair is especially appealing because
the feedback comes from the same verifier that defines success. Nevertheless,
iterative repair remains limited. It can become trapped in local edit cycles,
repair superficial errors while preserving a bad proof plan, or spend its
entire budget on a single trajectory that should have been abandoned earlier.

The main mode considered in this chapter is a \emph{hierarchical scaffold}.
Rather than asking a single model to produce a complete Lean artifact in one
pass, the system decomposes a hard verification task into smaller subgoals and
delegates those subgoals to worker model calls. Successful subproofs are then
recombined into the parent artifact and checked again by Lean. In this view,
hierarchy is primarily a mechanism for exposing intermediate proof structure:
the scaffold turns one difficult theorem or verified-coding task into a tree of
smaller obligations that can be solved, repaired, and verified locally.

These inference modes form a progression:
\[
\text{one-shot}
\;\rightarrow\;
\text{best-of-}n
\;\rightarrow\;
\text{iterative repair}
\;\rightarrow\;
\text{hierarchical subgoal decomposition}.
\]

The hierarchical scaffold is best viewed as a generalization of simpler inference
strategies: it still uses generation, verification, and repair, but organizes
them around recursively proposed subgoals rather than a single monolithic proof
attempt.

\paragraph{Agentic scaffolds.}
In this chapter, an \emph{agentic scaffold} is an inference-time control system
wrapped around a base language model. Rather than treating the model as a
standalone solver, the scaffold defines the interface through which model calls
interact with the formal environment. Concretely, it specifies five components:
the state maintained across attempts, the model roles available to the system,
the tools exposed to those roles, the actions they may take, and the controller
that executes those actions.

The scaffold can be described through five interfaces:
The scaffold can be described through five interfaces:
\begin{enumerate}
    \item \textbf{State:} the evolving proof context, including the original
    Lean task, the current candidate file, previous attempts, verifier
    diagnostics, local proof states, retrieved facts, and proposed intermediate
    claims.

    \item \textbf{Roles:} the model-level division between global planning and
    local solving. A planner proposes proof strategies or decomposes hard
    obligations into subgoals, while worker models attempt to solve those
    subgoals.

    \item \textbf{Tools:} the interface to Lean and the surrounding repository,
    including candidate checking, diagnostic extraction, proof-state inspection,
    and retrieval over relevant lemmas or examples.

    \item \textbf{Actions:} the operations available to the scaffold, including
    generation, repair, retrieval, subgoal proposal, and recombination of solved
    subproofs into the parent artifact.

    \item \textbf{Controller:} the policy that selects which action to execute
    next and determines when a branch or proof attempt should terminate.
\end{enumerate}

This definition emphasizes that the scaffold is not merely a multi-turn prompt.
It is a structured inference procedure over formal artifacts. Its central
mechanism in this chapter is hierarchical decomposition: a hard Lean
verified-coding task is broken into smaller obligations, delegated to worker
model calls, locally checked or repaired, and then recombined into a final Lean
artifact. Repair, retrieval, and verification are therefore supporting
operations inside a broader decomposition-driven search process.

This framing also clarifies the relationship between Part~I and Part~II. In
Part~I, verifier feedback was used primarily to improve the policy
\(\pi_\theta\). In Part~II, the policy is held inside an explicit scaffold, and
verifier feedback becomes an online observation used to guide decomposition,
local solving, repair, and recombination.

\paragraph{Why Lean is a natural setting.}
Lean is particularly well suited to agentic scaffolding because it provides
both a rigorous success criterion and structured intermediate feedback. A
completed proof either checks, or it does not. At the same time, Lean exposes
information that can guide future attempts: elaboration errors, tactic failures,
unsolved goals, local hypotheses, type information, and dependencies on
imported or local facts.

Lean verified-coding tasks also make the limits of one-shot generation visible.
The model may need to synthesize definitions, reason about program behavior,
prove auxiliary lemmas, and sequence tactics that interact correctly with local
definitions and imported libraries. A small implementation choice can simplify
or complicate the proof obligation. A proof may fail not because the overall
strategy is wrong, but because the model used the wrong lemma name, chose the
wrong induction variable, failed to expose a needed invariant, or attempted to
close a goal before establishing an intermediate fact. These are precisely the
kinds of failures that can benefit from tool use, retrieval, decomposition, and
stateful repair.

The dependence on library and local context is especially important. Many Lean
proofs are short once the right fact or premise is found, but difficult when the model must
infer that fact from the theorem statement alone. Conversely, some proof
a single retrieved lemma does not solve all obligations; they require a proof
plan that introduces intermediate claims and combines them in the right order.
This gives a natural role to hierarchical search. A hard Lean proof may be
inaccessible with a single generation, but accessible if the system first
proposes subgoals, proves them separately, retrieves supporting facts, and then
assembles the final artifact under verifier feedback.

\paragraph{Relation to prior agentic systems.}
The scaffold studied in this chapter sits at the intersection of several
threads of prior work. General language-agent systems introduced the basic
reason--act pattern, where a model interleaves reasoning with tool calls and
uses observations to condition later steps \citep{10.48550/arxiv.2210.03629}. Reflection and
self-repair systems showed how feedback from previous attempts can be stored or
reintroduced to improve future generations
\citep{shinn2023reflexion,madaan2023selfrefine,chen2023selfdebug}. Code-agent
systems extend this idea with explicit interfaces to files, editors, linters,
test runners, and search tools, showing that the design of the
agent--computer interface can be as important as the base model
\citep{yang2024sweagent,ridnik2024alphacodium}.

Formal theorem proving specializes these ideas to a verifier-rich environment.
LeanDojo and ReProver made interaction with Lean and retrieval-augmented theorem
proving a reproducible object of study, emphasizing that access to the proof
environment and relevant premises is a central bottleneck
\citep{10.48550/arxiv.2306.15626}. COPRA showed that a general-purpose model can be used
inside a stateful, backtracking proof-search loop in Lean and Coq, with proof
states, retrieved lemmas, and execution feedback informing later actions
\citep{thakur2024copra}. These systems already point toward the perspective
adopted here: the model is embedded in a loop that observes and acts within a
formal environment.

A second line of work studies inference-time scaling and decomposition.
Best-of-\(n\) and verifier-reranking methods demonstrate that additional
test-time compute can improve success when the verifier can identify correct
outputs, but flat sampling wastes compute on independent attempts. More
structured approaches allocate compute adaptively, score partial progress, or
search over intermediate states \citep{lightman2023letsverify,snell2024testtime}.
For Lean, decomposition has become especially important. Draft-Sketch-Prove
uses informal structure to guide formal proof construction \citep{jiang2023dsp};
DeepSeek-Prover-V2 uses recursive subgoal decomposition as part of its formal
reasoning pipeline \citep{10.48550/arxiv.2504.21801}; and Hilbert combines informal
reasoning, Lean verification, retrieval, and recursive decomposition to build
formal proofs from smaller subproblems \citep{10.48550/arxiv.2509.22819}. These
systems motivate the view that hard formal tasks are often more naturally
solved by constructing and checking a proof plan than by sampling a monolithic
final proof.

Recent Lean agents reinforce the same trend at the system level. 
Numina-Lean-Agent treats a general coding agent as the outer reasoning loop,
using a Lean MCP layer to support autonomous interaction with Lean, theorem
retrieval, informal reasoning, and auxiliary proof tools
\citep{10.48550/arxiv.2601.14027}. OpenGauss similarly emphasizes the
infrastructure around formalization: project detection, backend management,
workflow spawning, and coordinated agent execution over Lean projects
\citep{noauthor_math-incopengauss_2026}. Ax-Prover frames Lean proving as a multi-agent
workflow, combining LLM-based reasoning with MCP-based verification tools for
iterative formal proof construction \citep{requena2025axprover}. 
AxiomProver pushes this direction toward autonomous ensemble proving, where
multiple agents search for Lean proofs under verifier-backed checking
\citep{noauthor_axiommathputnam2025_2026}.

These systems differ in implementation, but they share a common design lesson:
state-of-the-art Lean proving increasingly depends on the system wrapped around
the model. Tool access, project-level context, retrieval, proof-state
inspection, decomposition, and verifier-backed iteration become part of the
inference algorithm itself, rather than post-hoc additions to direct proof
generation.

\paragraph{Benchmark motivation.}
The empirical sections of this chapter instantiate this perspective on Lean
verified-coding benchmarks introduced in Section~\ref{ch:benchmarks}, including Vericoding Lean tasks and the Dalek benchmark introduced earlier in
Section~\ref{sec:2.3.3}. These benchmarks are useful stress tests for
agentic scaffolding because they are not merely shallow syntax-completion
problems. They require proof construction, local reasoning about program
behavior, auxiliary facts, and interaction with Lean's verifier. They also
exhibit low base-model one-shot success, making them a natural setting for
asking whether additional inference-time structure can recover solutions that
direct generation misses.

\paragraph{Chapter contribution.}
While recent Lean agents have shown the value of tool use, retrieval, and
verifier-backed iteration for theorem proving, Lean verified coding introduces
a different inference problem. The system must synthesize artifacts that combine
program definitions, specifications, auxiliary lemmas, and proofs, where changes
to the implementation can alter the proof obligations that must later be
discharged. Decomposition is therefore not just a way to split a theorem into
mathematical subgoals; it must produce these subgoals that can be solved by worker
models and then recombined into a coherent Lean file that still verifies as a
whole.

This chapter studies that setting through three system components. First, we
use hierarchical multi-model decomposition: a planner proposes intermediate
subgoals, worker models attempt local proofs, and solved subproofs are inserted
back into the parent artifact. Second, we expose Lean through a structured MCP
interface that supports checking, diagnostic parsing, proof-state inspection,
retrieval, and guardrails against invalid shortcuts. Third, we use progress
evaluation to decide when a branch is making useful progress, when it should be
repaired, and when it should be abandoned. Together, these components instantiate
a verifier-guided scaffold for Lean verified coding, rather than a direct
generation or linear repair loop.

\section{System Overview}
\label{sec:system-overview}

We instantiate the verifier-guided scaffold described above as a two-level
system for Lean verified coding. The outer level first adapts the scaffold to the
structure of a VeriCoding task: it constructs and revises a sectioned Lean file
containing helper, definition, and theorem regions. The inner level is a
recursive proof-search engine, which we call \textsc{RecursiveWorker}: given a
theorem or lemma obligation, it attempts to prove the obligation directly,
decomposes difficult goals into intermediate subgoals, delegates those subgoals
to worker model calls, and recombines verified subproofs into the parent
artifact. This is directly inspired by \texttt{HILBERT} which uses recursive decomposition for theorem proving \citep{10.48550/arxiv.2509.22819}.

The system is verifier-in-the-loop at both levels. Candidate helper,
definition, and proof regions are checked against Lean as the artifact is
constructed, while theorem and lemma obligations are repeatedly checked during
recursive decomposition and recombination. A final successful output is a complete Lean
artifact that passes compilation and a no-bypass verification check: theorem statements must be
preserved, all proof obligations must be discharged, and the final file must
contain no remaining proof holes or unsound shortcuts.

The architecture therefore separates two concerns. The task-level adapter makes
the proof-search procedure compatible with verified-coding benchmarks, where
proof obligations are coupled to generated definitions and auxiliary facts.
\textsc{RecursiveWorker} is the central inference mechanism: it treats a hard
Lean obligation not as a single generation problem, but as a recursively
structured search over smaller verified obligations.

In more detail, the scaffold can be viewed as four interacting layers:
\begin{description}
    \item[Task adapter.] The task adapter manages the global Lean artifact. It
    tracks the editable helper, definition, and theorem regions; constructs
    candidate files; and identifies unresolved theorem or lemma obligations
    that require proof-level search.

    \item[RecursiveWorker.] The proof-search layer operates on individual
    theorem or lemma obligations. It is responsible for direct proof attempts,
    recursive subgoal decomposition, local worker calls, and recombination of
    verified subproofs into the parent artifact.

    \item[Verifier and tool layer.] Lean provides the authoritative
    accept/reject signal. In addition, an optional MCP tool layer exposes
    structured interfaces for diagnostics, proof-state inspection, retrieval,
    and bounded candidate checks. These tools guide search, but do not replace
    final verification.

    \item[Progress and guardrail layer.] The scaffold applies deterministic
    checks and progress evaluation at candidate boundaries. This layer enforces
    output format, preserves theorem signatures, rejects proof bypasses, and
    decides whether the system should continue, repair, decompose, promote, or
    stop.
\end{description}

\begin{figure}[ht]
    \centering
    \includegraphics[width=0.75\textwidth]{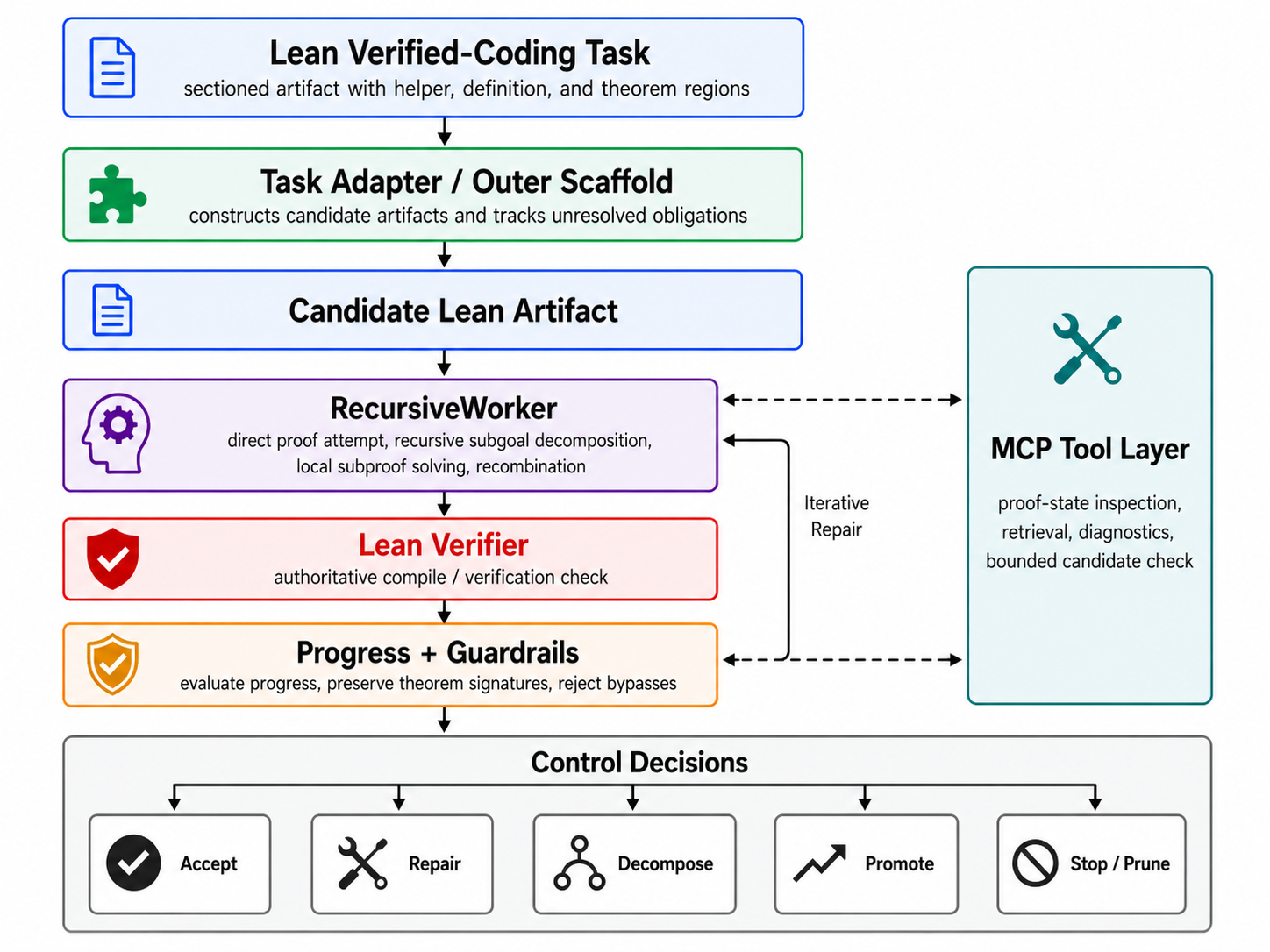}
    \caption[Overview of the verifier-in-the-loop Lean scaffold]{Overview of the verifier-in-the-loop scaffold for Lean verified
    coding. The task adapter manages sectioned VeriCoding artifacts, while
    \textsc{RecursiveWorker} performs recursive proof search over theorem and
    lemma obligations. Lean verification is the authoritative acceptance gate;
    the MCP tool layer provides auxiliary inspection and retrieval support.}
    \label{fig:system-overview}
\end{figure}

Figure~\ref{fig:system-overview} summarizes the overall architecture. The main
design principle is that Lean verified coding is treated as adaptive proof
search over a growing artifact, rather than as a single generation problem.
Candidate files are checked repeatedly, and verifier feedback is routed either
to the task-level adapter or to \textsc{RecursiveWorker}, depending on whether
the failure concerns the global artifact or a local proof obligation.

The following subsections unpack these layers. Section~\ref{sec:recursive-decomposition}
describes \textsc{RecursiveWorker}; Section~\ref{sec:verification-tool-access}
describes the verifier boundary and MCP tool layer; and
Section~\ref{sec:progress-evaluation} describes the progress and guardrail
mechanisms used to control search.

\subsection{Recursive Decomposition}
\label{sec:recursive-decomposition}

\begin{figure}[ht]
    \centering
\includegraphics[width=0.8\linewidth]{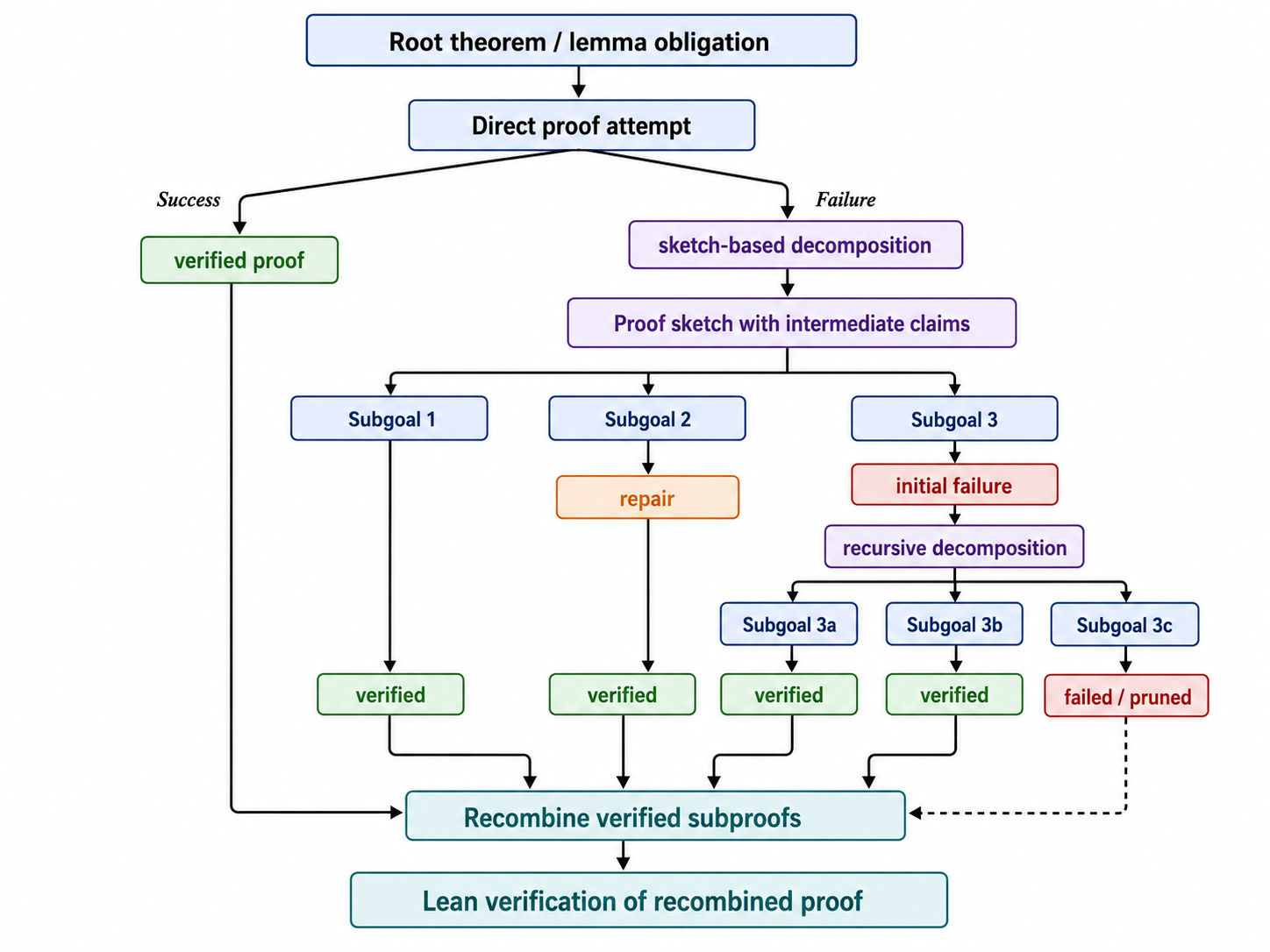}
    \caption[Diagram of recursive proof decomposition for Lean verified coding]{
\textbf{Recursive proof decomposition} for Lean verified coding.
Given a root theorem or lemma obligation, the scaffold first attempts a direct proof.
If the direct attempt succeeds, the resulting proof can be accepted as a verified proof.
If it fails, the system constructs a proof sketch containing named intermediate claims, which are promoted into standalone subgoals.
Each subgoal is then solved independently: some subgoals verify directly, some require repair, and harder subgoals may be recursively decomposed into smaller obligations.
Verified leaves are collected as reusable subproofs, while unsuccessful branches may be pruned.
The verified subproofs are then recombined into a candidate proof of the parent obligation, which is finally checked by Lean to ensure that the recombined proof satisfies the original theorem statement.
}
    \label{fig:recursive_decomp}
\end{figure}

The core proof-search mechanism is recursive decomposition. Rather than asking
one model call to synthesize a complete proof of a difficult Lean obligation,
the system treats the obligation as a target that can be broken into smaller
verified claims. This is useful because long Lean proofs often fail for local
reasons: a missing intermediate fact, an incorrect tactic sequence, or a proof
state that requires an auxiliary lemma before the final goal can be closed.
Recursive decomposition makes this intermediate structure explicit.

\paragraph{Sketch-based decomposition.} The procedure begins with a direct proof attempt. Given a theorem or lemma
obligation, a worker first asks a formal prover model to generate a complete
Lean proof. If this candidate verifies, the obligation is solved immediately.
If it fails, the system switches to sketch-based decomposition. A reasoning
model proposes a proof sketch containing named intermediate claims, typically
expressed as \texttt{have} statements whose bodies are left unresolved. These
claims serve as candidate subgoals: instead of solving the original theorem
monolithically, the system extracts the intermediate claims as standalone proof
obligations.

\paragraph{Worker and sub-worker roles.} Each extracted subgoal is then assigned to worker model calls. A worker may try
to solve the subgoal directly, repair a failed subgoal proof using Lean
diagnostics, or invoke the same decomposition procedure recursively if the
subgoal remains too difficult. Thus the proof attempt induces a tree of
obligations. The root is the original theorem, internal nodes are obligations
that have been decomposed further, and leaves are either verified subproofs or
failed branches. Recursion is bounded by depth limits and proof-attempt budgets,
so decomposition does not become an
unbounded proof exploration.

\paragraph{Recombination.} This is done explicitly. Once all required subgoals from a proof sketch have
verified, their proofs are inserted back into the parent proof context. The
system then reconstructs the parent theorem using the verified subproofs and
reruns Lean on the combined artifact. This final check is necessary because
subgoals that verify in isolation must still compose correctly with the parent
proof, local names, imports, and surrounding definitions. A branch is accepted
only when the recombined theorem verifies in context.

This \textbf{theorem-level} decomposition is distinct from the \textbf{task-level} artifact
management described in the previous section. The task adapter may construct or
revise helper, definition, and theorem regions of a VeriCoding file, but
\textsc{RecursiveWorker} is applied only to theorem and lemma obligations. In
other words, recursive decomposition is a proof-search mechanism: it solves
formal obligations by introducing, proving, and recombining intermediate claims,
while the outer scaffold is responsible for inserting the resulting proof back
into the full verified-coding artifact.

\paragraph{Termination.} The procedure terminates when the root theorem verifies, when a required subgoal
cannot be proved within the configured attempt budget, when the maximum
recursion depth is reached, or when a global resource limit is exhausted. These
termination conditions ensure that recursive decomposition can exploit
additional inference-time structure without allowing failed branches to consume
unbounded compute.

\subsection{Verification MCP and Tool Access}
\label{sec:verification-tool-access}
\paragraph{Verifier boundary.}
Lean verification is the authoritative accept/reject path. A candidate artifact
is accepted only if the required declarations elaborate and all proof
obligations are discharged under the final no-bypass policy. The tool layer has
a different role: it provides additional information to guide model calls.
Tools may expose diagnostics, proof states, library search, or bounded candidate
checks, but their outputs are still never acceptance evidence.

\paragraph{MCP tool layer.}
We implement tool access through an MCP-style tool layer. MCP, or Model Context
Protocol, provides a structured interface through which a model can call
external tools rather than relying only on raw prompt text \citep{lean_lsp_mcp}.
In practice, a tool-calling model emits typed requests to inspect a Lean file,
query the current proof state, search for relevant declarations, or check a
candidate snippet. The tool layer executes these requests against Lean-facing
services and returns structured outputs such as diagnostics, local hypotheses,
goal states, retrieved lemmas, or compact validity reports.

The purpose of this layer is to make the formal environment queryable during
proof search. It gives models access to information that would otherwise be
implicit in Lean's state or scattered across the library.

\paragraph{Tool categories.}
The tools exposed to the model fall into four broad categories. First,
\emph{candidate-checking tools} run bounded Lean checks on generated snippets or
candidate files and return compact validity information. Second,
\emph{diagnostic tools} parse Lean errors and warnings into a form suitable for
repair prompts. Third, \emph{proof-state tools} expose local goals, hypotheses and 
types at a specified location in a materialized Lean
file. Fourth, \emph{retrieval tools} search over Mathlib (Lean Standard Library), local declarations,
examples, or premise suggestions relevant to the current goal. These tools
support recursive decomposition by helping workers prove subgoals, repair local
failures, and identify facts needed for recombination. Table~\ref{tab:tool-classes}
summarizes the main tools and their subsequent classes exposed through the scaffold.

\paragraph{Guardrails.}
The same boundary also supports guardrails. The system rejects candidates that
change protected theorem signatures, leave unresolved proof holes, or use
forbidden shortcuts such as \texttt{sorry}, \texttt{admit}, unsound axioms, or
unsafe constructs. Intermediate stages may allow unresolved obligations when
the scaffold is still constructing the artifact, but final acceptance always
requires a complete Lean file with no proof bypasses.

\begin{table}[ht]
\centering
\footnotesize
\renewcommand{\arraystretch}{1.18}
\begin{tabular}{p{0.20\linewidth} p{0.28\linewidth} p{0.42\linewidth}}
\toprule
\textbf{Tool class} & \textbf{Representative tools} & \textbf{Role in proof search} \\
\midrule
Candidate checking
& \texttt{lean\_check\_candidate}
& Runs a bounded Lean check on a candidate snippet or file and returns validity
status with compact diagnostics. \\
\addlinespace[0.35em]

Diagnostics
& \texttt{lean\_diagnostic\_messages}$^\ast$
& Extracts Lean errors and warnings from a materialized file, including
locations and repair-relevant messages. \\
\addlinespace[0.35em]

Proof-state inspection
& \texttt{lean\_goal}$^\ast$,
  \texttt{lean\_term\_goal}$^\ast$,
  \texttt{lean\_hover\_info}$^\ast$
& Exposes local goals, hypotheses, types, and hover information at a concrete
file position. \\
\addlinespace[0.35em]

Retrieval / search
& \texttt{lean\_leansearch}$^\ast$,
  \texttt{lean\_loogle}$^\ast$,
  \texttt{lean\_leanfinder}$^\ast$,
  \texttt{lean\_leandex}
& Searches Mathlib, local declarations, APIs, examples, signatures, and
semantically related theorems. \\
\addlinespace[0.35em]

Premise search
& \texttt{lean\_state\_search}$^\ast$,
  \texttt{lean\_hammer\_premise}$^\ast$
& Suggests premises or automation-relevant facts for a concrete goal at a
specified file position. \\
\bottomrule
\end{tabular}
\caption[Tool classes exposed through the MCP-style layer]{Tool classes exposed through the MCP-style tool layer. Starred tools
are drawn from the Lean LSP MCP interface \citep{lean_lsp_mcp}; unstarred tools
are additional integrations used by the scaffold, such as LeanDex lookup and
bounded candidate checking. All tools provide guidance for proof search, while
final acceptance is determined only by Lean verification.}
\label{tab:tool-classes}
\end{table}

\subsection{Progress Evaluation}
\label{sec:progress-evaluation}

\paragraph{Partial progress under binary verification.}
Lean verification yields a binary accept/reject verdict, but inference need not
be all-or-nothing. A failed candidate can still represent useful progress. It
may close select proof holes, produce a valid helper lemma, repair a proof
sketch, or move the error
from a syntax failure to a smaller unsolved goal. Progress evaluation is the
mechanism that uses these intermediate signals to decide what action the overall scaffold should take: should it continue repairing a branch, decompose it further, route information
back to the task-level adapter, or stop the branch completely.

\paragraph{Deterministic gates.}
The first layer consists of deterministic gates. These checks enforce a number of needed controls: output
format, replacement counts, import hygiene, theorem-signature preservation,
no-bypass constraints, Lean verification status, recursion-depth limits, and
cost or call budgets. They serve to reject candidates
that violate hard constraints and accept those that satisfy the final
verification policy.

\paragraph{Task-level evaluation.}
The second layer operates over the full VeriCoding artifact, at the task level. It is invoked
when an iteration leaves unresolved obligations after constructing or revising
the candidate file. Its inputs include the current Lean artifact, recent
verifier diagnostics, the unresolved theorem or lemma blocks, and prior
control decisions. Its role is to decide whether the task-level scaffold
should continue, request additional helper theorems, revise generated
definitions, or abandon the current trajectory. This lets local proof failures
influence the overall file structure when the failure appears to depend on
missing auxiliary facts or unsuitable definitions.

\paragraph{Proof-level evaluation.}
The third layer operates inside \textsc{RecursiveWorker}, where it evaluates a
single theorem, lemma, or subgoal branch during recursive proof search. Rather
than judging the candidate in isolation, the evaluator can query the local Lean
state through read-only tools: diagnostics, goal inspection, term-goal
inspection, and hover/type information. These tool calls give the evaluator a
more precise view of the current obligation, the latest proof candidate or
sketch, and the surrounding local context. The evaluator then combines this
local information with the branch history and decomposition tree to choose an action.

\begin{figure}[t]
\centering
\begin{tikzpicture}[
    node distance=1.15cm,
    box/.style={draw, rounded corners, align=center, inner sep=5pt,
                minimum width=3.0cm, minimum height=0.9cm},
    smallbox/.style={draw, rounded corners, align=center, inner sep=5pt,
                     minimum width=3.1cm, minimum height=0.9cm},
    arrow/.style={->, thick}
]
\node[box] (candidate) {Candidate artifact\\or proof branch};
\node[box, below=of candidate] (gates) {Deterministic\\gates};
\node[box, below=of gates] (eval) {Progress\\evaluator};

\node[smallbox, below left=1.0cm and 2.4cm of eval] (continue)
    {Local continuation\\repair / tool use};
\node[smallbox, below=1.0cm of eval] (decomp)
    {Structural revision\\sketch / decompose};
\node[smallbox, below right=1.0cm and 2.4cm of eval] (escalate)
    {Escalation or pruning\\helper / def / abandon};

\draw[arrow] (candidate) -- (gates);
\draw[arrow] (gates) -- (eval);
\draw[arrow] (eval) -- (continue);
\draw[arrow] (eval) -- (decomp);
\draw[arrow] (eval) -- (escalate);
\end{tikzpicture}
\caption[Diagram of the progress-evaluation control flow]{
Progress-evaluation control flow. Deterministic gates first enforce hard
constraints such as format validity, theorem-signature preservation, no-bypass
checks, Lean verification status, and resource limits. If a candidate is not
accepted or rejected outright, the progress evaluator chooses among three
classes of response: local continuation, structural revision or further decomposition,
and escalation or pruning. These classes group the concrete task-level and
proof-level actions used by the scaffold.
}
\label{fig:progress-evaluation}
\end{figure}
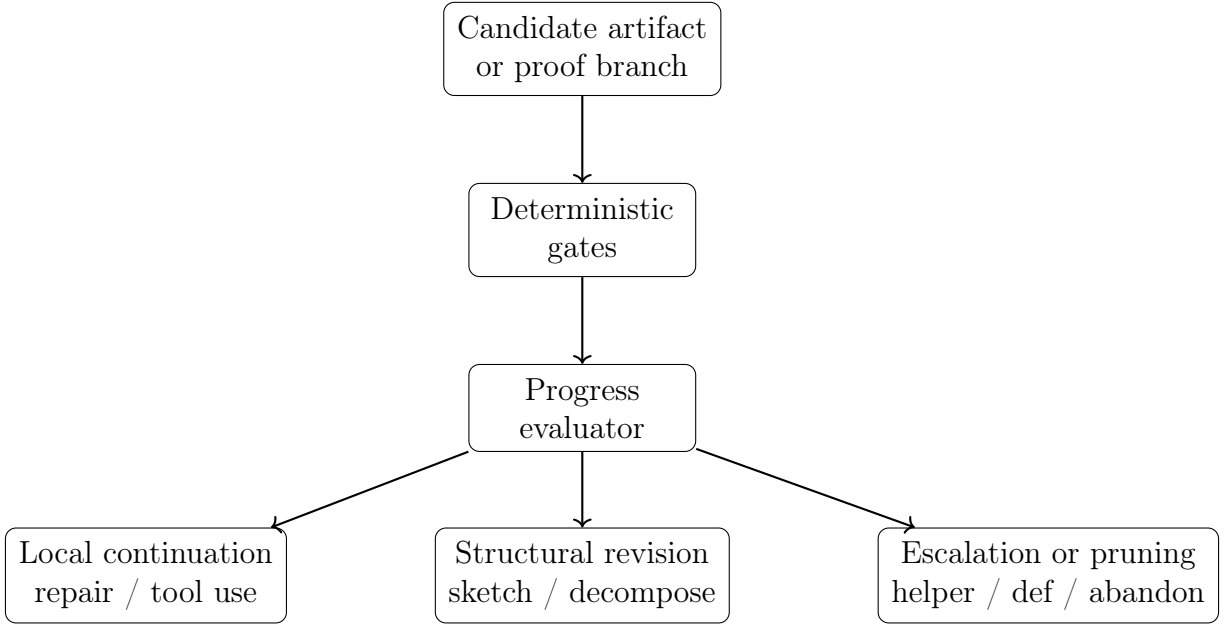

\paragraph{Action space.}
The evaluators emit discrete control actions, determining which
part of the scaffold should receive the next inference step. Some decisions
keep the branch local, allowing further repair or tool-assisted proof search.
Others change the proof structure, for example by revising the current proof
sketch or recursively decomposing a difficult subgoal. A third class escalates
the failure to the surrounding task-level scaffold or prunes the branch
entirely. Figure~\ref{fig:progress-evaluation} summarizes this control flow.

\paragraph{Control rationale.}
The purpose of this action space is to avoid treating all verifier failures as
equivalent. Some failures are local: a malformed tactic or
small type mismatch can remain within the repair loop. Others are structural:
a plausible proof sketch may need to be revised, or a hard subgoal may need to
be decomposed into smaller obligations. Still others indicate that the proof
branch is blocked by the surrounding artifact, such as a missing helper lemma or
an implementation choice that makes the specification difficult to prove. In
those cases, the evaluator routes the failure upward rather than spending more
budget on local repair.

\paragraph{Budgeted termination.}
Progress evaluation also prevents unbounded search. Branches terminate when
they exceed the configured repair budget, recursion depth, model-call budget,
or cost budget, or when the evaluator determines that further repair is
unlikely to help. Conversely, branches that show measurable progress can
receive additional attempts. This gives the scaffold a budgeted notion of
search as progress evaluation
determines how inference-time compute is allocated before final verification.

\section{Experimental Setup}
\label{sec:experimental-setup}

The previous section described the architecture of our verifier-in-the-loop
scaffold: a task-level adapter manages the global Lean artifact, while
\textsc{RecursiveWorker} performs recursive proof search over individual
theorem and lemma obligations. In this section, we describe how that scaffold
is evaluated. The goal is to specify
the task suites, experimental conditions, acceptance criteria, and measurement
protocol used in the experiments.

Our scaffold is built on an implemented HILBERT-style proof-search framework,
which introduced recursive proof construction with Lean verification, proof
sketching, subgoal decomposition, and recombination
\citep{10.48550/arxiv.2509.22819}. We therefore do not treat recursive
proof decomposition itself as a novel component of this thesis. Instead, our
contribution is to adapt and extend that proof-level worker into a verified-
coding scaffold for sectioned Lean artifacts.

Concretely, we add four pieces around the HILBERT-style worker. First, we add
task-level artifact management for VeriCoding tasks: the scaffold constructs,
updates, and verifies Lean files within distinct helper, definition, and theorem
regions, and promotes unresolved theorem or lemma obligations to
\textsc{RecursiveWorker}. Second, we add stronger final no-bypass checks, so
that a candidate is accepted only if the target signatures are preserved, all
obligations are discharged, and the final artifact contains no proof holes or
unsound shortcuts. Third, we add MCP-based tool access as specified in Section~\ref{sec:verification-tool-access}. Fourth, we implement a
Proof Reviser agent as specified in Section ~\ref{sec:progress-evaluation}, an additional model role.

Within this terminology, \textsc{RecursiveWorker} refers only to the
HILBERT-like proof-search component. The surrounding task adapter, verifier
interface, MCP tool layer, no-bypass guardrails, Proof Reviser, and experiment
runner are collectively referred to as our scaffold.

\subsection*{Task Suites}

We evaluate on three task suites. The first is a Vericoding pilot suite
containing 26 Lean tasks. This suite was constructed from an initial base run:
the baseline prover discharged the other candidate examples, and the remaining
unsolved tasks were retained as a compact pilot set for evaluating scaffolded
inference. Thus, the pilot suite is not intended to be a representative random
sample of Vericoding, but rather a small residual set where direct one-shot proving was
insufficient and verifier-guided search could provide additional signal. Each
task consists of a natural-language description, a formal statement, and a Lean
source file with editable regions. These regions include helper, definition, and
theorem sections, and proof obligations are represented by missing proof terms
(i.e., \texttt{sorry}). The pilot suite is used as a small end-to-end evaluation
setting for comparing direct repair, tool use, decomposition, and proof-reviser
variants.

The second suite is Verina, a larger VeriCoding-style dataset containing 189
Lean tasks. Verina uses the same general task schema as the pilot suite, but
provides a broader evaluation distribution. We use the same scaffold interface,
verification policy, and success criteria for both VeriCoding suites. This
allows the pilot suite to serve as a smaller validation setting before applying
the same procedure to the larger Verina evaluation.

The third suite is the Dalek prove-helpers worksheet sample. This benchmark is
not treated as another VeriCoding split. Instead, it is a separate stress test
for proof synthesis in a setting derived from systems-code verification. The
benchmark is extracted from Lean specifications produced by Aeneas for the Rust
\texttt{curve25519-dalek} crate. In this context, Aeneas is the translation
layer that turns Rust code and its associated correctness obligations into Lean
artifacts. The full extracted benchmark contains 190 self-contained Lean
projects, each centered on a main correctness theorem. We evaluate a fixed
sample of 30 tasks as a representative set and use the same sample across Dalek conditions.

The Dalek worksheet format differs from the VeriCoding format. Each worksheet
contains proof-only helper goals that must be solved before the main theorem
can be completed. These helper goals are introduced as fresh obligations, while
the original helper declarations in the surrounding source are redacted. A
successful solution therefore cannot reuse the original helper proofs. It must
instead prove the fresh helper goals in the submitted worksheet and then use
those proved helpers to close the main target. This makes the Dalek sample a
test of intermediate-lemma construction and proof composition, rather than
only local proof-hole completion, more representative of a true formal verification task.

\subsection*{Verification Infrastructure}

Lean verification is authoritative in all experiments. For the VeriCoding
suites, candidate artifacts are checked through Kimina, an implemented Lean 4
server that provides a scalable interface for Lean verification, diagnostics,
and proof-state feedback \citep{santos_kimina_2025}. Kimina supplies
the verifier interface used by our scaffold. Intermediate checks may guide
search while some obligations remain deferred, but final acceptance always
requires a fresh verification of the completed artifact under the no-bypass
policy.

The tool layer described in Section~\ref{sec:verification-tool-access} is used
only as an aid to proof search. In tool-enabled conditions, the model may
request structured information such as diagnostics, local proof states,
candidate-check results, or relevant declarations. These tool calls can help
\textsc{RecursiveWorker} repair failed proofs, identify useful facts, or decide
whether to decompose a goal further. They do not change the acceptance rule:
a task is solved only when the final artifact is accepted by the relevant
verifier or grader.

For Dalek tasks, final acceptance is determined by the Dalek grader rather than
by the VeriCoding verifier alone. The grader checks that the submitted
worksheet builds, satisfies the syntactic restrictions, and proves the required
main and helper targets without relying on forbidden \texttt{axioms} or redacted helper
proofs.

\subsection*{Experimental Conditions}

The experiments compare direct generation, sampling-based baselines, and
scaffold-based proof search. Across ablations, we hold fixed the task set,
model family, provider, verifier configuration, prompt family, and attempt
budget except for the factor being varied. This is important because formal
proof generation is sensitive to inference budget, verifier feedback, and the
availability of tool calls.

The simplest baseline is one-shot generation. In this condition, the model
receives the task prompt and produces a single Lean candidate. The candidate is
checked after generation, but verifier feedback is not returned to the model
during the attempt. This baseline measures how much of each task suite can be
solved by direct synthesis alone.

The second baseline is best-of-\(n\) sampling. The model produces multiple
independent candidates under the same task prompt, and the first candidate that
passes the verifier is accepted. This controls for additional sampling budget
without adding structured repair, recursive decomposition, or progress-guided
search. A tool-enabled variant gives each attempt bounded access to the Lean
tool layer while preserving the same overall sampling budget. This isolates the
effect of tool access from the effect of the scaffold's proof-decomposition and
repair procedure.

We then evaluate ablations involving \textsc{RecursiveWorker}. 
In the recursive-without-tools condition, \textsc{RecursiveWorker} can generate
proof sketches, extract subgoals, prove subgoals recursively, and recombine
verified subproofs, but it does not receive MCP tool access. This tests whether
decomposition alone improves proof search. In the recursive-with-tools
condition, bounded tool access is added during subgoal proving and correction.
This tests whether proof-state inspection, diagnostics, and retrieval help the
worker close extracted subgoals. The full scaffold condition combines ann the above with the Proof Reviser as detailed in Section~\ref{sec:progress-evaluation}. This
condition represents the complete experimental system rather than a single
isolated mechanism. Where a table reports only a subset of these ablations, we
interpret comparisons only among conditions with matched task sets and matched
budgets.

The Dalek worksheet experiments are reported separately from the VeriCoding
experiments. For Dalek, we compare a no-tools direct baseline against a
tool-enabled baseline under the same worksheet sample and retry budget. These
conditions use the Dalek grader as the final acceptance mechanism. We do not
pool Dalek results with VeriCoding results, since the task format, grading
procedure, and proof obligations differ.

\subsection*{Success Criteria}

For VeriCoding tasks, a task is counted as solved only if the final Lean
artifact passes the configured final verification check. The target theorem or
declaration signature must be preserved, and all proof obligations must be
discharged. The final artifact must not contain unresolved proof holes or
bypass constructs, including uses of \texttt{sorry}, \texttt{admit},
\texttt{sorryAx}, arbitrary axioms, unsafe casts, external implementations, or
other constructs that make the proof vacuous. A candidate that passes an
intermediate check while leaving deferred obligations unresolved is not counted
as a success.

For Dalek worksheet tasks, a task is counted as solved only if the submitted
worksheet passes the Dalek grader. The worksheet must build successfully, pass
the syntactic anti-cheat checks, and satisfy the semantic axiom checks for the
main theorem and all helper targets. The proof may rely only on the permitted
base axioms of the Lean environment. It may not rely on redacted helper proofs,
hidden proof gaps, or any occurrence of \texttt{sorryAx} that might bypass the helper theorem proofs. Since the worksheet
variant introduces fresh helper obligations, a successful submission must prove
those helpers in the worksheet itself and use them to complete the main target.

\subsection*{Metrics}

The primary metric is task success rate, reported as the number of solved tasks
divided by the number of attempted tasks for each condition. We report success
rates separately for the VeriCoding pilot suite, Verina, and the Dalek
worksheet sample.

We also record efficiency and process metrics. These include total input and
output tokens, estimated model cost, wall-clock time per task, the number of model calls, and the number of Lean verifier
calls. For tool-enabled conditions, we record the number and type of tool calls.
For conditions using \textsc{RecursiveWorker}, we additionally record the
number of generated subgoals, the number of solved and failed subgoals, the
maximum recursion depth reached, and the point at which proof search
terminates. When the progress evaluator or proof-revision layer is active, we also
record the distribution of its control decisions.

Failures are bucketed for post-hoc analysis. VeriCoding failures may arise
during non-theorem generation, direct proof search, subgoal extraction,
recursive subgoal solving, recombination, final verification, or timeout. Dalek
failures are categorized according to whether they arise from syntactic
rejection, build failure, unauthorized axioms, unsolved helper goals, or failure
to close the main theorem. These buckets are diagnostic only; they do not alter
the binary success criterion.

\subsection*{Evaluation Controls}

The remaining role of the evaluation protocol is
to ensure that comparisons are controlled and reproducible. For each condition,
we use a fixed task manifest and fixed run configuration. In ablation
comparisons, the task set, model, provider, verifier or grader, prompt family,
attempt budget, and token budget are held fixed; only the mechanism under
evaluation is varied. In particular, comparisons between the
recursive-with-tools condition and the full scaffold isolate the addition of
the \textsc{ProofReviser}, since both conditions otherwise use the same
\textsc{RecursiveWorker}, verifier feedback, and tool-access setting.

We report VeriCoding and Dalek results separately rather than pooling them. The
VeriCoding suites are evaluated through the Lean/Kimina final-verification
pipeline, whereas the Dalek worksheet sample is evaluated through the Dalek
grader. Because the task formats, acceptance checks, and proof obligations
differ, aggregate success rates across these settings would not be
interpretable as a single benchmark score.

All model-facing API calls in the reported experiments are routed through
OpenRouter using an OpenAI-compatible chat-completion interface. This includes
calls made by the direct-generation baselines, the formal prover, reasoning
or sketch-generation calls inside \textsc{RecursiveWorker}, tool-calling model
roles, and the \textsc{ProofReviser} when it is enabled. Routing all model
calls through the same provider interface standardizes authentication,
request formatting, usage accounting, and model selection across experimental
conditions.

\section{Results}
\label{sec:part-ii-results}

This section evaluates the inference-time verification scaffold on the
Vericoding pilot set, VERINA, and Dalek.
Unless otherwise noted, all methods in this
section use the same GPT-5.3 Codex base model; the comparisons therefore isolate
differences in inference-time control, tool use, decomposition, repair, and
proof-reviser feedback rather than differences in the underlying model. The main
results compare direct generation with repair, top-level tool use, the previous
E5 scaffold, and the full scaffold with proof reviser under reported inference
budgets. We then use the Vericoding pilot subset for more detailed configuration
comparisons, since that subset has the most complete scaffold ablations. Finally
we analyze tool use, cost--success trade-offs, and common failure modes.
\subsection*{Main Benchmark Results}
\label{sec:part-ii-main-results}

Table~\ref{tab:part-ii-main-results} reports the headline verification results
for the Vericoding pilot set, VERINA, and Dalek. On the Vericoding pilot set, direct
generation with five verifier-guided repairs solves 12 of 26 tasks
(46.2\%). Exposing only top-level tools does not improve this baseline in this
run, solving 6 of 26 tasks (23.1\%). The previous E5 scaffold solves 17 of 26
tasks (65.4\%), while the strongest pilot result is the full scaffold with proof
reviser, which solves 18 of 26 tasks (69.2\%). Because these rows use the same
base model, the improvement reflects the inference procedure: the higher-scoring
methods structure search around decomposition, verifier feedback, repair, and
proof-reviser checks rather than relying only on direct retry or outer-level
tool access.

The gain comes with a substantially larger inference budget. Direct repair uses
far fewer model and verifier/tool calls than the full scaffold with proof
reviser. Thus, the scaffold should be interpreted as a cost-sensitive search
procedure: it improves verification success by spending additional structured
inference budget, not by providing a cheaper single-pass method. Cost comparisons
should also be treated cautiously for rows where provider spend was not fully
captured; in those cases, call counts and token usage are more reliable than the
reported dollar estimate.

On VERINA broad, the direct repair baseline solves 117 of 187 tasks (62.6\%).
The top-level-tools-only variant solves 109 of 187 tasks (58.3\%), indicating
that simply allowing tool calls at the outer level does not improve pass rate in
this setting. We also report a harder VERINA residual setting, consisting of
tasks left unsolved by earlier direct-repair or top-level-tool baselines. On
this residual subset, whole-task decomposition with tools and the proof reviser
solves 7 of 42 tasks (16.7\%). Because this subset is selected from previously
unsolved tasks, the result should be interpreted as a stress test for the
scaffold rather than as a broad VERINA pass rate.

Dalek-Bench (Dalek Benchmark) provides a harder cross-benchmark stress test. On the 30-task
Dalek-Bench subset, direct generation without Lean tools solves 3 of 30 tasks
(10.0\%), while direct generation with Lean tools solves 5 of 30 tasks
(16.7\%). Thus, tool access provides a small improvement in this setting, but
the overall solve rate remains low. The Hilbert-corrected
recursive-decomposition scaffold solves \(2\) of 30 tasks (\(6.7\%\)), indicating
that the scaffold gains observed on the Vericoding pilot set do not yet transfer
cleanly to Dalek-Bench. The failure mode is not merely insufficient search
depth: the recursive framework often spends substantial token and tool-call
budget pursuing decompositions that do not produce repairable intermediate proof
states. These results suggest that progress on Dalek-Bench will require more
directed decomposition, stronger progress evaluation, and better task-specific
tool-use policies rather than simply increasing recursive search budget.

\begin{table}[!ht]
  \centering
  \small
  \setlength{\tabcolsep}{3.5pt}
  \renewcommand{\arraystretch}{1.08}

  \begin{tabularx}{\linewidth}{p{0.17\linewidth}Xrrrrrr}
    \toprule
    Benchmark & Method & Solved & Pass
    & \shortstack{Model\\calls}
    & \shortstack{Verif./tool\\calls}
    & Cost
    & \shortstack{Cost/\\solve} \\
    \midrule

    Vericoding pilot
    & Direct + Repair
    & 12 / 26 & 46.2\% & 159 & 109 & \$18.85 & \$1.57 \\

    & Direct + Repair + Tools
    & 6 / 26 & 23.1\% & 56 & 261 & \ -- & \ -- \\

    & RecursiveWorker + Proof Reviser
    & 18 / 26 & 69.2\% & 3,231 & 2,938 & \$132.69 & \$7.37 \\
    
    & RecursiveWorker + Tools & 17 / 26 & 65.4\% & 3,194 & -- & \$43.51 & \$2.56 \\ \\

    \midrule

    VERINA broad
    & Direct + Repair
    & 117 / 187 & 62.6\% & 945 & 551 & \ -- & \ -- \\

    & Direct + Repair + Tools
    & 109 / 187 & 58.3\% & 435 & 1,314 & \-- & \-- \\

    \midrule

    VERINA hard
    & RecursiveWorker + Tools + Proof Reviser
    & 7 / 42 & 16.7\% & 7,354 & 10,363 & \$298.89 & \$42.70 \\

    \midrule
    
    Dalek-Bench (Subset 30)
    & Direct + Repair
    & 3 / 30 & 10.0\% & 142 & 142 & \$8.75 & \$2.92 \\
    
    & Direct + Repair + Tools
    & 5 / 30 & 16.7\% & 2,102 & 271 & \$207.53 & \$41.51 \\
    
     & RecursiveWorker + Tools + Proof Reviser & 2 / 30 & 6.7\% & 2,084 & 4,969 & \$127.51 & \$63.76 \\

    \bottomrule
  \end{tabularx}

    \caption[Headline verification results across benchmarks and scaffolds]{Headline verification results for the Vericoding
  pilot set, VERINA,
    and the Dalek Track C sample using the same GPT-5.3 Codex
  base model. The
    Vericoding pilot rows compare direct repair, top-level tool
  use, the previous
    E5 scaffold, and the full scaffold with proof reviser on the
  same 26-task
    pilot suite. E5 denotes the fifth configuration in the
  earlier
    scaffold-ablation series and serves as the prior full-
  scaffold reference point
    before adding the proof reviser; the full E1--E5
  configuration series is
    reported separately in Table~\ref{tab:part-ii-pilot-configs}.
  The VERINA broad
    rows evaluate methods on the 187-task VERINA split used in
  this experiment.
    The VERINA hard row is evaluated only on tasks left unsolved
  by earlier
    direct-repair or top-level-tool baselines, so it should be
  read as a stress
    test for harder remaining examples rather than as an all-
  VERINA pass rate.
    The Dalek rows use a fixed 30-task Track C sample; the
  RecursiveWorker Dalek
    row is a live partial run and is marked with observed lower-
  bound calls and
    cost until the run finishes. Costs are reported run estimates
  when available;
    several Codex-backend rows did not fully capture provider
  spend, so cost
    comparisons should be interpreted together with model-call
  and
    verifier/tool-call counts.}
  \label{tab:part-ii-main-results}
\end{table}

\subsection*{Pilot-Subset Configuration Comparisons}
\label{sec:part-ii-pilot-configs}

The Vericoding pilot set provides the cleanest setting for scaffold
configuration comparisons because all runs share the same 26-task workload and
the same GPT-5.3 Codex base model. The E1--E5 labels refer to successive
configurations of the RecursiveWorker scaffold, not to different underlying
models. These configurations vary how inference budget is allocated across
decomposition, subgoal solving, repair, tool use, and progress evaluation.

Table~\ref{tab:part-ii-pilot-configs} summarizes these pilot-only ablations.
The main pattern is that scaffold performance depends on how the search budget
is structured, not only on how much budget is spent. E1 tests shallow
decomposition without a strong repair loop and solves only 7 of 26 tasks. E2
adds repair to the shallow setting and doubles the solve count to 14, showing
that verifier-guided retry is a major source of improvement. E3 makes
decomposition recursive, but without enough effective repair or budget control,
it underperforms E2 despite using more model calls. E4 increases the recursive
budget and improves over E3, but at the highest historical cost, illustrating
diminishing returns from expanding search alone.

The E5 configurations improve the allocation of budget across decomposition,
repair, and progress checks. The full E5 scaffold solves 17 of 26 tasks while
using fewer calls and lower reported cost than the larger-budget E4 setting.
The recursive-repair variant sacrifices one solved task but reduces cost, while
the recursive-repair plus subgoal-tools variant recovers the 17-task solve count
at the lowest reported cost among the 17-solve configurations. Finally, the
newer full scaffold with proof reviser achieves the highest pilot solve count,
18 of 26, but with higher reported cost. This makes the proof-reviser variant
an accuracy-oriented configuration rather than the most cost-efficient one.

\begin{table}[!ht]
  \centering
  \footnotesize
  \setlength{\tabcolsep}{3pt}
  \renewcommand{\arraystretch}{1.12}

  \begin{tabularx}{\linewidth}{p{0.28\linewidth}Xrrrr}
    \toprule
    Configuration & Description & Solved / n & Pass
    & \shortstack{Model\\calls} & Cost \\
    \midrule

    E1: shallow decomposition
    & Tests whether a shallow decomposition scaffold is useful without a strong
    repair loop; low success despite moderate cost.
    & 7 / 26 & 26.9\% & 4,204 & \$57.59 \\

    E2: shallow decomposition + repair
    & Adds verifier-guided repair to shallow decomposition; substantially higher
    success, but higher cost.
    & 14 / 26 & 53.8\% & 4,945 & \$97.58 \\

    E3: recursive decomposition
    & Adds recursive decomposition; spends more calls than E2 but underperforms,
    suggesting that deeper decomposition alone is not sufficient.
    & 11 / 26 & 42.3\% & 5,962 & \$101.65 \\

    E4: recursive decomposition + larger budget
    & Expands the recursive search budget; improves over E3 but at the highest
    historical cost, showing diminishing returns from budget alone.
    & 15 / 26 & 57.7\% & 8,775 & \$154.15 \\

    E5: full scaffold
    & Balances decomposition, repair, and progress evaluation; improves solve
    count while using less budget than E4.
    & 17 / 26 & 65.4\% & 5,104 & \$76.36 \\

    E5: recursive repair
    & Shifts budget toward recursive repair; reduces cost and calls, with a
    small loss in solve count.
    & 16 / 26 & 61.5\% & 3,327 & \$56.16 \\

    E5: recursive repair + subgoal tools
    & Adds subgoal-level tool use; recovers the 17-task solve count with the
    lowest reported cost among the 17-solve settings.
    & 17 / 26 & 65.4\% & 3,194 & \$43.51 \\

    Full scaffold + proof reviser
    & Adds proof-reviser feedback but no tool access; achieves the highest solve count, but with
    higher reported cost.
    & 18 / 26 & 69.2\% & 3,231 & \$132.69 \\

    \bottomrule
  \end{tabularx}

  \caption[Scaffold configuration comparisons on the Vericoding pilot]{Pilot-only configuration comparisons on the 26-task Vericoding pilot
  subset, using the same GPT-5.3 Codex base model. The E1--E5 rows compare
  successive RecursiveWorker scaffold configurations and show how performance
  changes as budget is allocated differently across shallow decomposition,
  recursive decomposition, repair, subgoal tools, and progress evaluation. The
  final row adds proof-reviser feedback and achieves the highest solve count,
  while the recursive-repair plus subgoal-tools variant is the most
  cost-efficient among the 17-solve configurations. These rows should be read as
  scaffold design trends rather than as an exhaustive hyperparameter sweep.}
  \label{tab:part-ii-pilot-configs}
\end{table}

\subsection*{Tool Use and Cost--Success Trade-offs}
\label{sec:part-ii-tool-cost}

\begin{figure}[t]
  \centering
  \includegraphics[width=\linewidth]{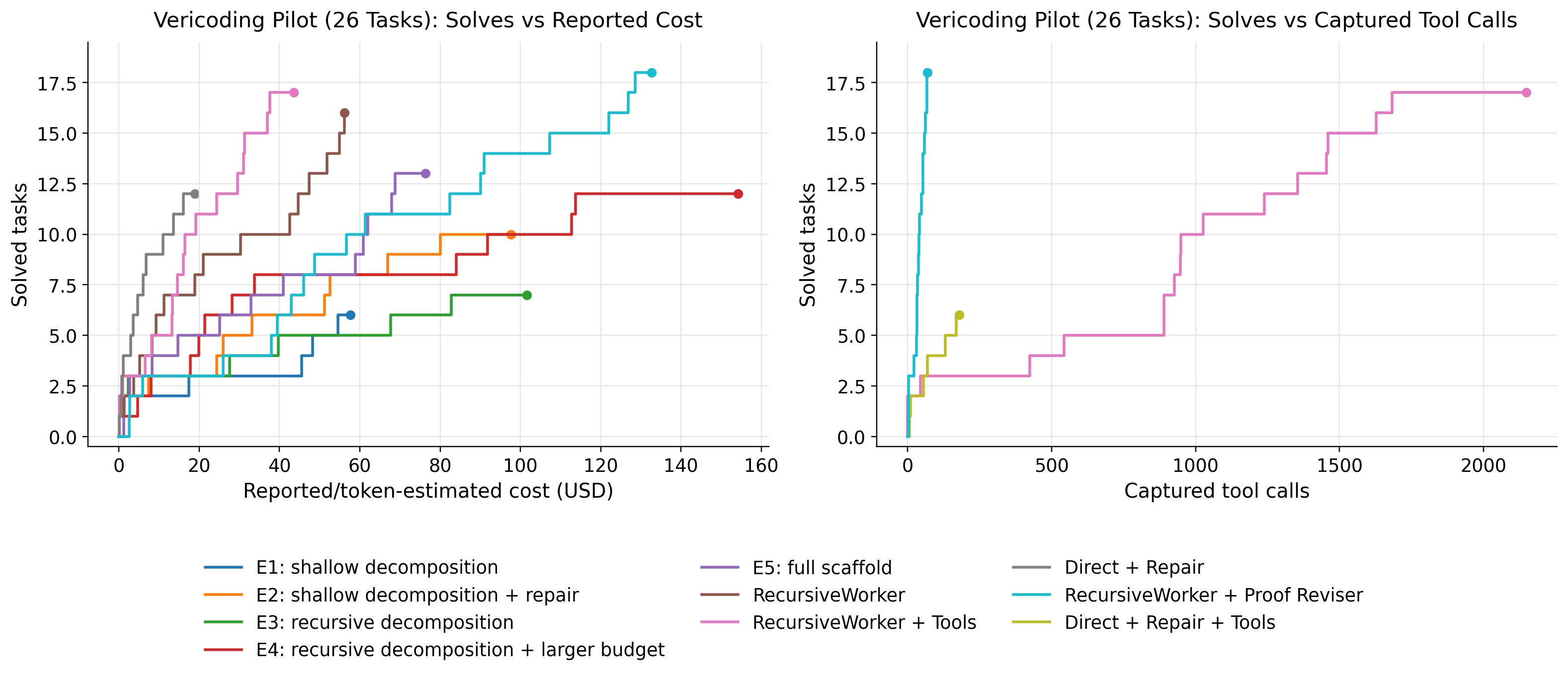}
  \caption[Budget-efficiency curves on the Vericoding pilot set]{Budget-efficiency curves on the 26-task Vericoding pilot set. The
  left panel plots cumulative solved tasks against reported or token-backed
  dollar cost, omitting runs whose provider cost was not reliably captured. The
  right panel plots solved tasks against captured interactive tool calls and
  therefore includes only runs with nonzero tool-event logging. The full
  scaffold with proof reviser reaches the highest pilot solve count, but does
  so at higher reported spend, while tool-heavy configurations vary
  substantially in the amount of captured tool use required per additional
  solve.}
  \label{fig:part-ii-budget-efficiency}
\end{figure}

\begin{figure}[t]
  \centering
  \includegraphics[width=\linewidth]{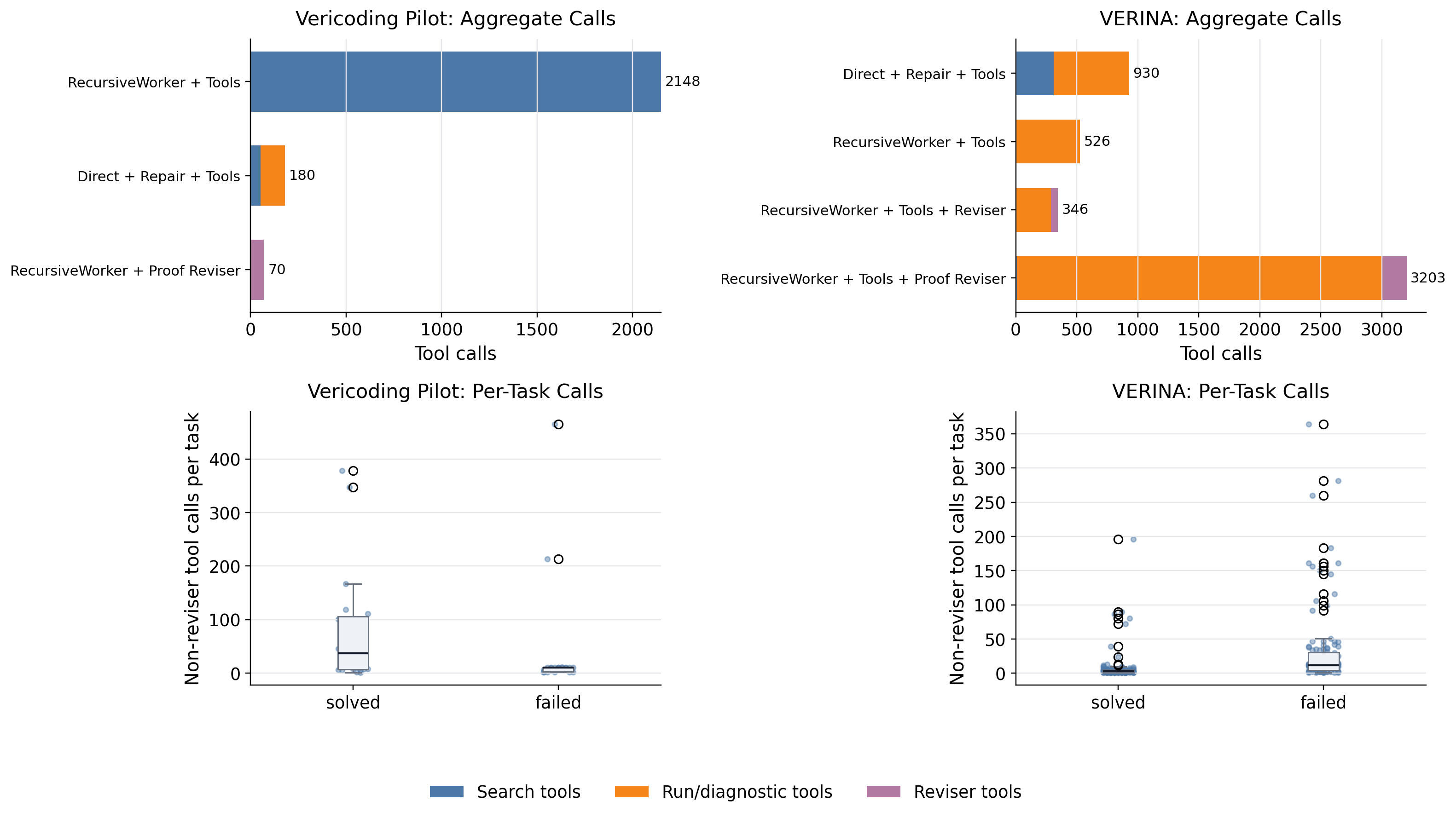}
  \caption[Captured tool-call usage by benchmark and tool group]{Captured interactive tool-call usage by benchmark family and tool
  group. The top row reports aggregate calls, grouped into search tools,
  run/diagnostic tools, and proof-reviser tools; the bottom row reports
  per-task non-reviser tool counts for solved and failed trajectories.
  Separating reviser calls prevents proof-reviser activity from being conflated
  with prover-side search. On the Vericoding pilot set, the historical
  subgoal-tools run is dominated by search calls, whereas the proof-reviser run
  records primarily reviser-side tool use. On VERINA, the whole-task
  decomposition run spends most of its captured tool budget on run/diagnostic
  calls.}
  \label{fig:part-ii-tool-use}
\end{figure}

Figures~\ref{fig:part-ii-budget-efficiency} and~\ref{fig:part-ii-tool-use}
analyze how the scaffold spends its inference budget. The budget-efficiency
curves show that the strongest pilot configuration reaches the highest solve
count only after spending substantially more inference budget than direct
repair. This reinforces the interpretation of the scaffold as a structured
search procedure rather than a low-cost replacement for direct generation.

The tool-use diagnostics separate captured interactive tool calls from verifier
or checker calls. This distinction is important because many runs use the
verifier heavily without exposing additional interactive tools to the model.
Tool access by itself is also not sufficient: the top-level-tools-only variants
do not outperform direct repair, while the higher-performing scaffolded runs use
tools inside a broader control loop involving decomposition, repair, and
progress evaluation.

\subsection*{Failure Modes}
\label{sec:part-ii-failure-modes}

\begin{figure}[t]
  \centering
  \includegraphics[width=\linewidth]{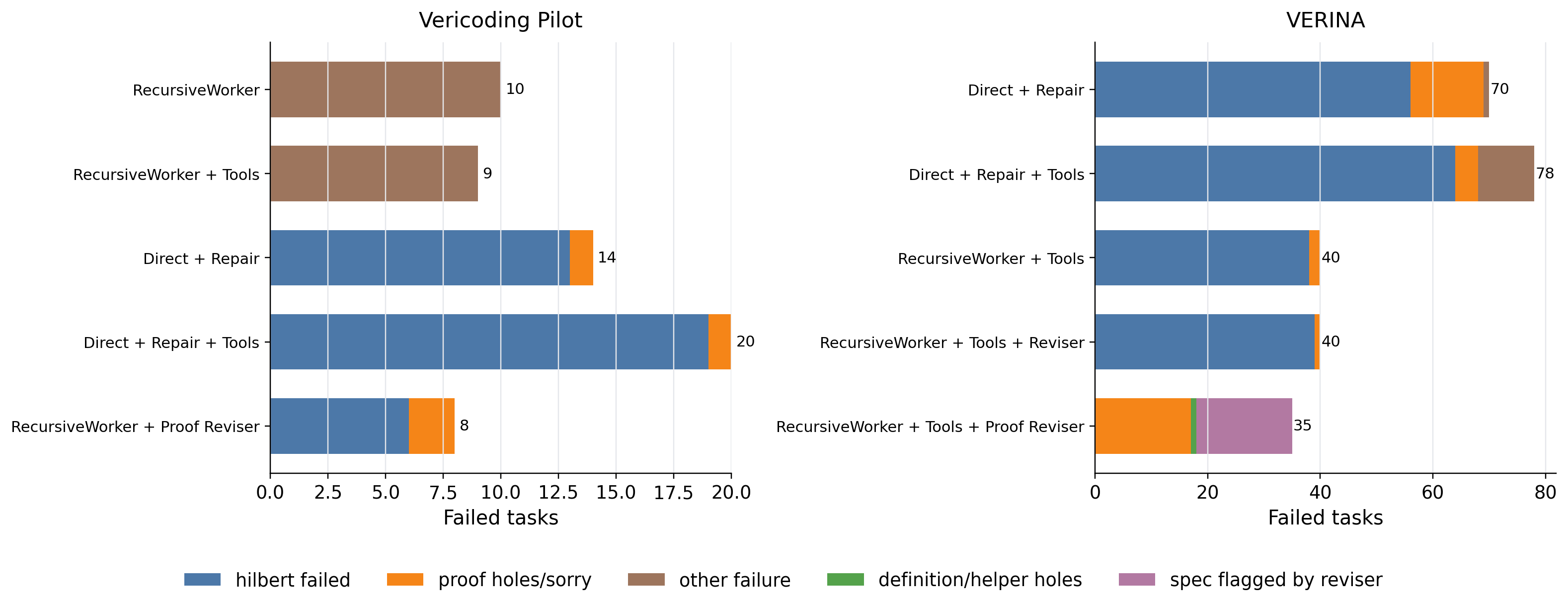}
  \caption[Failure-mode counts by benchmark and failure group]{Failure-mode diagnostics for the Vericoding pilot and VERINA runs.
  Bars group failed tasks into heuristic buckets inferred from trace errors,
  verifier diagnostics, proof-reviser decisions, and final Lean files. On the
  Vericoding pilot set, the full scaffold with proof reviser reduces the number
  of failures relative to direct repair and top-level tools, but still leaves
  Hilbert-stage and proof-hole failures. On VERINA, the broad direct-repair and
  top-level-tool baselines mostly fail in the Hilbert stage, while the hard
  residual whole-task run shifts some remaining failures toward proof holes or
  reviser-flagged trajectories. These buckets are debugging signals rather than
  formal, mutually exclusive error labels.}
  \label{fig:part-ii-failure-modes}
\end{figure}

Figure~\ref{fig:part-ii-failure-modes} summarizes failed trajectories using
heuristic buckets derived from trace errors, verifier diagnostics,
proof-reviser decisions, and final Lean files. These buckets are intended as
debugging signals rather than formal error labels. On the Vericoding pilot set,
the full scaffold with proof reviser reduces the number of failures relative to
direct repair and top-level tools, but the remaining failures are still
concentrated in Hilbert-stage failures and proof holes. This indicates that the
scaffold improves search, but does not eliminate the need for correct helper
definitions, proof structure, and theorem-level completion.

The VERINA failures show the same broader pattern under a harder stress-test
setting. Broad VERINA failures are dominated by Hilbert-stage failures, while
the residual runs expose more cases involving proof holes, unproductive
trajectories, or specification-related reviser decisions. These failures suggest
that verifier feedback is useful only when the scaffold can convert it into a
repairable intermediate state. When decomposition is poor, helper definitions
are incomplete, or the progress evaluator misjudges a branch, the system can
continue spending budget without reaching a verified solution.

\subsection*{Qualitative Examples}
\label{sec:part-ii-qualitative}

We include one successful trajectory and one representative failure to make the
aggregate trends concrete. The successful trajectory is
\texttt{rustbench\_array\_concat}. Direct repair and top-level tools both fail
on this task, while the full scaffold with proof reviser verifies it. The
verified artifact implements concatenation as \texttt{a ++ b}, then proves the
postcondition by decomposing it into three smaller obligations: the size of the
result, correctness of indices inherited from the left array, and correctness of
indices inherited from the right array. The final theorem recombines these
helper facts into the required postcondition, as shown in
Listing~\ref{lst:array-concat-success}.

\begin{lstlisting}[
language=Lean,
caption={Excerpt from the successful \texttt{rustbench\_array\_concat} trajectory. The final theorem recombines separately proved helper facts for result size, left-side indexing, and right-side indexing.},
label={lst:array-concat-success}
]
def arrayConcat
    (a : Array Int) (b : Array Int)
    (h_precond : arrayConcat_precond a b) : Array Int :=
  a ++ b

theorem arrayConcat_spec_satisfied
    (a : Array Int) (b : Array Int)
    (h_precond : arrayConcat_precond a b) :
    arrayConcat_postcond a b
      (arrayConcat a b h_precond) h_precond := by
  simp only [arrayConcat_postcond, arrayConcat]

  have h_size :
      (a ++ b).size = a.size + b.size := by
    exact h_size_arrayConcat_spec_satisfied a b h_precond

  have h_left :
      ∀ i : Nat, 0 ≤ i ∧ i < a.size →
        (a ++ b)[i]! = a[i]! := by
    exact h_left_arrayConcat_spec_satisfied
      a b h_precond h_left_bound h_left_index

  have h_right :
      ∀ i : Nat, 0 ≤ i ∧ i < b.size →
        (a ++ b)[i + a.size]! = b[i]! := by
    exact h_right_arrayConcat_spec_satisfied
      a b h_precond h_right_bound h_right_rewrite

  exact And.intro h_size (And.intro h_left h_right)
\end{lstlisting}

This example illustrates the value of scaffolded search. The successful run is
not merely a direct retry; it constructs a proof tree in which smaller
verifier-checkable obligations are proved independently and then assembled into
the final theorem. In this case, decomposition gives the model a useful proof
structure that simpler baselines do not find.

The representative failure is \texttt{rustbench\_remove\_elements}. In the full
scaffold with proof reviser, the task fails before theorem repair can make
useful progress. The generated non-theorem artifact still contains an incomplete
implementation, and the postcondition includes a malformed indexed expression.
In particular, the intended array access is passed as an argument without the
parentheses needed for Lean to parse it correctly, as shown in
Listing~\ref{lst:remove-elements-failure}.

\begin{lstlisting}[
language=Lean,
caption={Malformed candidate fragment from the failed \texttt{rustbench\_remove\_elements} trajectory. The implementation remains incomplete and the postcondition contains malformed indexed expressions such as \texttt{inArray a c[k]!}.},
label={lst:remove-elements-failure}
]
def removeElements
    (a : Array Int) (b : Array Int)
    (h_precond : removeElements_precond a b) : Array Int :=
  sorry

def removeElements_postcond
    (a : Array Int) (b : Array Int) (c : Array Int)
    (h_precond : removeElements_precond a b) : Prop :=
  (∀ k, k < c.size → inArray a c[k]! ∧ ¬inArray b c[k]!) ∧
  (∀ i j, i < j ∧ j < c.size → c[i]! ≠ c[j]!)

theorem removeElements_spec_satisfied
    (a : Array Int) (b : Array Int)
    (h_precond : removeElements_precond a b) :
    removeElements_postcond a b
      (removeElements a b h_precond) h_precond := by
  sorry
\end{lstlisting}

Lean rejects this candidate during non-theorem validation with a syntax error:

\begin{lstlisting}[
caption={Validation error for the failed \texttt{rustbench\_remove\_elements} candidate.},
label={lst:remove-elements-error}
]
Error message from Lean: unexpected token '!'; expected ')', ',' or ':'

The error was encountered while trying to process the following lines:
  (∀ k, k < c.size → inArray a c[k]! ∧ ¬inArray b c[k]!) ∧
\end{lstlisting}

The contrast between these examples clarifies both what the scaffold contributes
and where it remains brittle. In \texttt{rustbench\_array\_concat}, the scaffold
turns a difficult postcondition into smaller obligations that can be proved and
recombined. In \texttt{rustbench\_remove\_elements}, the system fails earlier:
the generated artifact is not yet a valid Lean object, so downstream
theorem-level repair cannot act on a meaningful proof state. This motivates
stronger whole-task validation, helper-region repair, and syntax-aware progress
checks before allocating large budgets to proof search.

\section{Discussion and Future Work}
\label{sec:part-ii-discussion}

\subsection{Discussion}
\label{sec:part-ii-discussion-main}

\paragraph{Tool access is not the same as tool use.}
A central result of this chapter is that simply exposing tools to the model is
not sufficient, and can even reduce performance. On the Vericoding pilot set,
direct repair solves 12 of 26 tasks, whereas the top-level-tools variant solves
only 6 of 26. The same pattern appears on VERINA broad, where direct repair
solves 117 of 187 tasks and top-level tools solve 109 of 187. Because these
comparisons use the same GPT-5.3 Codex base model, the difference is not
explained by model variance. Instead, it suggests that tool access introduces a
larger action space and additional opportunities for unproductive search unless
tool calls are embedded in a control structure. The higher-performing runs are
not the ones that merely expose tools, but the ones that decide when to
decompose, when to repair, when to inspect verifier feedback, and when to stop.

\paragraph{Verifier feedback as structured search.}
The strongest pilot result comes from the full scaffold with proof reviser,
which solves 18 of 26 tasks. This improves over direct repair and over the
previous E5 full-scaffold result, but the improvement should be interpreted as a
property of the inference procedure rather than of the base model. The scaffold
turns verifier feedback into structured search: failed attempts are not only
retried, but routed through decomposition, subgoal attempts, progress checks,
and proof-reviser feedback. This changes the role of the verifier from a
terminal pass/fail oracle into an intermediate control signal.

\paragraph{Cost-sensitive inference.}
The gains from scaffolded inference come with a substantially larger inference
budget. Direct repair is much cheaper in model calls, verifier/tool calls, and
reported cost, while the full scaffold with proof reviser spends many more
calls to recover additional solved tasks. Thus, the scaffold should be viewed as
a cost-sensitive search procedure rather than a lower-cost replacement for
direct generation. This distinction matters for deployment: in high-value
formal-verification settings, an additional verified task may justify many
extra calls, but in latency- or budget-constrained settings, the relevant metric
is cost per solve rather than pass rate alone.

\paragraph{Benchmark variation and residual difficulty.}
The evaluation sets play different roles. The Vericoding pilot set provides a
compact controlled workload for comparing scaffold configurations, while VERINA
broad tests the same high-level methods on a larger split. The VERINA hard
residual set is selected from tasks left unsolved by earlier baselines, so it
acts as a stress test for the remaining tail of difficult examples. Dalek-Bench
adds a different kind of stress test: it checks whether the recursive
decomposition framework transfers to a distinct benchmark family. The low
Dalek-Bench solve rates show that this transfer is not automatic. In this
setting, additional tools or deeper decomposition can consume substantial budget
without reliably producing useful intermediate proof states, motivating more
directed search and learned control.

\paragraph{Remaining brittleness.}
The failure analysis shows that verifier feedback is useful only when the
system can convert it into a repairable intermediate state. The scaffold still
fails when decompositions are irrelevant, helper definitions are incomplete,
proof holes remain in non-theorem regions, or the progress evaluator continues
an unproductive branch. The qualitative failure on
\texttt{rustbench\_remove\_elements} illustrates this problem: if the generated
artifact is malformed before theorem proving begins, then downstream repair may
spend budget without ever receiving a meaningful proof state. This motivates
stronger whole-task validation and repair, not only better theorem-body search.

\paragraph{Connection to Part~I.}
Part~I and Part~II use verifier feedback in complementary ways. In Part~I, the
verifier supplies a reward signal used to update the policy. In Part~II, the
policy is fixed and the verifier guides inference-time search around it. These
settings expose different design problems. RLVR depends on reward faithfulness,
task filtering, and avoiding reward hacking. Scaffolded inference depends on
control: deciding when to call tools, how to decompose, when to repair, and when
to abandon a branch. In both cases, the verifier is powerful only when its
feedback is made actionable by the surrounding training or inference
environment.

\subsection{Future Work}
\label{sec:part-ii-future-work}

\paragraph{Learned progress evaluation.}
The current scaffold relies on progress judgments to decide whether to continue,
repair, or abandon a branch. Future work could train a verifier-aware progress
model from scaffold traces, using successful and failed trajectories to predict
which partial states are likely to lead to verified solutions. This would make
branch selection and stopping decisions less heuristic and could reduce wasted
budget on unproductive repair loops
\citep{leanprogress2025,hubert2025alphaproof}.

\paragraph{Trained revisers and value networks.}
The proof reviser in this work is still largely a prompted component. A natural
next step is to replace or augment it with trained value networks and reviser
models. A value network could estimate the probability that a partial proof
state, candidate artifact, or search branch will eventually verify. A trained
reviser could classify the failure type, identify which region of the artifact
should be edited, and predict whether the next action should be repair,
decomposition, retrieval, or abandonment. Such models could be trained from
scaffold traces by labeling partial states according to eventual success,
remaining proof depth, verifier-error type, or cost-to-solve. This would move
the scaffold from hand-designed control rules toward learned verifier-guided
search, closer in spirit to proof-progress prediction and reinforcement-guided
formal reasoning systems \citep{leanprogress2025,hubert2025alphaproof}.

\paragraph{Training tool use rather than prompting tool access.}
The negative top-level-tools result suggests that future systems should not
only provide tools, but train the model to use them. Tool-use training could
treat calls to the verifier, diagnostic inspector, retriever, proof-state
viewer, or proof reviser as explicit actions. Successful scaffold traces could
provide imitation data for when to call each tool and how to incorporate its
output, while failed traces could provide negative examples of unproductive tool
use. This connects to general tool-use methods such as ReAct and Toolformer,
but the formal-verification setting provides a stronger supervision signal:
tool outputs can be grounded in compiler errors, proof states, and final
verification outcomes \citep{10.48550/arxiv.2210.03629,schick_toolformer_nodate}. Lean-specific
systems such as LeanDojo/ReProver and Lean Copilot also suggest that tool use,
premise retrieval, and proof-environment interaction can be treated as trainable
components rather than only as prompt-time affordances
\citep{yang2023leandojo,song2024leancopilot}.

\paragraph{Task-specific model specialization.}
All main comparisons in this chapter use the same GPT-5.3 Codex base model in
order to isolate scaffold effects, but future systems need not use a single
model for every decision. Different parts of the pipeline may benefit from
different model capabilities. A large general model may be useful for high-level
decomposition, while a smaller Lean-specialized model may be better for tactic
selection, premise retrieval, syntax repair, or local proof completion. A
router could assign subtasks to specialized models based on the current proof
state, failure mode, or expected cost. This would turn the scaffold into a
model portfolio rather than a single-model inference loop, potentially improving
both pass rate and cost per solve.

\paragraph{Whole-task repair and helper validation.}
The current scaffold still struggles when non-theorem regions are malformed or
underspecified. Future work should extend repair beyond theorem bodies to the
entire editable artifact: definitions, helper lemmas, specifications, imports,
and theorem statements. A whole-task validation stage could check whether helper
regions are syntactically valid, whether definitions match the intended
specification, and whether generated lemmas are actually useful for the final
proof. This would prevent the system from spending theorem-repair budget on
artifacts that are not yet valid Lean objects or that lack the supporting
structure needed for the final theorem.

\paragraph{Adaptive inference budgets.}
The experiments show that additional search can improve pass rate, but not all
tasks require the same amount of computation. Future scaffolds should allocate
budget adaptively. Easy tasks could terminate after direct repair or shallow
decomposition, while harder tasks receive deeper search, more repair attempts,
richer retrieval, or proof-reviser intervention. A learned value model could
also estimate the expected utility of additional computation, allowing the
system to optimize cost per solve rather than raw pass rate alone.

\paragraph{Combining training-time and inference-time verifier feedback.}
Finally, the two parts of this thesis suggest a combined direction. RLVR can
train a policy to produce better formal artifacts, while scaffolded inference
can generate structured traces of decomposition, tool calls, repairs, and
reviser decisions. Future work could train policies directly inside the
scaffold: models would learn not only to produce final proofs, but also to
choose tools, propose subgoals, interpret diagnostics, and decide when a branch
is no longer worth pursuing. Conversely, scaffold traces could provide training
data for RLVR or supervised fine-tuning, turning inference-time search behavior
back into improvements in the underlying policy.
\chapter*{Conclusion}
\addcontentsline{toc}{chapter}{Conclusion}
\label{ch:conclusion}

This thesis asked whether the generation of formally verified code from written
specifications can be automated with large language models (LLMs). Verified-code synthesis poses difficulties
beyond automated theorem proving: specifications must be conceived alongside the
implementation, the corpus of verified code is far smaller than the
mathematical proof libraries that drive theorem-proving systems, and the space
of valid solutions is less structured. Rather than committing to a single
solution, the thesis studied two complementary approaches: post-training the model itself, and scaffolding a fixed model
at inference time. Both were grounded in a concrete evaluation suite: the
Vericoding Benchmark for broad cross-language coverage, and Dalek Bench, a new
repository-grounded Lean benchmark constructed in this work from the
\texttt{curve25519-dalek-lean-verify} project.

Part~I showed that post-training from Dafny verifier feedback can serve as an
effective capability-distillation technique, but only when data quality, model
capacity, and reward signal are jointly controlled. The central difficulty was
not merely optimization, but reward design. The verifier deterministically
checks the specification it is given, not the task that was intended, so the
data distribution is itself part of the reward function. The initial
single-turn run reached high aggregate reward, but rollout-level analysis
revealed that much of this progress was specification hacking: trivial
postconditions admitted constant implementations, leaked helper functions
encoded the answer directly, and the staged reward paid partial credit for buildable code. Filtering the data toward faithful, low-leakage,
nontrivial specifications made the signal honest but the task much harder, and a
multi-turn verifier-feedback environment partially recovered performance,
reaching \(31.1\%\) verification within a four-attempt repair budget on the hard
subset. The broader lesson is that verifier feedback can train verified-code
capability, but only when the reward and data are faithful enough to measure the intended
task.

Part~II held the policy fixed and instead built a verifier-in-the-loop scaffold
for Lean verified coding, combining recursive subgoal decomposition, an
MCP-style tool layer, proof revision, and progress evaluation. The recurring
finding was that \emph{tool access is not the same as tool use}: naively
exposing tools to the model reduced performance, while stronger configurations
were those that directed when to decompose, repair, inspect feedback, and stop.
Structured search consistently beat raw budget, with the full scaffold and
proof reviser reaching the best pilot result of 18 of 26 tasks. The proof
reviser was especially important as an efficient local-repair mechanism,
recovering many failures without paying the full cost of recursive search. On
VERINA, a broader set from the Vericoding benchmark, most solved tasks were handled by the simpler base-plus-reviser and
reviser-plus-tools configurations; recursive decomposition's distinctive
contribution was not that more recursion was always better, but that it cracked
a few hard tasks those configurations could not solve on their own. Dalek Bench
remained considerably harder because success required recovering repository
context: local library conventions, dependency structure, existing lemmas,
naming patterns, and the surrounding proof architecture. This confirmed that
repository-grounded proof repair is a brittle and more realistic task, and does
not transfer automatically from standardized, mostly self-contained vericoding
benchmarks.

Recursive subgoal decomposition is only one loop within a broader scaffolding
design space, and it is informative to place it alongside concurrent
 work by my peer Leo Yao, who explored a complementary point in that
space for his master's thesis. Rather than a single recursive worker, his approach samples many
candidate agents in parallel, and rather than a wide array of tools, it equips
those agents with a small set of well-informed ones. Qualitatively, his results
point in the same direction as Part~II while emphasizing a different axis of
control: parallel breadth is effective at surfacing promising candidates across
VERINA and related benchmarks, but it buys that performance through additional
sampling cost. By contrast, recursive decomposition targets the hard tail of
tasks that require intermediate structure, while the proof reviser provides a
cheaper local-repair loop for cases where full decomposition is unnecessary.
Together, these systems suggest that effective scaffolding is not a matter of
adding more calls, more tools, or more recursion indiscriminately. It depends on
choosing the right control structure for the task. Tool \emph{selection} matters
more than tool \emph{breadth}; recursion helps when it creates useful
intermediate obligations, but can be wasteful when local repair is sufficient;
and progress evaluation is needed to decide which loop should run next. A
natural next system would use breadth as a first pass, decomposition for the
residual hard tasks, and a shared layer of curated tooling and progress
evaluation to keep search efficient.

Taken together, the two parts suggest that the path forward is not simply
larger models, more samples, or more tools, but better mechanisms for converting
verifier feedback into learning and search. At training time, verifier feedback
is useful when it becomes a faithful capability-distillation signal. At
inference time, verifier feedback is useful when it is embedded in a scaffold
that actively structures tool use, repair, decomposition, and stopping
decisions. The two approaches could ultimately reinforce each other. Scaffolds produce
the supervision that training needs: examples of when to decompose a task, which
tools to call, how to repair failed proofs, and when to stop. Training can then
turn those traces into model behavior, making future scaffolds cheaper, more
automatic, and less dependent on hand-designed control. This thesis does not close the gap between written
specifications and verified code, but it crystallizes several techniques for
narrowing it. That gap matters: as AI-generated code is written and deployed at
a scale that outpaces human review, machine-checkable verification is one of
the few safeguards that scales with it, and getting language models to produce
verified artifacts reliably is a problem worth continuing to attack from both of the proposed
directions.


\appendix


\defbibheading{bibintoc}{\chapter*{#1}\addcontentsline{toc}{backmatter}{\refname}} 

\printbibliography[title={\refname},heading=bibintoc]


\end{document}